\documentclass[usletter, oneside]{amsart} 

\usepackage[table]{xcolor}

\usepackage{titletoc}

\usepackage{amsmath,amsthm,amssymb,amsfonts,mathrsfs,color, mathtools,crop, graphicx, todonotes, bbm, tikz, subfig}
\usepackage{nicematrix}
\usepackage[]{geometry}
\usepackage{enumitem}
\usepackage{hyperref}

\hypersetup{
	colorlinks,
	linkcolor={blue!50!black},
	citecolor={blue!50!black},
	urlcolor={blue!80!black}
}

\usetikzlibrary{fixedpointarithmetic,calc, intersections,through,backgrounds,spy}

\definecolor{plotred}{RGB}{192 0 0}
\definecolor{darkgreen}{rgb}{0.0, 0.39, 0.0}

\theoremstyle{plain}
\begingroup
\newtheorem{claim}{Claim}

\newtheorem{theorem}{Theorem}[section]
\newtheorem*{theorem*}{Theorem}
\newtheorem*{"theorem"}{``Theorem''}
\newtheorem{corollary}[theorem]{Corollary}
\newtheorem{proposition}[theorem]{Proposition}

\newtheorem{lemma}[theorem]{Lemma}
\endgroup

\theoremstyle{definition}
\begingroup
\newtheorem{definition}[theorem]{Definition}
\endgroup

\theoremstyle{remark}
\begingroup
\newtheorem{remark}[theorem]{Remark}
\newtheorem{example}[theorem]{Example}
\endgroup 

\numberwithin{equation}{section}
\newcommand{\N}{\mathbb N} 
\newcommand{\T}{\mathbb T} 
\newcommand{\Z}{\mathbb Z} 
\newcommand{\R}{\mathbb R}
\renewcommand{\SS}{S}
\newcommand{\E}{{\mathbb E}}
\newcommand{\HH}{{\mathbb H}}
\renewcommand{\P}{{\mathbb P}}
\DeclareMathOperator{\Ric}{Ric}
\DeclareMathOperator{\inj}{inj}
\DeclareMathOperator{\conv}{Conv}
\DeclareMathOperator{\vcdim}{VC-dim}

\newcommand{\dist}{{\rm dist}}
\newcommand{\Dist}[2]{\text{dist}^{#1}\left(#2\right)}
\newcommand{\diam}{{\rm diam}}

\newcommand{\id}{\mathrm{id}}

\renewcommand{\H}{{\mathcal H}}

\newcommand{\LRa} {\Leftrightarrow}
\newcommand{\Ra} {\Rightarrow}

\renewcommand{\d}{\,\mathrm{d}}

\newcommand{\dx}{\,\mathrm{d}x}

\newcommand{\ds}{\,\mathrm{d}s}
\newcommand{\dt}{\,\mathrm{d}t}

\newcommand{\cal}{\mathcal}
\newcommand{\scr}{\mathscr}

\newcommand{\eps}{\varepsilon}
\newcommand{\average}{{\mathchoice {\kern1ex\vcenter{\hrule height.4pt
width 6pt depth0pt} \kern-9.7pt} {\kern1ex\vcenter{\hrule
height.4pt width 4.3pt depth0pt} \kern-7pt} {} {} }}

\allowdisplaybreaks

 \makeatletter
\@namedef{subjclassname@2020}{2020 Mathematics Subject Classification}
\makeatother

\newcommand{\B}{\mathcal{B}}

\newcommand{\arclength}[2]{\text{arclength}^{#1}\left(#2\right)}
\newcommand{\one}{\mathbbm{1}}
\newcommand{\red}[1]{\textcolor{red}{#1}}
\newcommand{\blue}[1]{\textcolor{blue}{#1}}
\newenvironment{claimproof}[1][Proof of Claim]{\noindent\textit{#1. }\space}{\hfill $\blacksquare$}

\begin{document}

\title[VC-Dimension and Generalization in Geometric Machine Learning ]{VC-Dimension and Generalization in Geometric Machine Learning with Applications to Active Learning}

\author{Delanna Do}
\address{Delanna Do\\
Department of Mathematics\\
University of Pittsburgh\\
Pittsburgh, PA 15213}
\email{delanna.do@pitt.edu}

\author{Gokul Bhusal}
\address{Gokul Bhusal\\
Department of Mathematics\\
University of California, Los Angeles\\
Los Angeles, CA 90095}
\email{bhusal@math.ucla.edu}

\author{Kevin Miller}
\address{Kevin Miller\\
Department of Mathematics\\
Brigham Young University, Provo\\
Provo, UT 84602}
\email{ksmiller@mathematics.byu.edu}

\author{Ekaterina Merkurjev}
\address{ Ekaterina Merkurjev \\
Department of Mathematics and CMSE \\
Michigan State University \\
East Lansing, MI 48824}
\email{merkurje@msu.edu}

\author{Stephan Wojtowytsch}
\address{Stephan Wojtowytsch\\
Department of Mathematics\\
University of Pittsburgh\\
Pittsburgh, PA 15213}
\email{s.woj@pitt.edu}

\date{\today}

% \subjclass[2020]{}
\keywords{VC-dimension, geometric machine learning, active learning, distance ball, Riemannian manifold, normed vector space}

\begin{abstract}
We compute the VC-dimension of metric ball classifiers in the model spaces of Riemannian geometry (Euclidean, spherical, hyperbolic) and present upper and lower bounds on the VC-dimension of metric balls in several other situations, including general normed vector spaces (potentially infinite VC-dimension), vector spaces with $\ell^p$-norms (VC-dimension grows at least linearly in the dimension, faster than Euclidean for $p=1$), flat tori, and general Riemannian manifolds (arbitrarily large VC-dimension after a conformal change). In several situations where the VC-dimension is large or even infinite, we prove generalization bounds under additional assumptions on the data distribution using covering arguments. Both in the VC-dimension and covering setting, we consider data on low-dimensional subsets of high-dimensional Euclidean spaces. As an application, we consider coverage-based algorithms in active learning.
\end{abstract}

\maketitle

%\setcounter{tocdepth}{1}
%\tableofcontents

\section{Introduction}

Making decisions based on incomplete information can lead to disaster \cite{romeo}. The science of statistics studies what we can say about a ground truth based on limited information, and how confident we can be in our assertions, assuming that we are given `random' examples of the underlying truth. Within this larger realm, statistical learning theory addresses such questions as how well we can expect a function $f$ to predict a `label' $y\in\{0,1\}$ given a point $x$, assuming that it does well on a randomly generated set $S= \{(x_i, y_i): i=1,\dots, n\}$. The true performance $\mathbb E_{(x,y)}|f(x) - y|$ and empirical estimate $\frac1n\sum_{i=1}^n |f(x_i) -y_i|$ are close for a fixed function $f: X\to\{0,1\}$ by the law of large numbers, but not necessarily if $f$ is selected from a function class specifically with the goal of doing well, e.g.\ by choosing a function which classifies as large a fraction of the sample set correctly as possible.

The answer to this question depends crucially on the `hypothesis class' $\mathcal H$ in which we choose the function $f$. If $\mathcal H$ contains a single element we are in the previous situation, whereas `large' hypothesis classes can contain functions which look promising on a given set $S$, but rely on spurious patterns in the finite sample which are not representative of the true distribution. A complete answer to the question is given in terms of Vapnik-Chervonenkis (VC) dimension: The VC-dimension of a hypothesis class is finite if and only if true performance and empirical estimate are {\em uniformly} close over the entire hypothesis class with high probability over the random sample for {\em any} probability distribution on the set of point-label pairs $(x,y)$. A quantitative statement is given in the Fundamental Theorem of PAC learning \cite[Theorem 6.8]{shalev2014understanding}. Uniform closeness in particular enables us to select a promising candidate on the dataset with confidence that it will perform well on new data (on average, and with high probability).

In this article, we compute and estimate the VC-dimension of several geometric hypothesis classes. Geometric machine learning leverages progress in well-studied fields like Riemannian geometry and spectral geometry for the purposes of data science, for instance by embedding data in non-Euclidean spaces or by recovering the intrinsic geometry of data in a Euclidean space using data graph approximations to a `data manifold'. Our focus is on classifiers based on distance, the coarsest geometric quantity. Specifically, we study the VC-dimension of the class of metric ball classifiers $\{1_{B_r(x)} : x\in X, r>0\}$ in various metric spaces $X$ and for fixed or variable $r>0$ and both open and closed balls. The cases we consider cover the most common spaces for data embeddings: Euclidean (flat), spherical (positively curved), hyperbolic (negatively curved), and vector spaces with general norms. We obtain partial results for quotients of such spaces, a useful notion for data which should be considered periodic in one or multiple directions. Given a bound on the VC-dimension of metric ball classifiers, we can also bound the VC-dimension of binary classifiers associated to unions of metric balls $1_{\bigcup_{i=1}^k B_r(x_i)}$.

We briefly summarize our results -- more precise statements can be found in the text below. 

\begin{enumerate}
\item The VC-dimension of metric ball classifiers in Euclidean and hyperbolic spaces of dimension $d$ is $d+1$, but in $d$-dimensional spheres, it is $d+1$ only if the radius of balls is either fixed or, if it is variable, constrained so that balls cannot cross the equator. It is $d+2$ for balls with unconstrained variable radius.

\item In quotients of Euclidean, hyperbolic and spherical spaces, the VC-dimension of metric ball classifiers with small radius coincides with that of small balls in the model geometry. For large balls, it may be larger due to the more complicated global structure. In flat tori $\R^d/\Z^d$, the VC-dimension grows at least as $5d/2 + O(1)$.

\item A similar lower bound $5d/3 + O(1)$ on the growth is true in $\R^d$ if Euclidean balls are replaced by $\ell^1$-balls. Both in $\ell^1(\R^d)$ and $\R^d/\Z^d$, a weaker upper bound $O(d\log d)$ on the VC-dimension holds.

\item If $d\geq 3$, the VC-dimension of metric ball classifiers with respect to a norm on $\R^d$ may be infinite. If $d\geq 3$ and $n\in\N$, for any Riemannian metric $g$ on a $d$-dimensional manifold $M$, there exists a conformally equivalent metric $\tilde g = u^2g$ for which the VC-dimension of metric ball classifiers exceeds $n$.

\item Even for points on the graph of a smooth function $f:\R^k\to\R^{d-k}$ (or a low-dimensional submanifold), it is not possible to obtain bounds on the number of points shattered by ambient space Euclidean balls purely in terms of the intrinsic dimension of the graph. This is true even if $f$ is small in $C^m$-norm (for any given $m\in\N$).
\end{enumerate}

Of particular didactic interest in our work is the observation that the family of binary classifiers $h_{x,r}:\R^d\to \{0,1\}$ which assign label $1$ to the interior of an $\ell^1$-ball $B^1_r(x) = \{y\in\R^d : \|x-y\|_1< r\}$ in $\R^d$ has $d+1$ parameters $x\in\R^d,\ r>0$, but VC-dimension at least roughly $5d/3$. While examples where the VC-dimension differs significantly from the number of parameters of a parametrized hypothesis class are well known (and several new ones are derived in this work), this one is particularly easy to prove and particularly natural with an eye towards applications.

While finite VC-dimension is both necessary and sufficient for PAC-learnability, often the guarantees of PAC-learning are not required in full generality. We can, for instance, make weaker assumptions on the hypothesis class at the cost of making stronger assumptions on the distribution of data. VC-dimension is a combinatorial concept which can be highly sensitive to, for instance, minor changes to the center $x$ of the ball $1_{B_r(x)}$. If the data distribution $\mu$ is such that minor changes to $x$ only cause minor changes to $1_{B_r(x)}$ in $L^1(\mu)$, on the other hand, good generalization bounds can be obtained by other means, such as covering arguments. In the second half of the article, we therefore address generalization guarantees based on covering bounds alongside those derived from VC-dimension.

As a particular instance, we observe that learning problems can be hard (depending on the dimension of an ambient embedding space) for general distributions, but easy (depending only on the intrinsic dimensionality of data) for `good' distributions, lending new support to a classical observation \cite{narayanan2009sample}. This encompasses even problems for which intrinsic distances are merely approximated, e.g.\ from estimates using a data cloud.

A particular motivation for this work are applications to coverage-based active learning, specifically algorithms such as CoreSet \cite{sener2017active} and ProbCover \cite{yehuda2022active}. In these applications, the goal is to cover the support of the data distribution $\mu$ by balls $B_r(x_{i_j})$ centered at $k$ points in a dataset $S = \{x_i : i=1,\dots, n\}$. The $k\ll n$ centers are selected for `data annotation', i.e.\ we obtain labels $y_{i_j}$ for the selected points for a downstream application. The covering sub-problem is used to select a `representative subset' $S' = \{x_{i_j} : j= 1,\dots,k\}$ of $S$ which falls within our annotation budget $k$. The coverage problem can be understood as a particular type of binary classification problem in which all points belong to the same class, i.e.\ the goal is to cover the entire space.

The article is structured as follows. Sections \ref{section half-spaces} and \ref{section vc metric balls} are devoted to the study of VC-dimension: In Section \ref{section half-spaces}, we compute the VC-dimension of families of (open or closed) half-space classifiers which are either required to contain the origin or cannot contain the origin. From this, we deduce the VC-dimension of metric ball classifiers in the model spaces of Riemannian geometry (Section \ref{section prototypical geometries}). We generalize this to obtain bounds on the VC-dimension of metric ball classifiers with small radius in quotients of the model spaces in Section \ref{section quotients}, along with partial results for the VC-dimension of metric balls in flat tori. Subsequently, we investigate VC-bounds in terms of intrinsic data dimension in a high-dimensional ambient Euclidean space in Section \ref{section intrinsic dimension} and bounds on the VC-dimension of balls in normed vector spaces in Section \ref{section normed spaces}, both for $\ell^p$-norms and general norms. Many of the positive results in Section \ref{section vc metric balls} are summarized in Table \ref{table vc-dim list}.

In Section \ref{section generalization}, we apply our bounds to derive generalization guarantees of metric ball classifiers in various spaces. As the VC bounds are vacuous in several examples considered above, we derive alternative covering-based generalization guarantees under stronger assumptions on the data distribution. Applications to supervised learning and active learning are considered in Section \ref{section applications}. Many proofs are given in the appendix.

While results in this article are related by a common theme, proofs in different sections are by and large unrelated on a technical level and can be read separately.

\subsection{Concepts and Conventions}

We refer the reader to \cite{shalev2014understanding} for the definition of the VC-dimension and classical results linking it to statistical learning theory, such as the Fundamental Theorem of PAC learning and the Sauer-Shelah Lemma. 

For concepts of Riemannian geometry as needed for Theorem \ref{theorem quotient spaces} in Section \ref{section quotients} and for Section \ref{section riemannian}, we refer the reader for instance to any of the introductory texts \cite{gallot1990riemannian, petersen2006riemannian, lee2018introduction}. Section \ref{section prototypical geometries} requires no deeper familiarity with Riemannian geometry than knowledge of spheres in Euclidean space and the hyperboloid model for hyperbolic spaces.

\section{VC-dimension of Half-space Classifiers}\label{section half-spaces}

In all three Riemannian model geometries -- Euclidean, spherical, hyperbolic -- we can reduce the computation of VC-dimension to that of certain classes of half-spaces in a vector space. We therefore present the following results on linear algebraic, not geometric, classifiers.

\begin{definition}
     Let $w\in\R^d\setminus \{0\}$ and $b \in \R$. Then we call $H_{w, b} = \{x\in\R^d : w^Tx\geq b\}$ a \textit{closed half-space} and $H_{w, b}^\circ = \{x\in\R^d : w^Tx > b\}$ an \textit{open half-space}. Let $\H$ denote the family of all closed half-spaces and $\H^\circ$ be the family of all open half-spaces. 
\end{definition}

For a given family of sets $\mathcal F$, let the \textit{element-wise complement of $\mathcal F$}, $\{X \setminus F : F \in \mathcal F\}$, be denoted by $\mathcal F^c$. By abuse of notation, we identify the families of sets with the associated hypothesis classes of classifiers that assign labels 1 to elements in a set and 0 to its complement. That is, we identify $F$ with $\one_F$ for each $F \in \mathcal F$. 

\begin{theorem} {\cite[Theorem 9.3]{shalev2014understanding}}
    The VC-dimension of $\H$ and $\H^\circ$ is $d+1$. 
\end{theorem}

Similarly to the way that the VC-dimension of $\H$ and $\H^\circ$ can be used to determine the VC-dimension of the families of Euclidean balls $\B$ and $\B^\circ$ in $\R^d$, we develop an analogous set of results on the VC-dimension of several restricted families of half-spaces in order to determine the VC-dimension of the families of geodesic balls in other constant curvature spaces.

\begin{definition}
   We call a half-space $H_{w, b}$ \textit{strictly exterior} if $0 \notin \overline{H_{w, b}}$ (i.e.\ if $b > 0$), \textit{strictly interior} if $0\in H_{w, b}^\circ$ (i.e.\ if $b < 0$), and \textit{homogeneous} if $0 \in \partial H_{w,b}$ (i.e.\ if $b = 0$). We call a half-space interior (exterior) if it is either strictly interior (exterior) or homogeneous.
   
    From these different classifications of half-spaces, we define the following restricted families of half-spaces:
    \begin{align*}
        \H_{<} &= \{H_{w, b} :H_{w, b} \text{ closed strictly interior half-space}\}, \\
        \H_{>} &= \{H_{w, b} :H_{w, b} \text{ closed strictly exterior half-space}\}, \\
        \H_0 &= \{H_{w, b} :H_{w, b} \text{ closed homogeneous half-space}\},\\
        \H_{\leq } &= \{H_{w, b} :H_{w, b} \text{ closed interior half-space}\} = \H_0\cup \H_{<}, \\
        \H_{\geq} &= \{H_{w, b} :H_{w, b} \text{ closed exterior half-space}\} = \H_0 \cup \H_>, 
    \end{align*}
    along with their open counterparts $\H^\circ_<,\ \H^\circ_{\leq},\ \H^\circ_0,\ \H^\circ_\geq$ and $\H^\circ_>$.
\end{definition}

\begin{theorem}{\cite[Theorem 9.2]{shalev2014understanding}}
    The VC-dimension of $\H_0$ and $\H_0^\circ$ is $d$.
\end{theorem}

From this established result, we derive the following extensions. 

\begin{lemma} \label{lm:half-space_d}
The VC-dimension of $\H_{>}$, $\H_{>}^\circ$, $\H_{<}$, $\H_{<}^\circ$, $\H_{\leq}$, and $\H_{\geq}^\circ$ is $d$. 
\end{lemma}

Perhaps surprisingly, there are two restricted classes of half-spaces which have the full VC-dimension of $\H$ or $\H^\circ$.

\begin{lemma} \label{lm:half-space_d+1}
    The VC-dimension of $\H_{\geq}$ and $\H_{\leq}^\circ$ is $d+1$.
\end{lemma}

\begin{remark}\label{remark contain 0}
The discrepancy in VC-dimension between the different classes of half-spaces studied in Lemmas \ref{lm:half-space_d} and \ref{lm:half-space_d+1} can be intuitively explained by the fact that the other restricted half-space families are collections of half-spaces that either all necessarily contain 0 or all necessarily do not contain 0. Comparatively, $\H_\geq$ contains both types of half-spaces. This can be exploited to liken $\H_\geq$ to the broader family $\H$ which has VC-dimension $d+1$.
\end{remark}

In the spirit of Remark \ref{remark contain 0}, we also note the following distinction, which is generally missed by textbooks discussing the VC-dimension of homogeneous half-spaces and half-spaces.

\begin{lemma}\label{lemma vc-dimension of homogeneous half-spaces}
The VC-dimension of $\H_{0} \cup \H_0^\circ$ is $d+1$.
\end{lemma}

Thus, the concept classes of open or closed homogeneous half-spaces individually have VC-dimension $d$, but their union has VC-dimension $d+1$, illustrating the subtlety of the notion. Some results in this section are summarized in Table \ref{table vc-dim list}.

\section{VC-dimension of Metric Ball Classifiers}\label{section vc metric balls}

\subsection{Setting and Notation}

Given a metric space $(X,d)$, consider the following respective families of closed and open balls with fixed radius $r > 0$:
\begin{equation}\label{eq ball classes 1}
\B_r = \big\{\overline{B}_r(x) : x\in X\big\},  \qquad \B_r^\circ = \big\{B_r(x) : x\in X\big\}. 
\end{equation}
Furthermore, for $I \subseteq (0, \infty)$, consider the following respective families of closed and open balls with varying radius $r \in I$,
\begin{equation}\label{eq ball classes 2}
\B_{I} = \bigcup_{r \in I}\B_r, \qquad \B_{I}^\circ = \bigcup_{r \in I}\B_r^\circ, \qquad\text{and simply } \B:= \B_{(0, \infty)}, \quad \B^\circ := \B_{(0, \infty)}^\circ.
\end{equation}

In the following, we establish bounds on the VC-dimension of these hypothesis classes in several important families of metric spaces, always suppressing the dependence of $\mathcal F$ on $(X,d)$ when it is clear from context. As a general result, we can focus on only open or closed balls.

\begin{lemma}\label{lemma closed vs open balls}
Let $(X,d)$ be a metric space.
\begin{itemize}
\item Assume that a finite set $S$ is shattered by $\B_r$. Then there exists $\eps>0$ such that  $S$ is shattered by both $\B_s$ and $\B_s^\circ$ for any $s\in(r,r+\eps)$.

\item Assume that a finite set $S=\{x_1,\dots, x_n\}$ is shattered by $\B_r^\circ$. Then there exists $\eps>0$ such that  $S$ is shattered by both $\B_s$ and $\B_s^\circ$ for any $s\in(r-\eps,r)$.

\item If a set is shattered by a family of (open or closed) balls of radius $r>0$, we can always choose a different radius $s$ and $\eps>0$ such that for all $I \in \mathcal P(\{1,\dots, n\})$, there exists $x_I \in X$ such that
\[
d(x_I, x_i)\ <\ s-\eps\ <\ s+ \eps\ <\ d(x_I, x_j) \qquad \forall\ i \in I,\ j\in I^c.
\]
In particular, the set can be shattered both by families of open and closed balls of the same radius with the same centers.

\item A finite set is shattered by $\B$ if and only if it is shattered by $\B^\circ$.
\end{itemize}
\end{lemma}

\subsection{Prototypical spaces in Riemannian Geometry} 
\label{section prototypical geometries}
Recall the following classical result.

\begin{proposition}\cite[Theorem 1]{dudley1979balls}\label{theorem flat space}
Let $X=\R^d$ equipped with the Euclidean metric on $\R^d$. Then the VC-dimension of any of the hypothesis classes in \eqref{eq ball classes 1}, \eqref{eq ball classes 2} is $d+1$.
\end{proposition}

We note that this result is at times misquoted to claim that the hypothesis classes have VC-dimension $d+2$.\footnote{\ This miscommunication is likely exacerbated by the fact that Dudley shows that $V(\mathcal B) = d+2$ where $V(\mathcal F) = \vcdim(\mathcal F)+1$ in modern terminology.} The weaker upper bound $d+2$ on the VC-dimension can be established more easily by applying \cite[Theorem A]{dudley1979balls}, while a key geometric insight used in Dudley's Theorem 1 provides the sharper result. In fact, Dudley demonstrates that his Theorem 1 {\em cannot} be derived from his Theorem A. 

The discrepancy between $d+1$ in Dudley's Theorem 1 and $d+2$ in Dudley's Theorem A is in fact a necessary artifact of differing proof strategies: Theorem A assumes a vector space structure which implies closure under complements, i.e.\ either label could be assigned to the interior and exterior of the ball respectively. If this is indeed the case, the shattering behavior changes.

\begin{corollary}\label{corollary complements}
Let $(X,d)$ be $\R^d$ with the Euclidean norm. Then $\vcdim(\B \cup \mathcal \B^c) = d+2$.
\end{corollary}

A more detailed discussion of the improvement from $d+2$ to $d+1$ for ball classifiers with fixed interior label can be found in Appendix \ref{appendix constant curvature}. In a nutshell, Dudley's key geometric insight lies in the fact  that for any $x, y\in \R^d$ and all $r, s>0$, the sets $B_r(x) \setminus B_s(y)$ and $B_s(y)\setminus B_r(x)$ can be separated by a hyperplane. Thus, one may compare the VC-dimension of the family of Euclidean balls to the VC-dimension of the family of \textit{half-spaces} for an upper bound. The matching lower bound is obtained directly by shattering a set of orthonormal vectors along with the origin.

The standard geodesic balls on $S^d \subseteq \R^{d+1}$ are given by the intersection of $S^d$ with half-spaces in $\R^{d+1}$. Hence the first of our extension theorems of constant curvature spaces addresses the VC-dimension of geodesic ball families on $S^d$ and the upper bounds follow from directly applying the above lemmas. Depending on the radius of the balls, the half-spaces may be interior or exterior.

\begin{figure}
\centering
\includegraphics[width=0.15\linewidth]{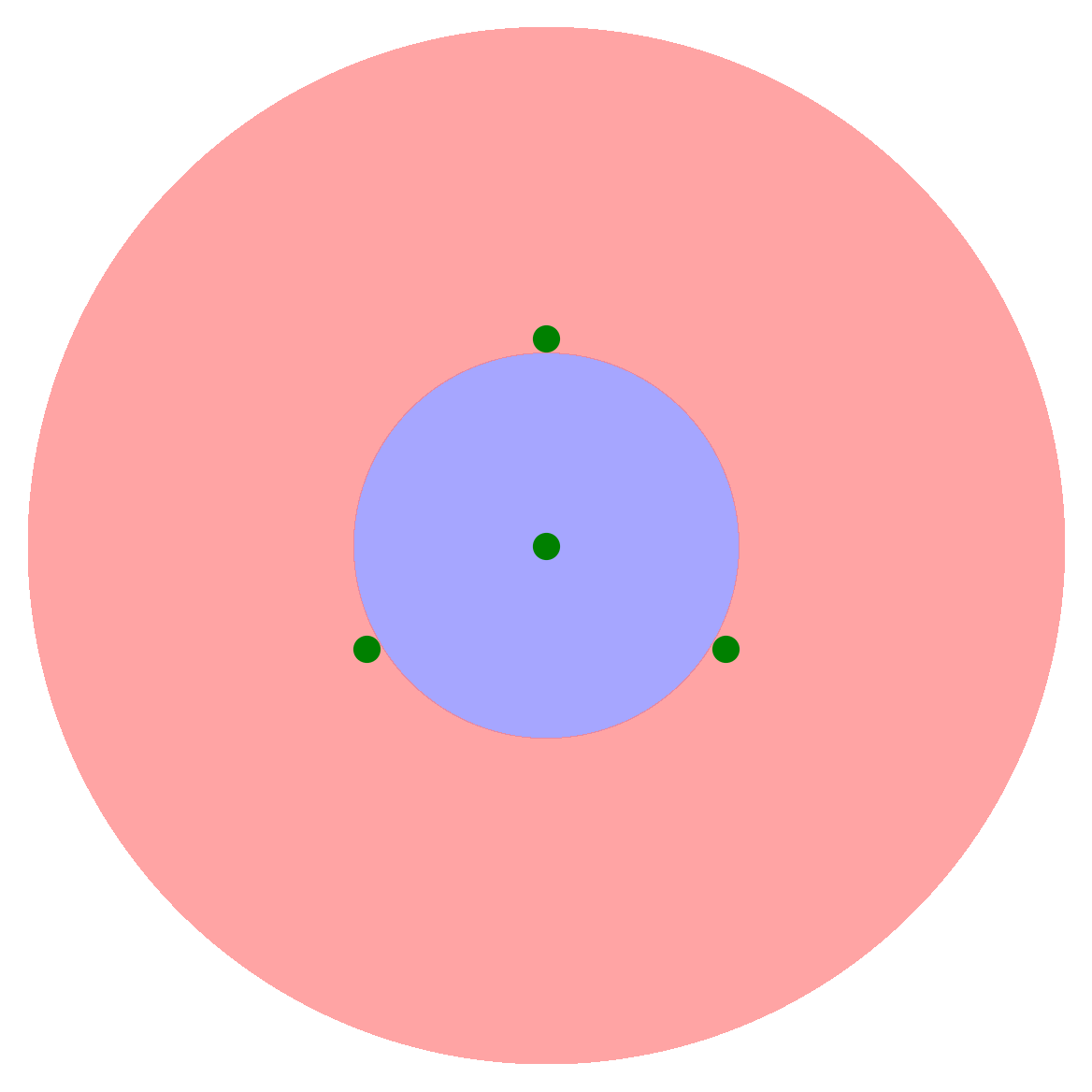}
\includegraphics[width=0.15\linewidth]{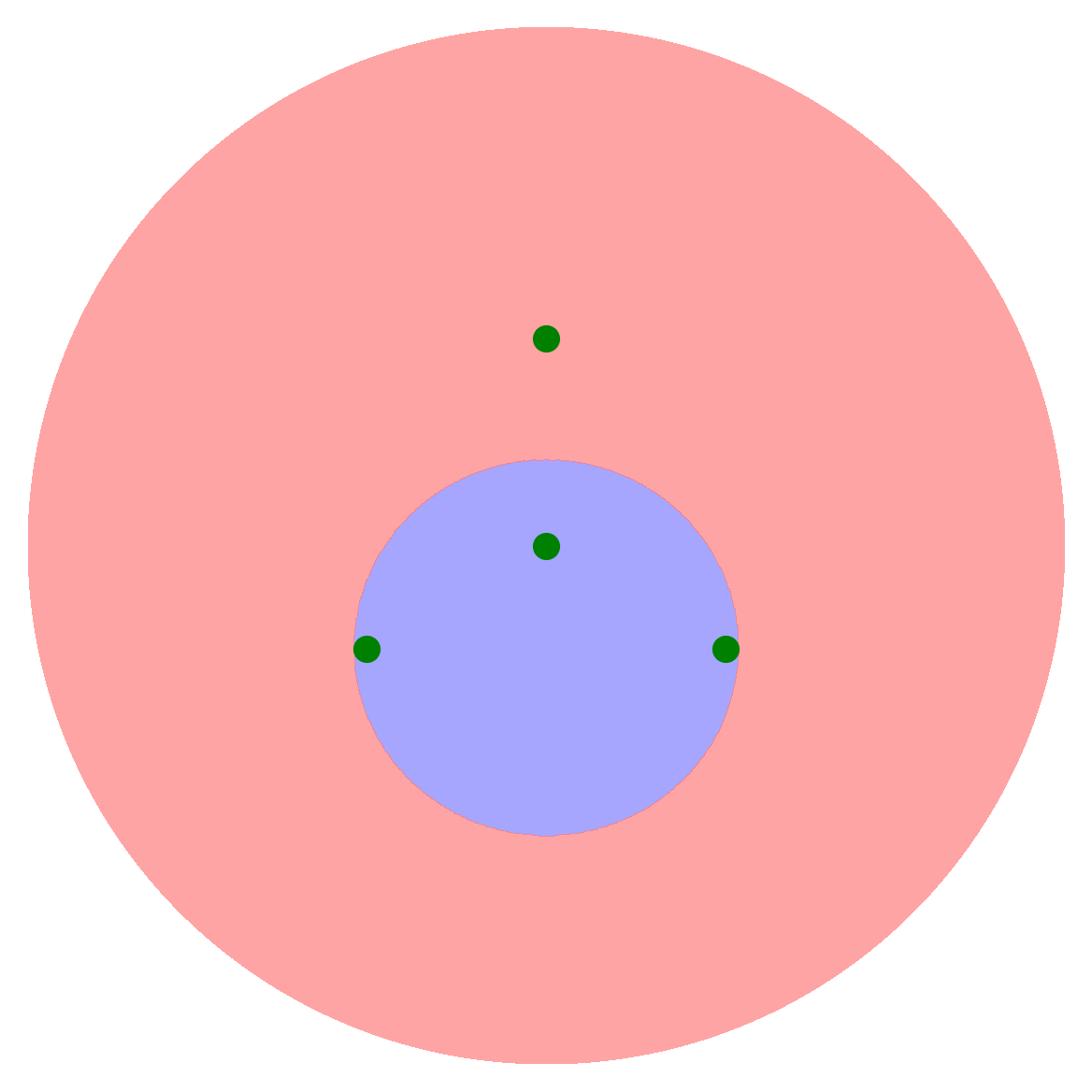}
\includegraphics[width=0.15\linewidth]{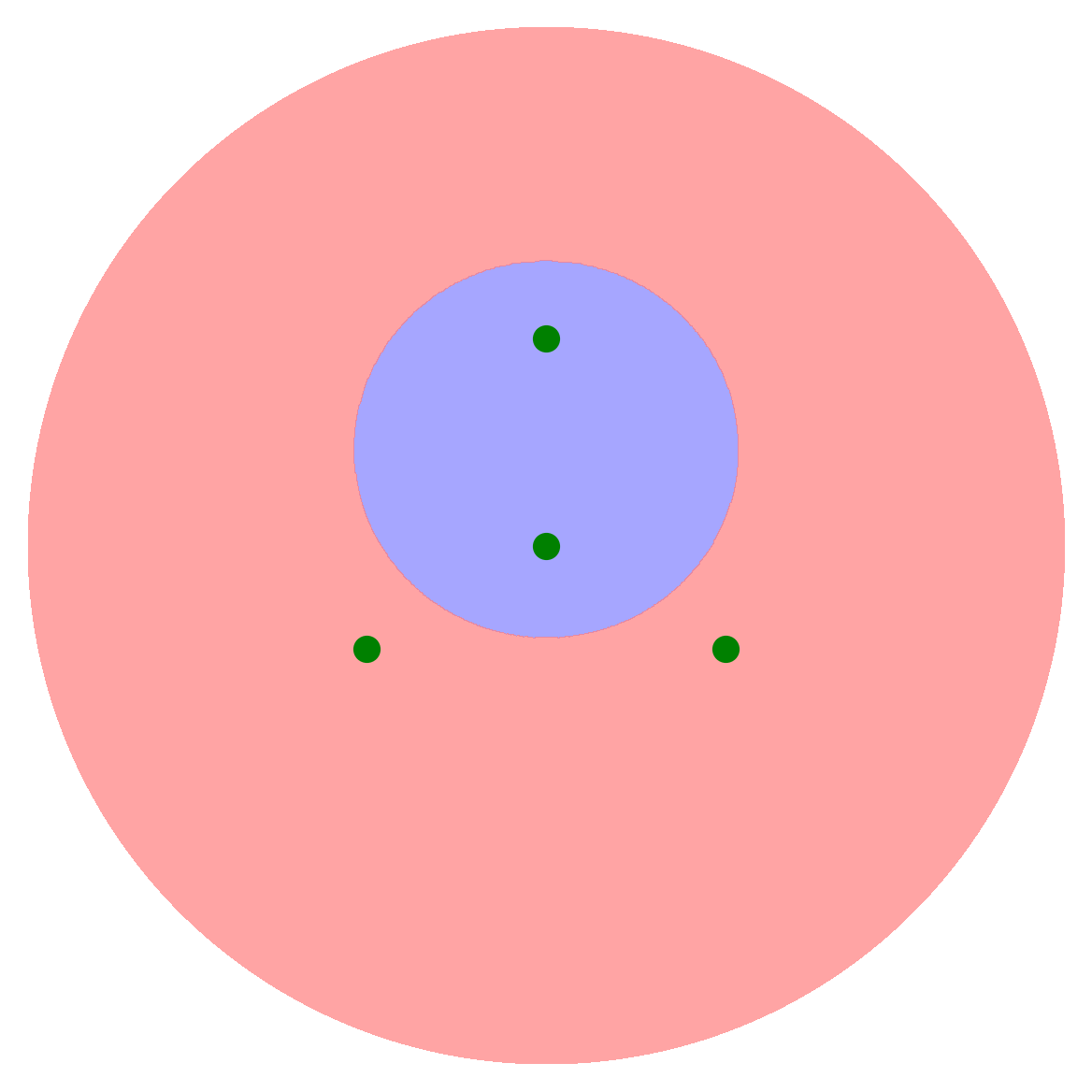}
\includegraphics[width=0.15\linewidth]{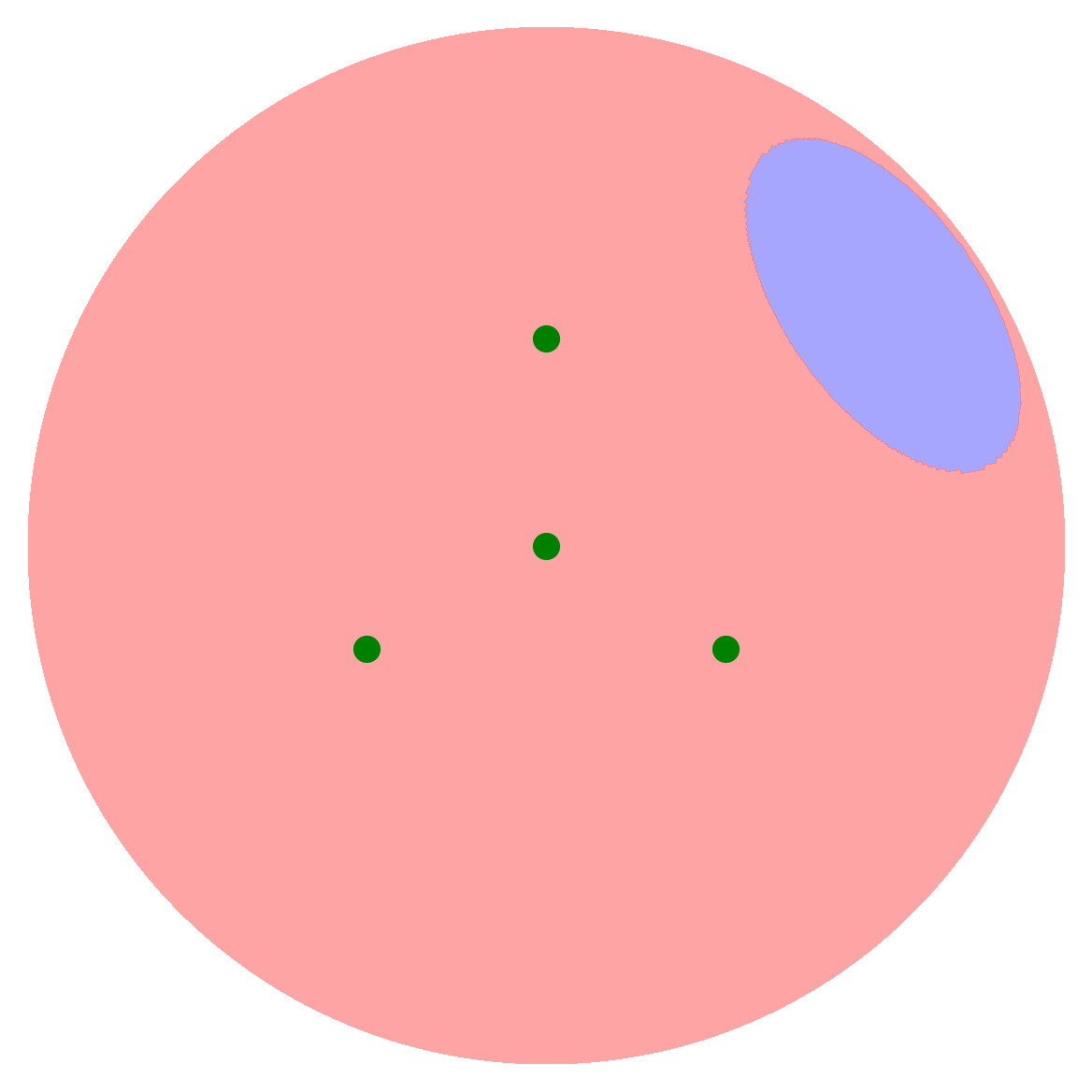}

\caption{\label{figure sphere}
Orthogonal projection of the upper hemisphere of $S^2$ onto the $xy$-plane. The blue shaded region and the red shaded region are complementary metric balls $B_r(x)$ and $B_{\pi - r}(-x)$ with respect to the intrinsic distance of $S^2$ respectively. The four-point set is shattered by the class of all metric balls -- in fact, it suffices to consider all balls of a single suitable radius $r$ and the complementary radius $\pi -r$ to shatter the four points. {\bf First, second:} One-point sets/three-point sets. {\bf Third:} Two-point sets. {\bf Fourth:} All/none. By symmetry, these are the only possible geometric arrangements. 
}
\end{figure}

\begin{theorem}\label{theorem sphere}
Let $X=S^d \subseteq \R^{d+1}$ equipped with the intrinsic (cosine similarity) distance 
\[
\dist(x, y) = \arccos(\langle x, y\rangle), \qquad \langle x, y\rangle = \sum_{i=1}^{d+1} x_iy_i.
\]
Then 
\begin{itemize}
    \item The VC-dimension of the fixed radius classes $\B_r$ and $\B_r^\circ$ is $d+1$ if $r<\pi$ (and $0$ if $r>\pi$).
    \item The VC-dimension of the bounded variable radius classes $\B_{(0, \pi/2)}, \B^\circ_{(0, \pi/2)},  \B^\circ_{(0, \pi/2]}$ is $d+1$. However, the VC-dimension of $\B_{(0, \pi/2]}$ is $d+2$.
    \item The VC-dimension of the unconstrained classes $\B$ and $\B^\circ$ is $d+2$.
\end{itemize}
\end{theorem}

The fact that $\B_r$ and $\B_r^\circ$ behave as in Euclidean spaces for small values of $r$ is unsurprising. If $r\geq \pi$, then $\overline B_r(x)$ coincides with the entire sphere for any $x\in S^d$, so $\overline B_r = \{S^{d}\}$ cannot shatter even a singleton set for $r\geq \pi$. The same is true for open balls of radius $r>\pi$, while $B_\pi(x) = S^d\setminus \{-x\}$, i.e.\ $\B^\circ_\pi$ can shatter singletons but no set of cardinality $\geq 2$.

The claim on $\B_{(0, \pi/2)}, \B^\circ_{(0, \pi/2)},  \B^\circ_{(0, \pi/2]},$ and $\B_{(0, \pi/2]}$ follows from the above lemmas on restricted half-space families. Lastly, the more subtle behavior of $\B$ and $\B^\circ$ owes to the fact that they are (essentially\footnote{\ Following the spirit, but not the letter, of Lemma \ref{lemma closed vs open balls}.}) closed under taking complements: in a sphere, the complement of an open ball is a closed ball and vice versa: $B_r(x)^c = \overline{B_{\pi -r}(-x)}$ -- see also Figure \ref{figure sphere} for an illustration. 

Euclidean spaces and spheres are the prototypical examples of flat and positively curved spaces respectively. The spherical distance or `cosine similarity distance' serves as a proxy in vector-database retrieval to measure how closely a document is aligned with a user's query.

We establish that the bound of the Euclidean case also applies to hyperbolic spaces, the prototypical spaces of negative curvature, without any distinctions as in Theorem \ref{theorem sphere}. Data embeddings into hyperbolic spaces rather than Euclidean spaces are popular as they often require much lower dimension, in particular for hierarchical data \cite{sarkar2011low, nickel2017poincare, chamberlain2017neural, sala2018representation, tifrea2018poincar}. However, algorithms and theoretical analyses often require modification from the Euclidean setting \cite{ganea2018hyperbolic,weber2020robust}. In this case, we demonstrate that metric ball classifiers behave exactly as expected.

\begin{theorem}\label{theorem hyperbolic}
In the $d$-dimensional hyperbolic space $\mathbb H^d$, all hypothesis classes described in \eqref{eq ball classes 1} and \eqref{eq ball classes 2} have VC-dimension $d+1$.
\end{theorem}

For simplicity, we presented the results for spherical and hyperbolic spaces of curvature $1$ and $-1$ respectively, but analogous results apply immediately for any (positive or negative) constant curvature by simple scaling: If all distances are scaled by a factor $\lambda$, curvature scales as $\lambda^{-2}$. VC-dimension results in Euclidean and hyperbolic spaces are scale-invariant while Theorem \ref{theorem sphere} applies to any sphere of constant curvature $K>0$ with the threshold value $\pi$ replaced by $\pi/\sqrt K$. The main results of this section are summarized in Table \ref{table vc-dim list}.

\subsection{Normed spaces}\label{section normed spaces}

In all examples so far, the VC-dimension of the various classes of metric ball classifiers was equal to either $d+1$ or $d+2$, where $d$ was the dimension of the parameter space $X$ in which the centers of balls could be selected. The same is not true even for $\ell^p$-norms on $\R^d$.

\begin{example}\cite{despres2014vapnik, gomez2021vapnik}\label{example linfty}
Let $X= \R^d$ equipped with the $\ell^\infty$-norm $\|x\|_\infty = \max_i|x_i|$. In this metric space, $\B$ and $\B^\circ$ both have VC-dimension $\lfloor (3d+1)/2\rfloor$ where $\lfloor z\rfloor$ denotes the largest integer $n\leq z$ for $z\in\R$.
\end{example}

To the best of our knowledge, the VC-dimension of $\ell^p$-balls on $\R^d$ is unknown except for $p=2$. We obtain the following partial results.

\begin{theorem}\label{theorem vc normed vector spaces}
\begin{enumerate}
\item Denote by $\B(p,d)$ the class of closed $\ell^p$-balls on $\R^d$. Then
\[
\liminf_{p\to p_0} \vcdim(\B(p,d)) \geq \vcdim(\B(p_0,d)).
\]

\item For $p=1$, we have
\[
5 \left\lfloor \frac d3\right\rfloor \leq \vcdim(\B(1,d)) \leq \max\big\{4(d+1)\,\log_2(4(d+1)),\ 110\big\}
\]

\item In particular:
\[
\liminf_{p\to 1} \vcdim(\B(p,d)) \geq 5\left\lfloor \frac d3\right\rfloor,\quad 
\liminf_{p\to \infty} \vcdim(\B(p,d)) \geq \left\lfloor \frac{3d+1}2\right\rfloor,
\]
i.e.\ there are `smooth' $\ell^p$-norms with uniformly convex unit balls which attain the same lower bounds as the extreme $\ell^p$-norms with polygonal unit balls.

\item Let $\|\cdot\|$ be a norm on $\R^2$. Then $\vcdim(\B) = 3$.

\item Let $d\geq 3$. Then there exists a norm on $\R^d$ such that $\vcdim(\B_1) = +\infty$.

\item Let $d\geq 3$ and $\eps>0$. Then there exists a norm on $\R^d$ such that 
\[
\left| \frac{\|x\| - \|x\|_2}{\|x\|_2}\right| < \eps \quad\forall\ x\neq 0, \qquad \vcdim(\B) = +\infty.
\]
\end{enumerate}
\end{theorem}

In particular, for dimension $d\gg 1$, the known bounds suggest that there exists $p_d>1$ such that $\vcdim(\B(p,d))$ is maximal for $p\in[1,p_d)$ in the sense that it beats both $\ell^2$- and $\ell^\infty$-balls. Whether there exists a norm on $\R^d$ such that $\B$ has VC-dimension $\leq d$ remains open. By direct experimentation, one may verify that $\vcdim(\B_r(1,3)) \geq 5> \vcdim(\B_r(2,3))$, i.e.\ even in three dimensions and with fixed radius, $\ell^1$-balls are able to shatter larger sets than $\ell^2$-balls of arbitrary radius. For an example, see Table \ref{table shattering fixed radius R3} in the Appendix.

The difference between the final two claims is whether the radius is fixed or variable (in which case we can prove a stronger result with a norm which is $C^0$-close to an inner product norm).

\renewcommand{\arraystretch}{1.2}
\begin{table}
\begin{tabular}{|p{0.45\textwidth}|p{0.45\textwidth}|}
    \hline
    \multicolumn{2}{|c|}{\cellcolor{gray!30} \textbf{Half-spaces in $\R^d$}} \\
    \hline
    \centering{\cellcolor{gray!30} \textbf{VC-dimension = $d$}} & \centering{\cellcolor{gray!30} \textbf{VC-dimension = $d+1$}}  \tabularnewline
    \hline

    \begin{itemize}[leftmargin=*, itemsep=0.5em, after = \vspace{-2mm}]
        \item Open homogeneous half-spaces $\H_0^\circ$
        \item Closed homogeneous half-spaces $\H_0$
        \item Closed strictly exterior half-spaces $\H_>$
        \item Closed strictly interior half-spaces $\H_<$
        \item Closed interior half-spaces $\H_\leq$
    \end{itemize}
     & 
     \begin{itemize}[leftmargin=*, itemsep=0.5em]
        \item All open half-spaces $\H^\circ$
        \item All closed half-spaces $\H$
        \item All homogeneous half-spaces $\H_0^\circ \cup \H_0$
        \item Closed exterior half-spaces $\H_\geq$
    \end{itemize} 
    \\
    \hline
\end{tabular}

\vspace{2mm}

\begin{tabular}{|p{0.45\textwidth}|p{0.45\textwidth}|}
    \hline
    \multicolumn{2}{|c|}{\cellcolor{gray!30} \textbf{Euclidean balls in $\R^d$ and metric balls in $\HH^d$}} \\
    \hline
    \centering{\cellcolor{gray!30}\textbf{VC-dimension = $d+1$}} & \centering{\cellcolor{gray!30}\textbf{VC-dimension = $d+2$}}  \tabularnewline
    \hline
    \begin{itemize}[leftmargin=*, itemsep=0.5em, after = \vspace{-2mm}]
        \item Open balls of fixed radius $\B_r^\circ$
        \item Closed balls of fixed radius $\B_r$
        \item Open balls of variable radius $\B^\circ$
        \item Closed balls of fixed radius $\B_r$
    \end{itemize} &
    \begin{itemize}[leftmargin=*, itemsep=0.5em]
        \item The union of the class of open balls of fixed radius and their complements $\B_r \cup \B_r^c$
        \item The union of the class of open balls of variable radius and their complements $\B \cup \B^c$
    \end{itemize}
    \\
    \hline
\end{tabular}

\vspace{2mm}

\begin{tabular}{|p{0.45\textwidth}|p{0.45\textwidth}|}
    \hline
    \multicolumn{2}{|c|}{\textbf{\cellcolor{gray!30}Metric balls in $S^d$}} \\
    \hline
    \centering{\cellcolor{gray!30}\textbf{VC-dimension = $d+1$}} & \centering{\cellcolor{gray!30}\textbf{VC-dimension = $d+2$}}  \tabularnewline
    \hline
    \begin{itemize}[leftmargin=*, itemsep=0.5em]
        \item Open balls of fixed radius $\B_{r}^\circ$, $r < \pi$
        \item Closed balls of fixed radius $\B_{r}$, $r < \pi$
        \item Open balls of variable radius $\B_{(0, \pi/2]}^\circ$
        \item Closed balls of variable radius $\B_{(0, \pi/2)}$
    \end{itemize}
     & 
     \begin{itemize}[leftmargin=*, itemsep=0.5em, after = \vspace{-2mm}]
        \item Open balls of variable radius $\B_{(0, \pi)}^\circ$
        \item Closed balls of variable radius $\B_{(0, \pi)}$
        \item Closed balls of variable radius $\B_{(0, \pi/2]}$
        \item Balls of fixed radius $r < \pi$ and their complements $\B_r \cup \B_r^c$
    \end{itemize} \\
    \hline 
\end{tabular}
    \vspace{2mm}

\begin{tabular}{|p{0.45\textwidth}|p{0.45\textwidth}|}
    \hline
    \multicolumn{2}{|c|}{\textbf{\cellcolor{gray!30}Metric balls in other spaces}} \\
    \hline
    \centering{\cellcolor{gray!30}\textbf{VC-dimension  $\geq 5\lfloor d/2\rfloor$}} & \centering{\cellcolor{gray!30}\textbf{VC-dimension  $\geq 5\lfloor d/3\rfloor$}} \tabularnewline
    \hline
    \begin{itemize}[leftmargin=*, itemsep=0.5em, after = \vspace{-2mm}]
        \item Open or closed balls of variable radius $\B$ or $\B^\circ$ in the flat torus $\T^d$
    \end{itemize}
     & 
     \begin{itemize}[leftmargin=*, itemsep=0.5em, after = \vspace{-2mm}]
        \item Open or closed balls of variable radius $\B$ or $\B^\circ$ in $\R^d$ with respect to the $\ell^1$-norm
    \end{itemize} \\
    \hline
    \centering{\cellcolor{gray!30}\textbf{VC-dimension  potentially infinite}} & \centering{\cellcolor{gray!30}\textbf{VC-dimension  potentially infinite (continued)}} \tabularnewline
    \hline
    \begin{itemize}[leftmargin=*, itemsep=0.5em, after = \vspace{-2mm}]
        \item Open or closed balls of fixed radius $\B_r$ or $\B^\circ_r$ in general normed vector spaces
        \item Open or closed balls of variable radius $\B$ or $\B^\circ$ in general normed vector spaces
    \end{itemize}
     & 
     \begin{itemize}[leftmargin=*, itemsep=0.5em, after = \vspace{-2mm}]
        \item Open or closed balls of fixed radius $\B_r$ or $\B^\circ_r$ in non-compact Riemannian manifolds
        \item Open or closed balls of variable radius $\B$ or $\B^\circ$ in Riemannian manifolds
    \end{itemize} \\
    \hline 
\end{tabular}
\vspace{2mm}

    \caption{\label{table vc-dim list}
    VC-dimension of some hypothesis classes based on geometric sets in constant curvature Riemannian manifolds and normed vector spaces. For precise statements, see the respective theorems.
    }
\end{table}

\subsection{Quotients of model spaces}\label{section quotients}

The three spaces described above -- Euclidean, spherical, and hyperbolic -- are the only {\em simply connected} spaces of constant (sectional) curvature. They conveniently admit exact formulas for intrinsic distances and are the most frequent targets for data embeddings. However, at times when data exhibits periodic structure -- for example, when dates should be treated as periodic in one or more directions to discern time-of-day or seasonal effects -- it is advisable to consider suitable quotients, such as cylinders or (flat) tori. Similarly, if cosine similarity/spherical distance is used to assess similarity between document statements and user queries, it may be important to return also documents asserting the opposite statement to a user's conjecture. Mathematically, this corresponds to operating on the quotient space $S^d/\sim$ where $x\sim z$ if and only if $z\in\{-x,x\}$ (the real projective space $\mathbb{RP}^d$).

\begin{figure}
\centering
\includegraphics[width=0.15\linewidth]{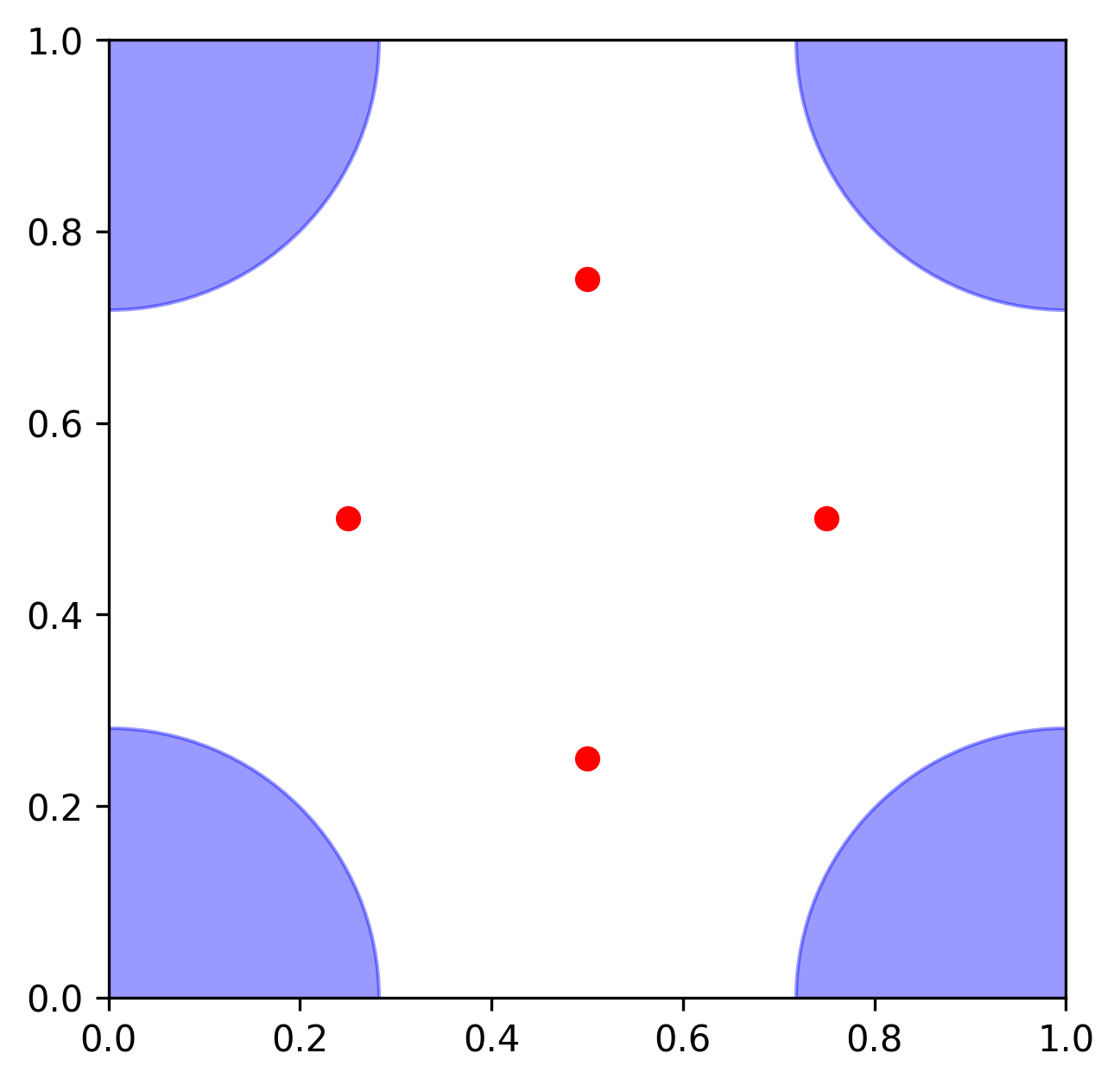}
\includegraphics[width=0.15\linewidth]{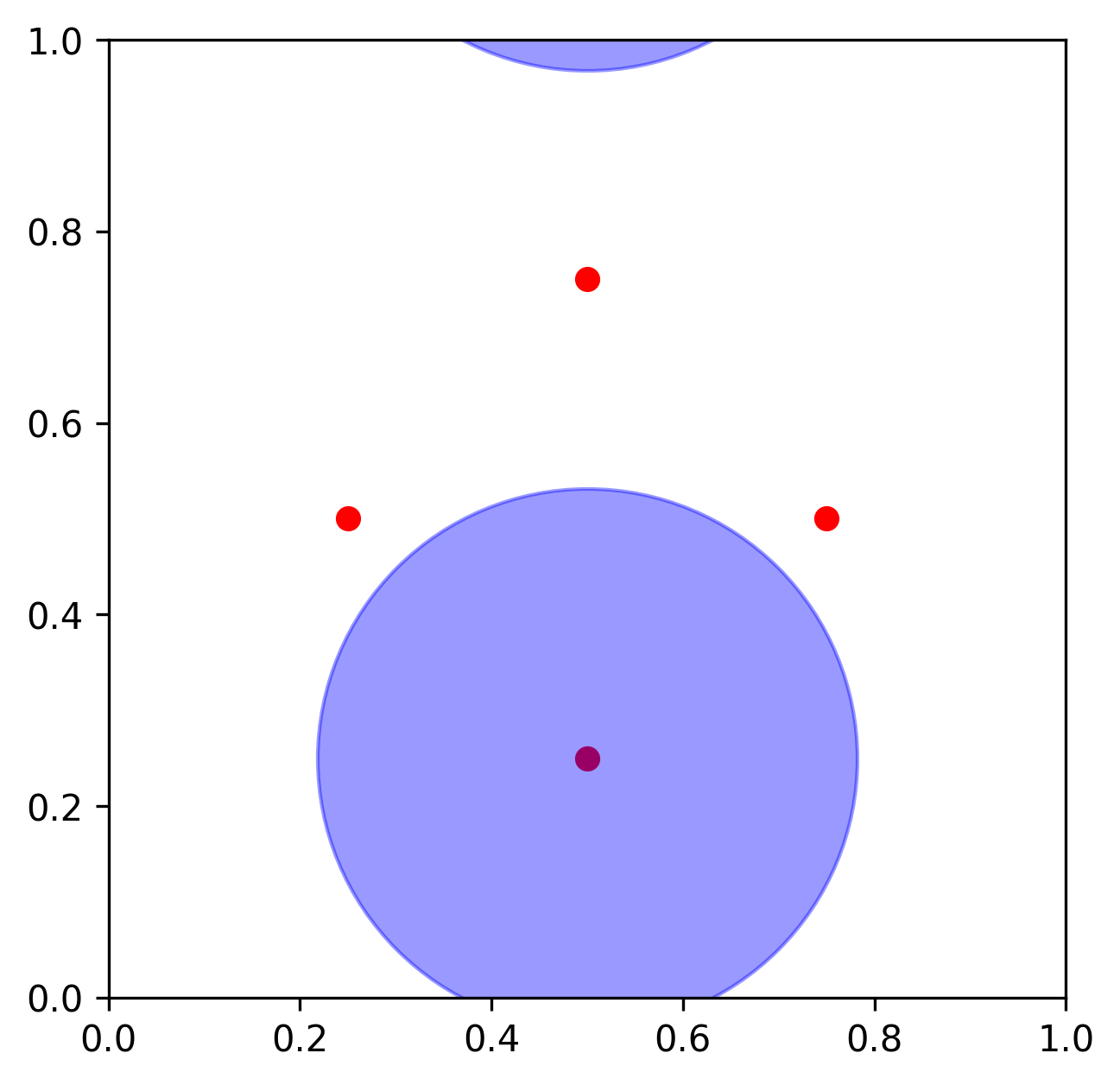}
\includegraphics[width=0.15\linewidth]{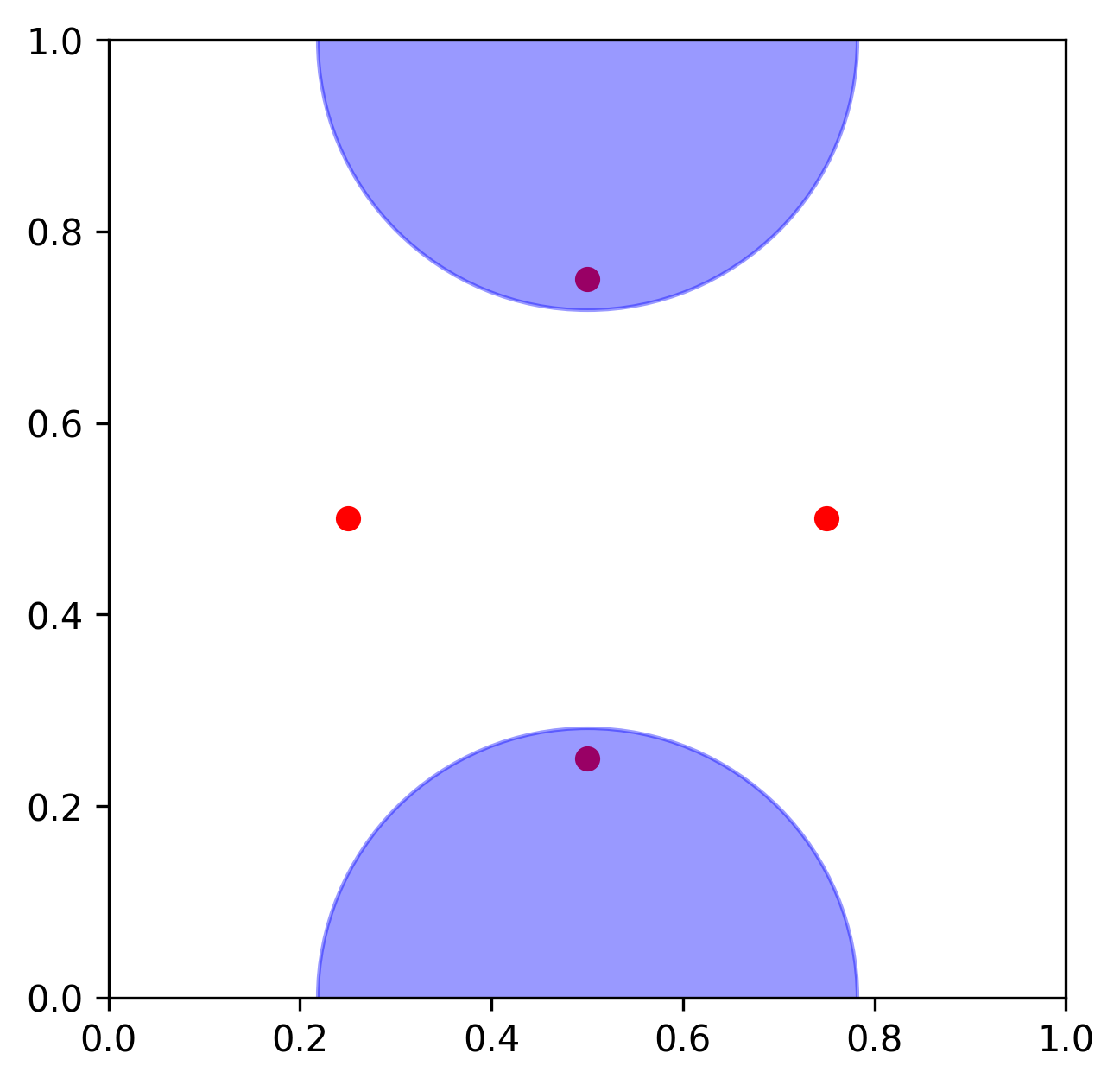}
\includegraphics[width=0.15\linewidth]{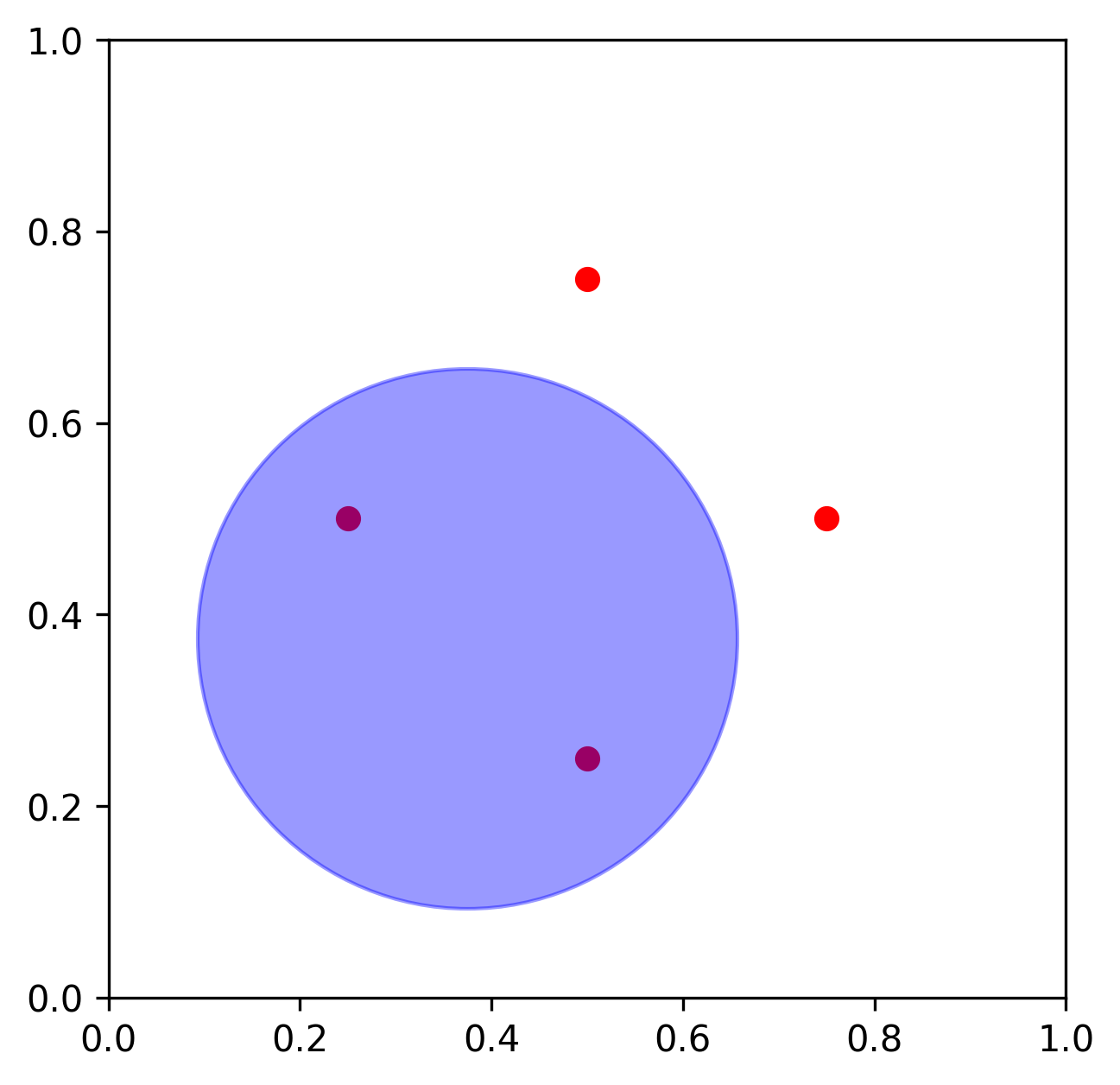}
\includegraphics[width=0.15\linewidth]{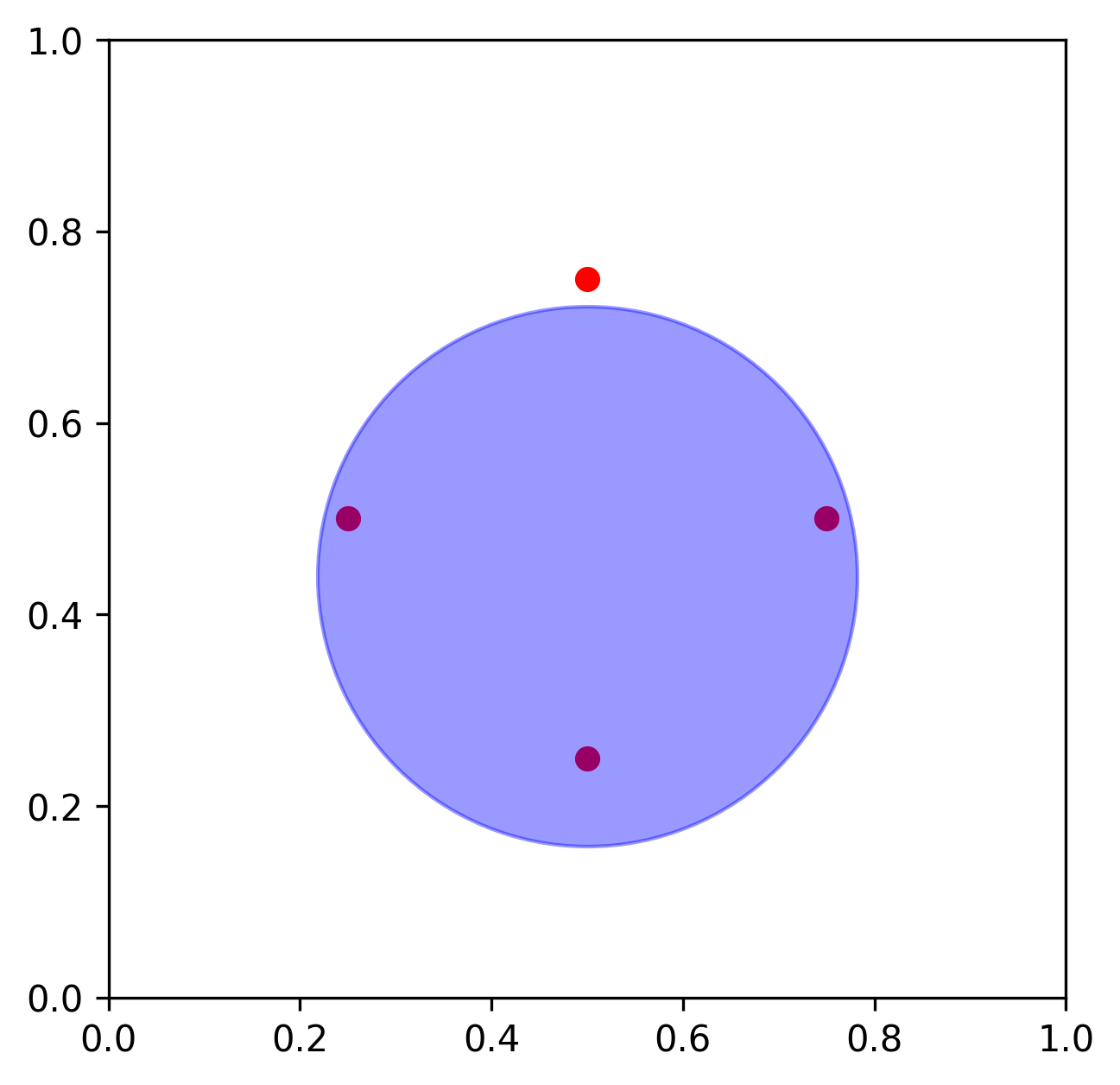}
\includegraphics[width=0.15\linewidth]{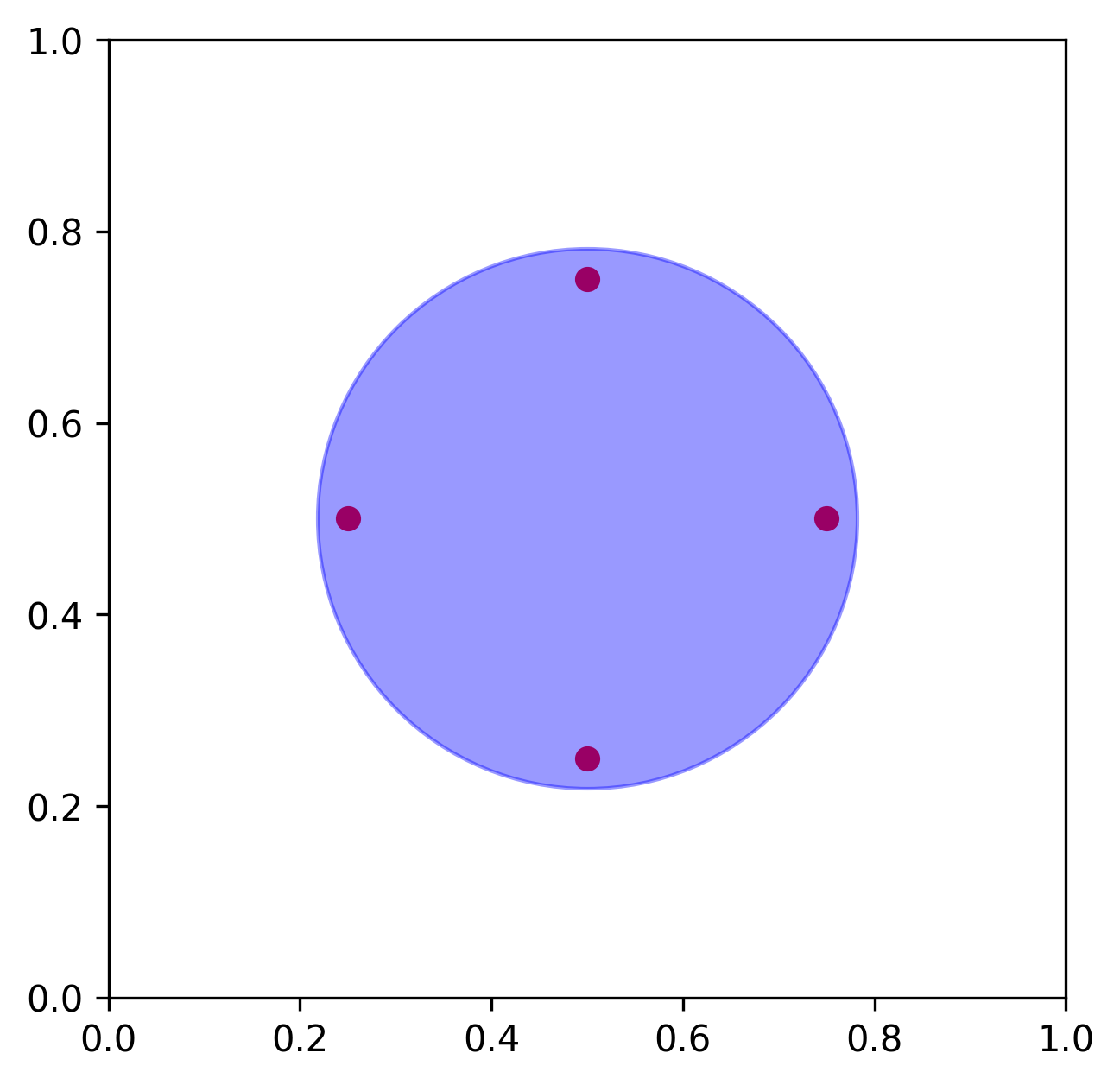}

\caption{\label{figure flat torus}
After the plane $\R^2$ and the sphere $S^2$, the torus $\R^2/\Z^2$ is the simplest example of a Riemannian surface. Small metric balls in the flat torus behave like balls in the plane (VC-dimension three) while balls of radius $r\in (0.25, 0.35)$ shatter the four-point set $\{(0.5, 0.25)$, $(0.5, 0.75)$, $(0.25, 0.5)$, $(0.75, 0.5)\}$, i.e.\ their VC-dimension is at least four.\\ {\bf Left to right:} Empty set, one-point set, two different types of two-point sets, three-point set, four-point set. Due to symmetry, the ones presented are the only subsets of the set of points, up to reflection or rotation.}
\end{figure}

Metric balls in such quotient spaces may have higher VC-dimension than in the original space due to the change in topology -- see Figure \ref{figure flat torus} for a simple example. However, small balls behave as in constant curvature spaces.

\begin{theorem}\label{theorem quotient spaces}
Let $(M, g)$ be a complete $d$-dimensional Riemannian manifold of constant curvature with injectivity radius $\inj(M,g)>0$. If $r< \inj(M,g)/3$, then $\B_r, \B_r^\circ, \B_{(0, r)}, \B_{(0, r]}, \B^\circ_{(0, r)}$ and $\B^\circ_{(0, r]}$ have VC-dimension $d+1$.
\end{theorem}

Since the complete Riemannian manifolds of constant sectional curvature are precisely the quotients of one of the three model spaces by the action of a properly discontinuous subgroup of the isometry group (Killing-Hopf Theorem \cite[Corollary 11.13]{lee2018introduction}), the theorem in particular applies to the motivating examples. To the best of our knowledge, it is unknown how large the VC-dimension of all metric balls (or metric balls of a fixed large radius) in a complete Riemannian manifold of constant curvature may be, even for flat tori. Similarly, the VC-dimension of metric balls inside a Riemannian product between a Euclidean space $\R^{d_1}$, a sphere $S^{d_2}$ and a hyperbolic space $\mathbb H^{d_3}$ or e.g.\ two spheres $S^{d_1}, S^{d_2}$ remains unknown. We give an initial partial result for flat tori.

\begin{lemma}\label{lemma flat tori}
Let $\T^d = \R^d/ \Z^d$. Then
\[
5\left\lfloor \frac d2\right\rfloor \leq \vcdim(\B)  \leq \max\big\{4(d+1)\,\log_2(4(d+1)),\ 110\big\}.
\]
\end{lemma}

The proof of Lemma \ref{lemma flat tori} exploits the product structure $\T^{2k} = (\T^2)^k$ of the torus and does not generalize to balls of fixed radius. We conjecture that a similar proof yields $\vcdim(\B) = O(d^2)$ on any torus $\R^d/\Gamma$, i.e.\ for any lattice $\Gamma\subset \R^d$. The difference between the two bounds is that for the lattice $\Z^d$, there are $d$ `lines' where a closest point representative may `jump' if $x$ crosses the line while there are $2^{d-1}(2^d-1)$ potential such lines in general grids associated to pairs of the vertices of a fundamental domain.

\subsection{Ambient dimension and intrinsic data dimension}\label{section intrinsic dimension}

All bounds above were data agnostic and thus depended on the ambient space dimension $d$. Often, realistic data is expected to lie on or near a low- or moderate-dimensional subset (a linear subspace or smooth sub-manifold) of a much higher-dimensional ambient space. It is a natural question whether the `effective hypothesis class' for such applications is smaller than suggested by the ambient space dimension. 

For instance, the restriction of the Euclidean norm to a $k$-dimensional subspace $V$ of $\R^d$ is an inner product norm on $V$ and thus isometric to the Euclidean space $\R^k$. Thus, if the probability distribution of real data is known to be supported on a $k$-dimensional affine subspace of $\R^d$, we are `effectively' working with a hypothesis class of Euclidean balls on a $k$-dimensional space and the $k$-dimensional VC-bounds govern generalization -- see Section \ref{section generalization}. Note that, by choosing the center of a metric ball $x\notin V$, we generate an `effective hypothesis class' of $k$-dimensional balls $\B_{(0, r]}$ of varying radius from the hypothesis class $\B_r$ of $d$-dimensional balls of fixed radius (and similarly for other settings).

\begin{corollary}\label{corollary vc dimension subspace}
\begin{enumerate} 
\item Let $V \subseteq \R^d$ be an affine subspace of dimension $k$ and $\mathcal X\subseteq V$ a subset of cardinality $k+2$. Consider the hypothesis classes $\B, \B^\circ$ {\em with respect to the Euclidean distance on the ambient space} $\R^d$. Neither class shatters $\mathcal X$. 

\item Consider a set $\mathcal X$ of $k+2$ points on the sphere $S^{k} = \{x\in\R^d : \|x\|_2 = 1\text{ and } x_{k+2} = \dots = x_d = 0\}$ in $\R^d$. Consider the hypothesis class $\B_r$ with respect to the Euclidean distance on the ambient space $\R^d$. 

\begin{itemize}
\item $\mathcal X$ cannot be shattered by $\B_r$ or $\B_r^\circ$ if $r< 1$. 
\item There are sets of $k+2$ points on $S^k$ which can be shattered by $\B_r^\circ$ if $r>1$ or by $\B_r$ if $r\geq 1$.
\item Sets of cardinality $k+2$ {\em cannot} be shattered by the restricted hypothesis classes
\[
\widetilde \B_r^\circ = \{ B_r(x) : x\in S^k\}, \qquad \widetilde \B_r = \{ \overline{B_r(x)} : x\in S^k\}
\]
irrespective of the radius $r$, 
where the balls are taken with respect to the Euclidean distance on the ambient space $\R^{d}$, but the centers of the balls are restricted to lying on the sphere $S^k$.
\end{itemize}
Since $S^k$ is contained in a $(k+1)$-dimensional linear subspace of $\R^d$, we immediately conclude that no subset of cardinality $k+3$ can be shattered.
\end{enumerate}
\end{corollary}

In general, it cannot be expected that real data lies {\em exactly} on a low-dimensional linear space, so it is natural to question whether the result is stable under small perturbations. Unfortunately, this is not the case. 

\begin{example}\label{example 2d graph}
Let $0<\eps \leq 1/\sqrt5$ and choose $\beta = \sqrt{1-\eps^2}$. Let 
\[
f_\eps(x) = \frac\eps{1-\eps^2} \,(x_1^2 - x_2^2).
\]
Then the hypothesis class $\B_1$ shatters the four-point set
\[
x_1 = (\beta, 0, \eps), \qquad x_2 = (-\beta, 0,\eps), \qquad x_3 = (0, \beta, -\eps), \qquad x_4 = (0, -\beta, -\eps).
\]
on the almost flat graph $\Gamma_\eps = \{(x, f_\eps(x)): x\in\R^2\}$.
To see that the set is shattered, one may consider points like the origin (all points), $\pm e_i$ for $i=1,2$ (one-point sets), $(0,0,\pm\eps)$ and $(\pm \beta/2, \pm \beta/2, 0)$ (two-point sets), and $(0, \pm \eps, \eps)$, $(\pm \eps, 0, -\eps)$ (three-point sets). For details, see Appendix \ref{appendix intrinsic vs ambient dimension}.
\end{example}

Naturally, as the ambient space is three-dimensional, no set of five or more points can be shattered by Euclidean balls.
With a slightly weaker guarantee in the general case, we show that a bound on the number of shattered points purely in terms of the intrinsic data dimension and independent of the co-dimension {\em cannot} be derived.

\begin{theorem}\label{theorem ambient dimension graph}
Let $k\leq d\in \mathbb N$, $\eps>0$ and $V = \R^k\times \{0_{d-k}\}\subseteq \R^d$. There exist $n = \max\{k,\ d-k\}$ points $z_i$ in $\R^d$ such that 
\begin{enumerate}
    \item $\|z_i\| = 1$ and $\dist(z_i, V) <\eps$ for all $i=1,\dots, n$, and
    \item $\{z_1, \dots, z_n\}$ is shattered by $\B_1$.
\end{enumerate}
The centers of shattering balls may be chosen in $B_{\sqrt\eps}(0)$.

If $k$ is so large (or $n\leq \max\{k, d-k\}$ is so small) that there are $n$ points $x_1,\dots, x_n\in S^{k-1}\subset \R^k$ such that $\|x_i -x_j\| \geq \delta\in (0,1]$ for all $i\neq j$, the points $z_i$ can be chosen to lie on the graph of a smooth function $f_{\eps,\delta} :V \to V^\bot$ such that the tensor of $m$-th derivatives of $f_{\eps,\delta}$ satisfies $\|D^mf_{\eps,\delta}\|_{L^\infty}\leq C_m\, (1-\eps^2)^{-m/2}\,\delta^{-m}\eps$ where $C_m$ depends on $m, d$ and $k$, but not $\eps$ or $n$. There are at least $\max\left\{2k,\ \sqrt{\frac 8\pi} \exp\left(\frac{(k-2)(2-\delta^2)^2}8\right) \right\}$ such points. 
\end{theorem}

The condition $\|z_i\|=1$ is selected along with $\dist(z_i,V)<\eps$ to ensure that $z_i$ truly appears $k$-dimensional at any length scale rather than a point cloud in a small sphere around the origin.

Thus, in a noisy setting or manifold setting, it is impossible to derive VC bounds on the `effective hypothesis class' derived from ambient space Euclidean balls which only depend on the intrinsic dimension of data and not the ambient space dimension, at least if the centers of balls are {\em not} required to lie exactly on the graph surface. Whether or not a natural barrier exists for a class of balls centered on the graph itself in the noiseless manifold setting remains open.

\subsection{General Riemannian manifolds, conformal metrics and data graph approximations}\label{section riemannian}

It is a natural question whether bounds depending only on the intrinsic dimensionality of data can be obtained using balls with respect to intrinsic distances instead of ambient space distances (or data-based estimates of intrinsic distances).
As a counterexample, we show that in general Riemannian manifolds, topological dimension and VC-dimension of metric balls may decouple entirely.

\begin{theorem}\label{theorem no vc bound}
\begin{itemize}
\item Let $n\in \mathbb N$. Then there exists a compact smooth two-dimensional Riemannian manifold $(M,g)$ such that the hypothesis classes $\B_1, \B_1^\circ$ have VC-dimension at least $n$.

\item Let $(M,g)$ be a Riemannian manifold of dimension $d\geq 3$ and $n\in \mathbb N$. Then there exists a Riemannian metric $\tilde g$ on $M$ which is conformally equivalent to $g$ such that the hypothesis classes $\B_1, \B_1^\circ$ have VC-dimension at least $n$.

\item If $M$ is a non-compact manifold of dimension $d\geq 3$, then for any Riemannian metric $g$ on $M$ there exists a Riemannian metric $\tilde g$ on $M$ which is conformally equivalent to $g$ such that the hypothesis classes $\B_1, \B_1^\circ$ have {\em infinite} VC-dimension.
\end{itemize}
\end{theorem}

If $(M,g)$ is complete, also the conformal metrics $\tilde g$ are complete since the conformal factor is constant outside a compact region of $M$. In particular, $g$ could be replaced by a conformally equivalent complete metric before modification to guarantee that $(M, \tilde g)$ is complete.

Conformal metrics are popular in data science with conformal factors which depend on the density of a data distribution on $M$.
These `Fermat metrics' allow for cheaper travel through higher density regions of the data domain \cite{sajama2005estimating}. Specifically, if $\rho$ is a (strictly positive) probability density on a $d$-dimensional Riemannian manifold $(M,g)$, the associated $p$-th order Fermat metric is associated with the re-weighted metric tensor $g_p = \rho^{-2(p-1)/d}g$ for $p\in[1, \infty)$. This coincides with the original geometry for constant $\rho$ or $p=1$, but density-dependent metrics often lead to improved performance on a variety of data science tasks \cite{mckenzie2019power, groisman2022nonhomogeneous, little2022balancing, chazal2023choosing, trillos2024fermat, manousidaki2024clustering}. In Theorem \ref{theorem no vc bound}, we observe that even if $(M,g)$ is a well-behaved model space, metric balls on $M$ with respect to the weighted Fermat metric may behave wildly depending on $\rho$ and $p$.

In the proof of Theorem \ref{theorem no vc bound}, we construct conformal factors which make travel along a specified system of curves cheap, akin to `building highways through a swamp'. Thus we require high data density, but only on a set of co-dimension $\geq 2$. This is easily realized as $\rho^{-2(p-1)/d}$ with a probability density $\rho$, assuming that $p>1$. While the construction of counterexamples is often artificial, real datasets may possess such structural properties e.g.\ if data is not sampled independently but from time series progressing along curves.

It remains open whether bounds on the (Riemann or Ricci) curvature tensor of a Riemannian manifold or on the covariant derivative of a curvature tensor can be used to bound the VC-dimension of metric ball classifiers at some fixed (finite and positive) length scale.

\subsection{VC-dimension of unions of metric balls}

Suppose $k\geq 1$. Denote by 
\[
\B^{k\cup} = \left\{ \bigcup_{i=1}^k B_i : B_1, ..., B_k \in \B \right\}
\]
the hypothesis class of classifiers associated to $k$-fold unions of sets in $\B$. By \cite[Lemma 3.2.3]{blumer1989learnability}, we have
\begin{equation}\label{eq vc-dimension of unions}
    \vcdim(\B^{k\cup}) \leq 2k\log(3k)\cdot \vcdim(\B).
\end{equation}
The logarithmic dependence on $k$ in \eqref{eq vc-dimension of unions} is unavoidable: If $d \geq 3$, the VC-dimension of a $k$-fold union of half-spaces satisfies
\begin{align*}
    \vcdim(\H^{k\cup}) = \Omega \big( (d+1) \cdot k\log(k) \big)
\end{align*}
by \cite[Theorem 1]{csikos2019tight}. While metric ball classifiers and half-space classifiers are distinct, the former are {\em more} expressive: {\em Every set which can be shattered by half-spaces can also be shattered by balls, and every set which can be shattered by a $k$-fold union of half-spaces can be shattered by a $k$-fold union of balls.} Given a set, any fixed but sufficiently large radius suffices. 

\begin{corollary}\label{corollary vc-dimension union lower bound}
In $\R^d$ with the Euclidean norm
\[
\vcdim(\B^{k\cup}) \geq \vcdim(\B_r^{k\cup}) \geq \vcdim (\H^{k\cup}) =  \Omega \big( (d+1) \cdot k\log(k) \big).
\]
\end{corollary}

The corollary holds for collections both of closed and open balls. A bound similar to \eqref{eq vc-dimension of unions} holds for the more general hypothesis classes derived from boosting \cite[lemma 10.3]{shalev2014understanding}, which aggregate simple geometric classifiers into complex predictors.

\section{Generalization of Metric Ball Classifiers}\label{section generalization}

\subsection{VC-dimension based bounds}
As a consequence of the bounds in the previous section, we know that metric ball classifiers and classifiers associated to the union of metric balls generalize well in `good' Riemannian manifolds. We present results in a `coverage estimation' setting for measures, but note that the classification setting follows easily if one considers $n_+$ points sampled from a measure $\mu_+$ of data with positive label and $n_-$ points sampled from a measure $\mu_-$ of data with negative label.

\begin{theorem} \label{thm:probability_bounds_using_VCdim}
    Suppose there are $n$ points $x_1, ..., x_n$ drawn i.i.d.\ from domain $X$ with respect to some probability measure $\mu$. Let $\hat \mu_n = \frac{1}{n}\sum_{i=1}^n \delta_{x_i}$ denote the empirical measure.
    
    For a given metric on $X$, if $\B$ has finite VC-dimension $\vcdim(\B)$ and $k\in \N$, then there exists a universal constant $C$ such that  
    \begin{align*}
       % \P\left(\sup_{x\in X} \big|\mu(B_r(x)) - \hat\mu_n(B_r(x))\big| \geq \epsilon \right) &\leq \frac{1}{\epsilon} \cdot C \sqrt{\frac{V(\cal H)}{n}}, \\
        \P\left(\sup_{z_1, \dots, z_k\in X} \left|\mu\left(\bigcup_{j=1}^kB_r(z_j)\right) - \hat\mu_n\left(\bigcup_{j=1}^kB_r(z_j)\right)\right| \geq \epsilon \right) &\leq \frac{1}{\epsilon} \cdot C \sqrt{\frac{2\vcdim(\cal H) \cdot k\log(3k)}{n}} .
    \end{align*}
\end{theorem}

Conversely, the Fundamental Theorem of PAC learning \cite[Theorem 6.8]{shalev2014understanding} and Theorem \ref{theorem no vc bound} imply in combination that there may be learning tasks which are not PAC learnable (or agnostic PAC learnable) in the class of metric ball classifiers when $(M,g)$ is a generic Riemannian manifold of dimension $d\geq 3$. Even if $(M,g)$ is a `nice' manifold, there exists a conformally equivalent metric $\tilde g$ which is pathological.

\subsection{Covering-based bounds: Intrinsic balls}\label{section covering intrinsic}
Using an approach which avoids VC-dimension entirely and bounds generalization directly by covering arguments under stronger assumptions on the data distribution, we can prove the following.

\begin{theorem}\label{theorem generalization covering}
Let $(X,d)$ be a separable metric space and $\mu, \nu$ probability measures on (the Borel $\sigma$-algebra of) $X$. Let $r>0$ and assume that there exist $c, C, \delta_0, d >0$ such that
\begin{enumerate}
\item $\nu(B_\delta(x))\geq c\,\delta^d$ for all $\delta<\delta_0$ and $x\in X$.
\item $\mu\big(B_{r+\delta}(x) \setminus B_{r-\delta}(x)\big)\leq C\delta$ for all $\delta<\delta_0$ and $x\in X$.
\end{enumerate}
Assume that $k\in\N$ and $\eps< 4Ck\delta_0$.
If $x_1, \dots, x_n$ are independent samples from $\mu$ and $\hat \mu_n := \frac1n \sum_{i=1}^n\delta_{x_i}$ is their empirical measure, then
\begin{equation}\label{eq generalization bound}
\P\left(\sup_{z_1,\dots,z_k\in X} \left|\mu\left(\bigcup_{i=1}^kB_r(z_i)\right) - \hat\mu_n\left(\bigcup_{i=1}^kB_r(z_i)\right)\right| \geq \epsilon \right)
\leq 6\,\frac{\sqrt{\log 2 + k\log(1/c) + dk\,\log(8Ck/\eps)}}{\eps\,\sqrt n}
\end{equation}
where $\P = \mu^{\otimes n}$ denotes the joint distribution of $x_1,\dots, x_n$.
\end{theorem}

The assumption on $\nu$ is often referred to as a lower Ahlfors regularity condition. We note that the measures $\mu, \nu$ may be distinct.
The theorem has an immediate application for manifolds with suitably bounded curvatures, at least on `small enough' scales.

\begin{corollary}\label{corollary ricci bound}
Let $(M,g)$ be a compact $d$-dimensional Riemannian $C^\infty$-manifold with the natural $d$-dimensional (Hausdorff) measure $\H^d(M) = v$. Assume that the Ricci curvatures of $(M,g)$ satisfy $\Ric \geq (d-1)\,K g$ for some $K\in\mathbb R$ in the sense of quadratic forms.

Let $\mu$ be a probability distribution on $M$ which has a density $\rho$ with respect to $\H^d$ such that $0\leq \rho \leq A$, and assume that $x_1,\dots, x_n$ are random points in $M$, sampled independently with distribution $\mu$. Denote by $\hat\mu_n$ their empirical distribution. 

Let $0<r<\inj(M,g)$. Then there exists $\eps_0>0$ depending on the geometry of $(M,g)$ such that \eqref{eq generalization bound} holds for all $\eps<\eps_0$ with 
\[
c = \frac{\omega_d}{2v}, \qquad C = 2A\,d\omega_d \max_{s\in[0,2r]}\phi_K(s)^{d-1} \qquad\text{where } \phi_{K}(s) = 
\begin{cases}
\sin(\sqrt{K}s)/\sqrt{K} & K>0\\ s&K=0\\\sinh(\sqrt{|K|}s)/\sqrt{|K|} & K<0.
\end{cases}
\]
\end{corollary}

The result applies in particular to Riemannian sub-manifolds of $\R^D$ for $D>d$ with their intrinsic geometry. We believe that the smoothness requirements can be relaxed to $C^k$ with $k\in \mathbb N$ sufficiently large, but are not aware of a reference in which smoothness conditions are carefully analyzed to this end and leave the question open for future work.

While we use the fact that the sectional curvatures of the compact Riemannian manifold $(M,g)$ are bounded from above in the proof of Corollary \ref{corollary ricci bound}, it only enters the statement implicitly as a condition on $\delta_0$. By contrast, the lower Ricci curvature bound enters in the constant $C$ for the growth of spherical shells and cannot be disregarded by passing to a sufficiently small scale. Crucially, the Ricci curvature can be estimated from point clouds at least when they are sampled from the uniform distribution on a sufficiently smooth compact sub-manifold of $\R^D$ \cite{garcia2026continuum}.

Due to its covering-based nature, there exists a little slack in the proof of Theorem \ref{theorem generalization covering} in that the metric $d$ on the space $X$ does not have to be known precisely for the conclusion to be valid, but only to sufficiently high precision. This applies e.g.\ when the manifold $(M,g)$ is not given, but reconstructed as a metric space from a point cloud in a high-dimensional ambient Euclidean space $\R^D$.

\begin{theorem}\label{theorem graph approximation of sub-manifold}
Assume that $(M,g), K, A, \rho, \eps_0,r, \mu, \hat\mu_n$ are as in Corollary \ref{corollary ricci bound}. Assume random distance estimates $d_{ij}$ are given (potentially depending on the sample $x_1,\dots, x_n$) such that
\[
\max_{i,j=1,\dots,n} \big|d_{ij} - d_{(M,g)}(x_i,x_j)\big|<\alpha\eps/k\qquad \text{for } \alpha = \frac1{2C}
\]
with probability at least $1-\gamma$ and set
\[
B_i := \{j =1,\dots, n : d_{ij}<r\}.
\]
Then  
\[
\P\left(\sup_{\substack{I\subseteq \{1,\dots,n\}\\ |I| \leq k}} \left| \mu\left(\bigcup_{i\in I} B_r(x_i)\right) - \frac{\left|\bigcup_{i\in I}B_i\right|}n\right| \geq \eps\right) \leq \gamma + \frac1\eps\,\sqrt{\frac{2k\,\log(2en)}{n}}.
\]
for all $\eps<\eps_0$ and $k=1,\dots,n$. If $\eps$ is taken slightly smaller, then also the estimate
\[
\P\left(\sup_{\substack{I\subseteq \{1,\dots,n\}\\ |I| \leq k}} \left|\mu\left(\bigcup_{i=1}^kB_r(x_i)\right) - \frac{\left|\bigcup_{i\in I}B_i\right|}n\right| \geq \epsilon \right)
\leq \gamma + 24\,\frac{\sqrt{\log 2 + k\log(1/c) + dk\,\log(16\,Ck/\eps)}}{\eps\,\sqrt n}
\]
holds for $k=1,\dots, n$ with the same constants $c,\ C$ as in Corollary \ref{corollary ricci bound}.
\end{theorem}

The first estimate is sharper for small $n$ and seemingly independent of the dimension of $M$ (although the convergence of empirical estimates to true distances depends on it, see below). It is obtained by bounding the generalization at the $n$ points $x_i$ directly without appealing to the geometry of $M$. The second estimate involves larger geometric constants derived from the generalization bound in Corollary \ref{corollary ricci bound} at {\em all} points $x\in M$, but it yields the faster asymptotic rate of convergence $n^{-1/2}$ as $n\to\infty$.

\begin{remark}\label{remark intrinsic distances}
Intrinsic geometric distances on sub-manifolds of Euclidean spaces can be estimated from point clouds in various settings.

\begin{enumerate}
\item A point cloud in $\R^d$ can be equipped with a graph structure whose vertices are the points. Two points $x_i, x_j$ are connected by an edge if $\|x_i-x_j\|<\eps$ for a threshold value $\eps $ (a geometric neighborhood or $\eps$-neighborhood graph) or if $x_i$ is among the $k$ nearest points to $x_j$ (a $k$-nearest neighbor or $k$NN digraph). Typically, the kNN digraph is symmetrized to an undirected graph.

Distances in this graph with edge weights $e_{ij} = \|x_i-x_j\|$ can be computed for instance by Dijkstra's algorithm. On a fully connected graph ($\eps = +\infty$), these distances coincide with the Euclidean distances by the triangle inequality, while they converge to intrinsic `data manifold' distances in $\eps_n$-neighborhood graphs where $\eps= \eps_n$ decreases slowly with the number of data points \cite{bernstein2000graph, alamgir2012shortest, arias2019unconstrained}. The decrease must occur in order to force paths to follow the manifold and must be slow in order for the graph to resemble $M$ at the resolution given by the random point samples.

If the same point cloud is used to obtain $\hat\mu_n$ and estimate intrinsic distances in this fashion, one can argue from \cite[Propositions 1 and 2]{arias2019unconstrained} that the approximation accuracy scales as 
\[
\qquad d_{ij} - \dist_{(M,g)}(x_i, x_j) = O\left(\max\left\{\eps_n^2, \frac{\delta_n}{\eps_n}\right\}\right)\qquad\text{where } \delta_n := \max_{x\in M} \min_{i=1,\dots, n}\|x-x_i\|
\]
is the filling length scale and $\delta_n \sim (n/\log n)^{-1/d}$. This is minimized when $\eps_n \sim \delta_n^{1/3}$ at a rate of convergence $(\log n / n)^{2/3d}$. To the best of our knowledge, a precise analysis over the entire manifold is not available.

In general, larger point clouds may be used to approximate manifold distances if all $\mu$ is the distribution of only one class of points on $M$.

\item In the same setting, one may choose edge weights $e_{ij} = \|x_i-x_j\|^p$ for $p>1$ and recover `Fermat metrics' \cite{sajama2005estimating, mckenzie2019power, groisman2022nonhomogeneous, little2022balancing, trillos2024fermat} which allow for cheaper travel along the data manifold in high density regions of the data distribution. If $p>1$, short jumps are automatically preferred and even Dijkstra graph distances on the fully connected graph recover the `Fermat metric' (after suitable rescaling).

In \cite[Theorem 4.6]{little2022balancing}, the authors demonstrate that -- under strong additional geometric assumptions -- that the distance between the (rescaled) graph Fermat metric and the Fermat distance on the manifold is no worse than $O(n^{-1/2d}\log n)$ on average. The result holds for flat domains, i.e.\ the intrinsic and ambient dimension coincide, but it trivially generalizes for flat domains within a high-dimensional ambient space.

\item For a sub-manifold $M$ of $\R^d$, \cite{memoli2005distance} demonstrate that distances within $M$ can be approximated by distances within a tubular neighborhood $M_\delta = \{x\in\R^d : \dist(x,M)<\delta\}$ (estimated from a point cloud), which can be approximated using a Cartesian grid. Cartesian grids allow for more efficient discretizations than arbitrary point clouds, but due to the dependence on the ambient dimension, the technique is restricted to sub-manifolds of low-dimensional spaces.

\item In \cite{calder2022hamilton}, the authors study an approximation of intrinsic distances based on the observation that distance functions solve the eikonal equation $\|\nabla u\| = 1$ (in the viscosity sense). For data sampled from a probability distribution on a domain $\Omega\subseteq\R^d$ and under regularity assumptions on domain and density (as well as uniform positivity and boundedness of the Lebesgue density), the authors prove the convergence of solutions to `graph-eikonal' equations to solutions to eikonal type equations in the domain $\Omega$. The authors prove results on fully connected graphs, but sketch in \cite[Remark 27]{calder2022hamilton} why the results are expected to hold on $k$-nearest neighbor graphs. This approach enjoys greater stability to large errors in the length of graph edges than e.g.\ shortest path distances in data graphs. Results on sub-manifolds rather than flat domains are not available, but expected to hold.
\end{enumerate}
\end{remark}

\subsection{Covering-based bounds: Ambient space balls}

In view of Theorem \ref{theorem ambient dimension graph}, the hardness of learning metric ball classifiers on moderate-dimensional sub-manifolds of high-dimensional spaces generally depends not only on the intrinsic dimensionality of the data but also on that of the ambient space, even if the sub-manifold is merely the graph of a function. Under additional assumptions on the data distribution however and for metric balls centered on the data manifold itself, it can be relaxed in the same way as for intrinsic balls in Section \ref{section covering intrinsic}.

\begin{theorem}\label{theorem generalization ambient balls}
Let $R, \delta>0$, $L\in[0,1)$, $m\in \{1,\dots, d\}$ and $M\subseteq \R^d$ such that $v:= \H^m(M) < \infty$. Assume that for all $x_0\in M$, there exists a rigid motion $\phi(x) = O(x-x_0)$ and a function $f:\R^m \to \R^{d-m}$ such that the following are true:
\begin{enumerate}
\item $f(0) = 0_{d-m}$ and $f$ is Lipschitz-continuous with Lipschitz constant $L$.
\item $\phi(M\cap B_R(x_0)) = \Gamma_f \cap B_R(0)$ where $\Gamma_f = \{(x, f(x)): x\in B_R(0)\subset \R^m\}$.
\end{enumerate}
Assume that $\mu$ is a probability distribution on $M$ which has a bounded density $\rho:M\to[0,A]$ with respect to the $m$-dimensional Hausdorff measure on $M$. Assume that $x_1,\dots, x_n\in M$ are independent random samples from $\mu$ and denote $\hat\mu_n = \frac1n\sum_{i=1}^n \delta_{x_i}$. Then, if $r$ is so small that $r+\delta<R$, the bound \eqref{eq generalization bound} holds with
\[
c = \frac{\omega_m}{v\,(1+L^2)^{m/2}}, \qquad C = 2m\omega_mR^{m-1} \,A\,\frac{(1+L^2)^{m/2}}{1-L}.
\]
\end{theorem}

We note that the generalization bounds of Theorem \ref{theorem generalization ambient balls} are independent of the dimension of the ambient space, but derived for balls with center on the data manifold $M$. They apply in particular to balls centered at data points which are sampled from a probability distribution on $M$, an important case for applications. However, they are not fully comparable to the lower bound of Theorem \ref{theorem ambient dimension graph}.

The Lipschitz constant $L$ of $f$ in Theorem \ref{theorem generalization ambient balls} has to be small. With slightly higher regularity assumptions on $M$, this is not an issue: If $M$ is a $C^{1,\alpha}$-sub-manifold of $\R^d$, we can write it locally as a graph over its tangent space with $Df(0) = 0$, so
\begin{align*}
\|f(x) - f(y)\| &\leq \int_0^1 \|Df(tx+(1-t)y)\|\,\|x-y\|\dt \\
    &\leq \|x-y\|\int_0^1 [Df]_{C^{0,\alpha}}\,\|tx+(1-t)y\|^\alpha\dt\\
    &\leq \max\{\|x\|, \|y\|\}^\alpha\,[Df]_{C^{0,\alpha}} \,\|x-y\|
    % &\leq [Df]_{Lip}\|x-y\|\sqrt{\int_0^1 t^2\|x\|^2 + 2t(1-t)\langle x,y\rangle + (1-t)^2\|y\|^2\dt}\\
    % &= [Df]_{Lip}\|x-y\|\sqrt{\frac12 \|x\|^2 - \frac13\langle x,y\rangle + \frac12\|y\|^2\,}\\
    % &\leq [Df]_{Lip}\|x-y\| \sqrt{\frac13\,\|x\|^2 + \frac13\,\|y\|^2 + \frac16\,\|x-y\|^2}
\end{align*}
i.e.\ the Lipschitz constant of $f$ grows as $r^\alpha$ on disks of radius $r$. In particular, we can guarantee that the assumptions will be met if $R$ is chosen sufficiently small. 

We conjecture that a similar result can be obtained assuming that $M$ is locally $C^1$-close to a quadratic surface rather than comparable to a plane as in the proof of Theorem \ref{theorem generalization ambient balls}. This is implied for $C^{2,\alpha}$-manifolds with $D f(0) = 0_{(d-m)\times m}$ under uniform bound assumptions on $D^2f(0)$ and $[D^2f]_{C^{0,\alpha}}$.

\section{Applications}\label{section applications}

\subsection{Active Learning}

\begin{figure}
    \centering
    \includegraphics[width=0.48\linewidth]{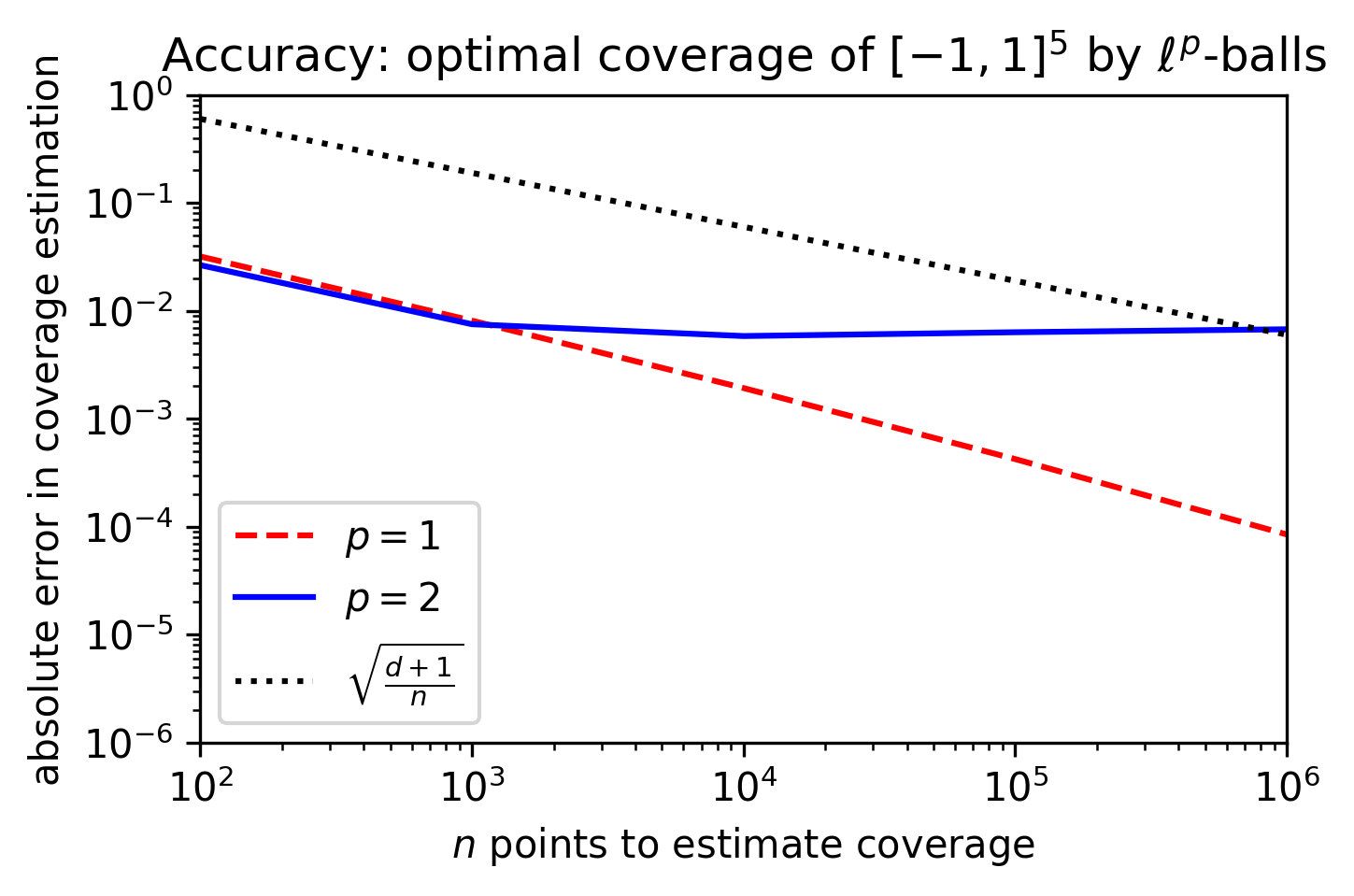}
    \includegraphics[width=0.48\linewidth]{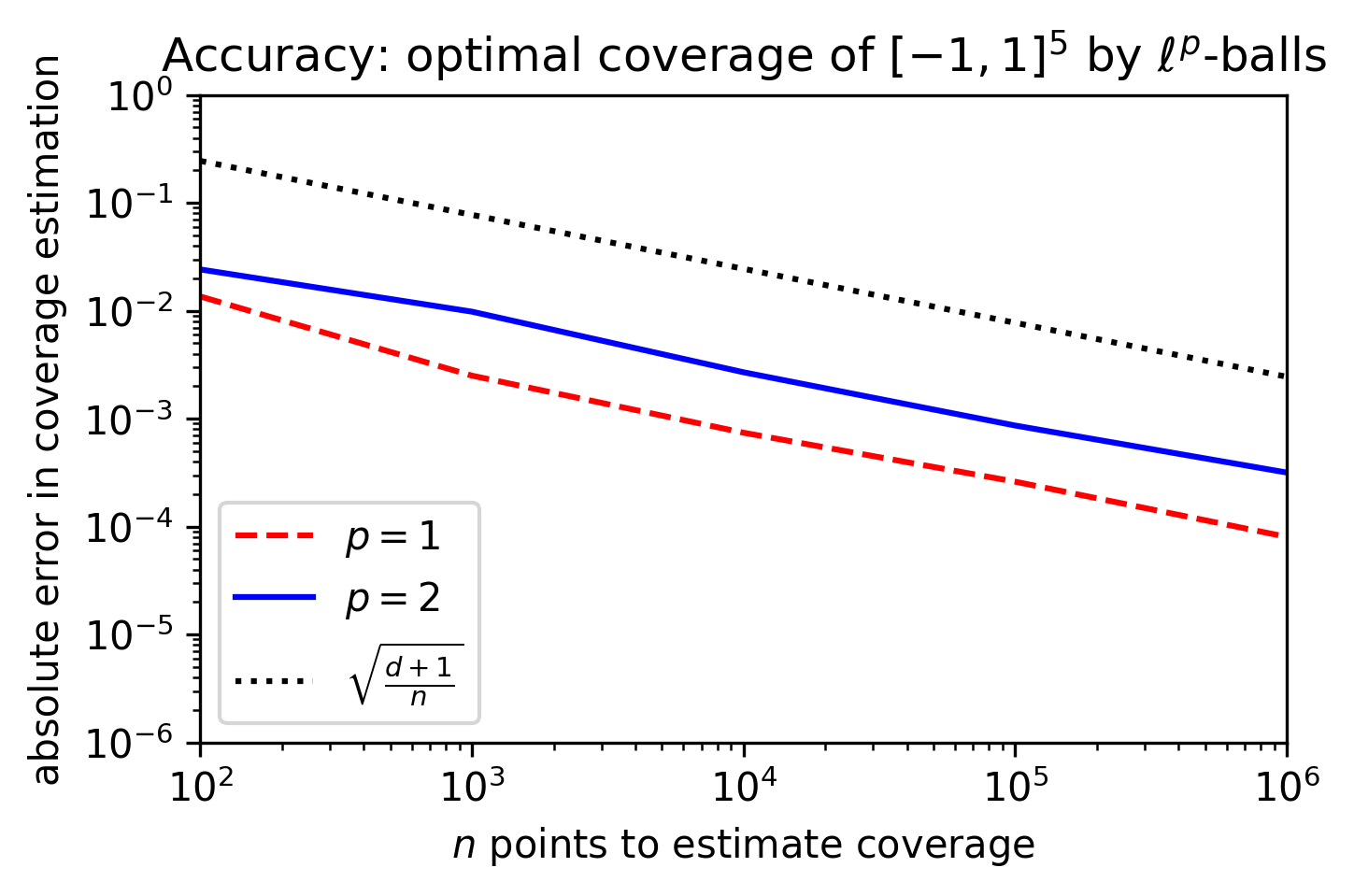}
\caption{\label{figure coverage by lp balls}
Given $x_1,\dots, x_n$ sampled uniformly from $[-1,1]^5$, we select the point $x_i$ which achieves the highest coverage of the empirical distribution, i.e.\ which maximizes $c_i:= \frac1n\sum_{j=1}^n 1_{\{\|x_i-x_j\|_p\leq 1\}}$ for $p\in\{1,2\}$ to estimate which fraction of $[-1,1]^d$ can be covered by an $\ell^p$-ball of radius $r=1$. We plot the discrepancy from the maximum coverage which is achieved by $B_1^p(0)$ for the uniform distribution on $[-1,1]^5$. Results are averaged over 100 runs.\\
To reduce computational complexity, we consider only $\min\{n, 500\}$ random indices (left plot) or $x_i$ such that $\|x_i\|_p \leq 1.1\cdot \min_j\|x_j\|_p$ (right plot). Despite the higher VC-dimension of the hypothesis class in $\ell^1$, the estimates are closer to the true coverage and as $n$ increases, the error from considering fewer centers dominates for $p=2$, but not $p=1$. As `spherical caps' carry more mass than caps at the vertices of the cross polytope, the coverage is more resilient towards a suboptimal choice of center. Effects due to using $x_i$ both for coverage estimation and as centers of balls for covering will be analyzed in a forthcoming article \cite{probcoverlimit}.
}
\end{figure}

Active learning algorithms aim to select points $x_{i_1},\dots, x_{i_k}$ from a much larger unlabeled dataset $x_1,\dots, x_n$ for annotation. To identify an informative subset, at least during the initial rounds of labeling, one aims to select a `representative' subset. Algorithms such as CoreSet \cite{sener2017active} and ProbCover \cite{yehuda2022active} select a set of points $x_1,\dots, x_k$ by

\begin{enumerate}
\item minimizing $r$ under the constraint that $\{x_1, \dots, x_n\} \subset \bigcup_{j=1}^k B_r(x_{i_j})$ or
\item maximizing the number of covered points $\Big|\{x_1,\dots, x_n\} \cap \bigcup_{j=1}^k B_r(x_{i_j})\Big|$ for given $r>0$.
\end{enumerate} 

We can cast the coverage problem as a binary classification problem in which all points have label 1, i.e.\ the goal is to cover the entire data space. The learning problem of ProbCover is not realizable unless $r$ is sufficiently large; while coverage for CoreSet is trivially realizable with a single ball of sufficiently large radius, the structural risk minimization problem of solving the problem with small radius is hard. In both cases, labeled data points serve both as centers for the balls covering the data set and to estimate coverage of the true data range. The more general generalization bounds presented in Section \ref{section generalization} are naturally cast in terms of coverage estimation. In Figure \ref{figure coverage by lp balls}, we illustrate the optimal coverage by $\ell^p$-balls of $[-1,1]^d$ as estimated by centers of a sub-sample of a dataset and its convergence to the true coverage.

While both algorithms were originally introduced using Euclidean distances, they rely only on the metric structure of data (and could, in fact, be generalized to non-metric notions of data similarity). In a companion article \cite{applicationpaper}, we explore ProbCover with Fermat metrics and find that it leads to an empirically more informative subset for data annotation.

\subsection{Metric balls and nearest neighbor classifiers}

Metric ball and $k$-nearest neighbor classifiers are two distinct classes of classifiers based on proximity in metric spaces. Despite obvious similarities -- multiple close points with one label raise the likelihood of that label in both cases -- there are clear differences. 

\subsubsection{Nearest neighbors}
Given a dataset $S = \{x_1,\dots, x_n\}$ of labeled points and a new point $x$, $k$-nearest neighbors (kNN) classifiers find the distance from $x$ to all points $x_i$, keeping track of the $k$ closest ones and predicting the label of the new point by majority vote. As new data is received, no training is necessary, but the computational complexity at deployment scales as $n d$ since all distances need to be computed. With pre-processing and storing data in suitable structures, the complexity can be reduced to $O(d\log n)$ by KD trees in low dimension and, at the cost of returning approximate nearest neighbors, to $O(dn^\rho)$ with $\rho<1$ by locality sensitive hashing.

The impact of a labeled point does not decay with distance: In general, the voting strength of a point is not weighted by inverse distance or similar. The classifiers treat both classes identically: Upon flipping the labels of known points, the labels of given points are also reversed (assuming that a tie break mechanism is also flipped or symmetric random to begin with).

\subsubsection{Unions of metric balls}

Metric ball and union-of-balls classifiers treat two classes differently: One is assigned the `interior domain' $B_r(x)$ or $\bigcup_{i=1}^k B_r(x_i)$ while the other is assigned its complement. They are potentially more light-weight than their kNN counterparts if $k\ll n$ since only the distance to the centers of the balls $x_1,\dots, x_k$ needs to be computed rather than to all points in $S$. A labeled point of the `interior type' has full control of the predicted label at a point up to a certain radius, even if points with the opposite label are nearby. On the other hand, its influence abruptly ends at the cut-off radius. 

Union-of-balls classifiers may require more extensive preparation to select centers and radii, unless simply all known points with a given label are selected as centers.

\subsubsection{Use cases}

Due to their fundamental asymmetry, union-of-balls classifiers are most naturally suited to problems with inherent asymmetry, such as a clear majority/minority divide. For instance, they naturally flag any newly received points which are sufficiently similar to known minority point candidates. In applications like fraud detection with a very pronounced gap between the number of legitimate and fraudulent transactions and a significant interest in avoiding false negatives, this describes a filter for potential candidates. In this case, $k$ could be chosen as the number of known fraudulent transactions in a given set.

\section*{Tool Use}

After the article was written by the human authors, {\tt Claude-Opus-5.5} was used to proofread the article and helped flag minor errors, both in terms of language/presentation and mathematical inconsistencies. 

Additionally, Claude was used to find examples of a shattered five-point set in $\T^2$ for Lemma \ref{lemma flat tori} and a five-point set in $\R^3$ shattered by $\ell^1$-balls for Theorem \ref{theorem vc normed vector spaces} as well as the example in Table \ref{table shattering fixed radius R3} of $\ell^1$-balls with fixed radius, presented in the appendix. The arguments exploiting the product structure of spaces for lower bounds in $\T^d$ and $\ell^1(\R^d)$ as well as the Sauer-Shelah based upper bound in both cases were designed by the human authors with Claude.

No text in the article was generated by automated tools. No results other than those listed above were obtained using artificial intelligence.

\section*{Acknowledgements}

The work of DD and SW was partly supported by grant NSF DMS-2424801. The work of EM was partly supported by grant NSF DMS-2512644.

\bibliographystyle{alpha}
\bibliography{bibliography.bib}

\newpage
\appendix
\section*{Appendix}
\startcontents[appendix]
\renewcommand{\baselinestretch}{1.3}\normalsize
\printcontents[appendix]{l}{1}
{\setcounter{tocdepth}{2}}
\renewcommand{\baselinestretch}{1}\normalsize

\section{VC-dimension: Constant Curvature Manifolds and Half-Spaces}\label{appendix constant curvature}

\subsection{Closed and open ball classifiers in metric spaces}

\begin{proof}[Proof of Lemma \ref{lemma closed vs open balls}]
We focus on the proof of the first statement as all others are analogous. Assume that $S = \{x_1, \dots, x_n\}$ is shattered by $\B_r$. Then for every $J \subseteq \{1,\dots, n\}$, there exists $x_J$ such that
\[
d(x_J, x_j) \leq r < d(x_J, x_k) \qquad\forall\ j \in J,\ k\notin J.
\]
The result holds for 
\[\eps = \min_{J\subseteq\{1,\dots, n\}} \min_{k\in J^c} d(x_J, x_k)-r\] with the same centers $x_J$ for the balls.
\end{proof}

\subsection{Dudley's Argument and the VC-dimension of families of half-spaces}
We isolate the geometric essence of the proof of the VC-dimension of $\B$ on $\R^d$ in \cite{dudley1979balls}. As above, we identify sets with their associated classifiers.

\begin{lemma}\label{lemma abstract dudley}
Consider two classes of sets $\mathcal E$ and $\mathcal S$. Assume the following: If $E, F\in \mathcal E$, then there exists $S\in \mathcal S$ such that $E\setminus F \subseteq S$ and $(F\setminus E) \cap S = \emptyset$. Then
\[
\vcdim (\mathcal E) \leq \vcdim(\mathcal S).
\]
\end{lemma}

\begin{proof}
Assume that $\mathcal X$ is shattered by $\mathcal E$. Let $\mathcal Y$ be any subset of $\mathcal X$ and $E, F \in \mathcal E$ such that $\mathcal X \cap E = \mathcal Y$ and $\mathcal X\cap F = \mathcal X\setminus \mathcal Y$. Take $S \in \mathcal S$ as above. Then if $x\in \mathcal X$, then
\[
x\in \mathcal Y \quad \Ra \quad x \in E\setminus F \quad \Ra\quad  x\in S\qquad \text{and}\qquad x\in \mathcal X \setminus \mathcal Y \quad \Ra \quad x\in F\setminus E \quad \Ra\quad x\notin S.
\]
Thus $\mathcal Y = \mathcal X \cap S$. As $\mathcal Y$ was arbitrary, we see that $\mathcal X$ is also shattered by $\mathcal S$.
\end{proof}

The following example illustrates the derivation of the upper bound on the VC-dimension of Euclidean balls. For this we recall that the hypothesis class of closed half-spaces $\H$ and hypothesis class of open half-spaces $\H^\circ$ both have VC-dimension $d+1$.

\begin{example}
Let $\mathcal E = \{B_r(x) : r>0,\ x\in\R^d\}$ with respect to the Euclidean metric and $\mathcal S = \{ H_{w,b} : w\in \R^d,\ b\in \R\}$. Then $\mathcal E, \mathcal S$ satisfy the conditions of Lemma \ref{lemma abstract dudley}. Let $E, F \in \cal E$. If $E \subseteq F$ or $F \subseteq E$, then $E$ and $F$ trivially satisfy the conditions. Assume $E \not \subseteq F$ and $F \not \subseteq E$. Then the boundaries $\partial E$ and $\partial F$ of $E$ and $F$, respectively, are $(d-1)$-dimensional spheres whose intersection is either empty, a point, or a $(d-2)$-dimensional sphere which lies on a $(d-1)$-dimensional hyperplane.

If their intersection is empty, then consider the line segment connecting the center of $E$ to $F$. Pick $S \in \cal S$ to be the half-space $H_{w, w^Tp}$ where $p$ is the midpoint of the line segment, and $w$ is a unit vector running along the line segment, pointing towards the center of $E$.

If their intersection is a point, then we pick $S \in \cal S$ to be the half-space $H_{w, w^Tp}$ where $w$ is the unit normal vector to the common tangent plane based at the intersection point $p$ pointing towards the center of $E$.

Lastly, if their intersection forms a $(d-2)$-dimensional sphere on a $(d-1)$-dimensional hyperplane, then we pick $S \in \cal S$ to be the half-space $H_{w, b}$ where $w$ is the unit normal vector to the hyperplane pointing towards the center of $E$, and $b$ is the center of the $(d-2)$-dimensional sphere. 

Therefore, in all cases, we can separate $E$ and $F$ by some $S \in \cal S$. Hence, by Lemma \ref{lemma abstract dudley} and knowing the VC-dimension of the class of half-spaces, the VC-dimension of Euclidean ball classifiers is at most $d+1$.
\end{example}

We prove the bounds in hyperbolic spaces and spheres (for fixed radius) by the same technique. Both follow as special cases of a general result for a restricted class of half-spaces.

\begin{proof}[Proof of Lemma \ref{lm:half-space_d}]
    We begin by showing that 
    \[\vcdim(\H_{>}^\circ) \leq \vcdim(\H_0^\circ)= d.\] 
    by proving that the conditions of Lemma \ref{lemma abstract dudley} are satisfied for $\cal E = \H_{>}^\circ$ and $\cal S = \H_0^\circ$.
    
    Let $E, F \in \cal E$ with $E = H_{w, b}^\circ$ and $F = H_{v, c}^\circ$. Note that $b > 0$ and $c > 0$ as $E$ and $F$ are  exterior half-spaces. 
    Furthermore, we may assume without loss of generality that $w$ and $v$ are unit vectors since $H_{w, b}^\circ = H_{\frac{w}{||w||}, \frac{b}{||w||}}^\circ$.
        
    If $w \parallel v$, then the half-spaces $E$ and $F$ must satisfy one of the following three scenarios: 
    \begin{itemize}
        \item when $w = v$ and $c \geq b$, then $F \subseteq E$,
        \item when $w = v$ and $b \geq c$, then $E \subseteq F$, 
        \item when $w = -v$, then $E \cap F = \emptyset$, 
    \end{itemize}
    where the last statement follows from the fact that if $v^Tx \geq c$, then $w^Tx \leq -c < 0 < b$. In other words, $F \subseteq E^c$. Thus, these are the only three possible scenarios in the case that $w \parallel v$. 
    
    Now, if $F \subseteq E$ or $E \cap F = \emptyset$, we let $S = H_{w, 0}$ so that $E \setminus F \subseteq E \subseteq S$ and $(F \setminus E) \cap S = \emptyset \cap S = \emptyset$. Otherwise, if $E \subseteq F$, we let $S = H_{-w, 0}$ so that $E \setminus F = \emptyset \subseteq S$ and $(F \setminus E) \cap S \subseteq F \cap S = \emptyset$.
    
    If $w \nparallel v$, then on the plane $P = \text{span}(w, v)$, the lines $\{x \in P : w^T x = b\}$ and $\{x \in P : v^T x = c\}$ intersect at a unique point $p = (\mu_1 w + \mu_2 v) \in P$. Note that $p \neq 0$ as $E$ and $F$ are strictly exterior half-spaces. Let $q = (\lambda_1 w + \lambda_2 v) \in P$ be a nonzero vector orthogonal to $p$ so that
    \begin{equation}\label{eq:q_orthog_p}
        \lambda_1 b + \lambda_2 c = \lambda_1 w^T p +  \lambda_2 v^T p = (\lambda_1 w + \lambda_2 v)^T p = q^T p = 0.
    \end{equation}
    In particular, following the Gram-Schmidt process, one can choose
    \begin{align*}
        q \;\;=\;\; w - \text{proj}_p w \;\;=\;\; w - \frac{w^Tp}{p^Tp}\;p
        \;\;=\;\; \underset{\lambda_1}{\underbrace{\left(1 - \frac{w^Tp}{p^Tp} \cdot \mu_1 \right)}} w + \underset{\lambda_2}{\underbrace{\left(- \frac{w^Tp}{p^Tp} \cdot \mu_2 \right)}} v.
    \end{align*}

    We now show that $\lambda_1 \lambda_2 \leq 0$. Observe that because $w^T p = b$ and $v^T p = c$, then 
    \begin{equation*}\label{eq:exterior_half-space_pTp}
        p^Tp = (\mu_1w + \mu_2 v)^T p = \mu_1 b + \mu_2 c.
    \end{equation*}
    Therefore, because $b > 0$ and $c > 0$, then
    \begin{align*}
        (p^Tp)^2 \cdot \lambda_1 \lambda_2 
        &= (p^Tp)^2 \cdot \left(1 - \frac{w^Tp}{p^Tp} \cdot \mu_1 \right) \left(- \frac{w^Tp}{p^Tp} \cdot \mu_2 \right) \\
        &= \left(p^Tp - w^Tp \cdot \mu_1 \right) \left(- w^Tp \cdot \mu_2 \right) \\
        &= (\mu_1 b + \mu_2 c - \mu_1 b) (-\mu_2 b) \\
        &= -\mu_2^2 \,bc \leq 0
    \end{align*}
    means that $\lambda_1 \lambda_2 \leq 0$.
    We now assume without loss of generality that $\lambda_1 \geq 0$ and $\lambda_2 \leq 0$ as we can redefine $q$ to be the orthogonal vector $-\lambda_1 w - \lambda_2 v$ otherwise. 
    Thus, for every $x_0 \in (E \setminus F) = \{x \in \R^d : w^Tx > b, v^Tx \leq c \}$, because $\lambda_1b + \lambda_2c = 0$ by Equation (\ref{eq:q_orthog_p}), $\lambda_1 \geq 0$, and $\lambda_2 \leq 0$, then
    \[(\lambda_1 w + \lambda_2 v)^T x_0 \;=\; \lambda_1 w^T x_0 + \lambda_2 v^T x_0 \;>\; \lambda_1 b + \lambda_2 c \;=\; 0.\]
    Similarly, for every $x_0 \in (F \setminus E) = \{x \in \R^d : w^Tx \leq b, v^Tx > c \}$,
    \[(\lambda_1 w + \lambda_2 v)^T x_0 \;=\; \lambda_1 w^T x_0 + \lambda_2 v^T x_0 \;<\; \lambda_1 b + \lambda_2 c \;=\; 0.\]
    Finally, this proves that $(E \setminus F) \subseteq S$ and $(F \setminus E) \cap S = \emptyset$ for $S = H_{q, 0}^\circ$. Hence by Lemma \ref{lemma abstract dudley}, $\vcdim(\H_{>}^\circ) \leq \vcdim(\H_0^\circ) = d$. 

    Next, we show the reverse inequality by proving that $\H_{>}^\circ$ shatters the set of $d$ points $\{e_i\}_{i=1}^d$. For every non-empty $J \in \scr P(\{1, ..., n\})$, let $w = \sum_{j \in J}e_j$ and $b = 1/2$. Since $b > 0$, then $0 \notin H_{w, b}^\circ$, and so $H_{w, b}^\circ \in \H_{>}^\circ$. Moreover, 
    \begin{align*}
        w^T e_j &= 1 > b & \text{for every $j \in J$,} \\
        w^T e_k &= 0 < b & \text{for every $k \notin J$.} 
    \end{align*}
    And in the case where $J = \emptyset$, let $w = -\sum_{j=1}^n e_j$ and $b = 1/2$ so that $w^T (e_k) = -1 < b$ for all $k \in \{1, ..., n\}$.
    Hence $\H_{>}^\circ$ shatters $\{e_i\}_{i=1}^d$, and so $\vcdim(\H_{>}^\circ) \geq d$. Thus, we have $\vcdim(\H_{>}^\circ) = d$. 

    We now prove the VC-dimension of the other listed hypothesis classes is also $d$. Because all above arguments were made with strict inequalities, with minor changes, one can see that $\vcdim(\H_{>}) = d$. Secondly, because the hypothesis class of complement sets $\H^c$ has the same VC-dimension as the original class $\H$ for any given $\H$, we have that 
    \begin{align*}
         &\vcdim(\H_{<}) = \vcdim(\H_{>}^\circ) = d, \\
        &\vcdim(\H_{<}^\circ) = \vcdim(\H_{>}) = d.
    \end{align*}
    
    Next, we will show that $\vcdim(\H_{\geq}^\circ) = d$. 
    Suppose to the contrary that there exist $d+1$ points $\textbf{p}_0, ..., \textbf{p}_d \in \R^d$ that is shattered by $\H_{\geq}^\circ$. In particular, for each fixed $J \in \scr P(\{0, ..., d\})$, there exists $H_{\textbf{w}, b}^\circ \in \H_{\geq}^\circ$ such that $\langle \textbf{w}, \textbf{p} \rangle > b \geq 0$ if and only if $j \in J$. Let $\epsilon \geq 0$ be the minimum distance between $\{\textbf{p}_j\}_{j\in J}$ and the boundary of $H_{\textbf{w}, b}^\circ$. 
    As $H_{\textbf{w}, b}^\circ$ is an open set containing $\{\textbf{p}_j\}_{j\in J}$, it must be that $\epsilon > 0$. Hence, for all $j \in J$, $\langle \textbf{w}, \textbf{p} \rangle > b+\frac{\epsilon}{2} > 0$. Therefore, for every $H_{\textbf{w}, b}^\circ \in \H_{\geq}^\circ$ which separates $\{\textbf{p}_j\}_{j \in J}$, there exists $H_{\textbf{w}, b+\frac{\epsilon}{2}}^\circ \in \H_{>}^\circ$ which separates $\{\textbf{p}_j\}_{j \in J}$. In other words, $\H_{>}^\circ$ shatters $\{\textbf{p}_j\}_{j=0}^d$. This contradicts the fact that $\vcdim(\H_{>}^\circ) = d$. Thus, $d = \vcdim(\H_{>}^\circ) \leq \vcdim(\H_{\geq}^\circ) < d+1$ implies that $\vcdim(\H_{\geq}^\circ) = d$, and by preservation of complement sets, $\vcdim(\H_{\leq}) = d$.
\end{proof}

\begin{proof}[Proof of Lemma \ref{lm:half-space_d+1}]
    For each $j \in \{1, ..., d\}$, let $\textbf{p}_j = \textbf{e}_j$, and let $\textbf{p}_{d+1} = -\textbf{e}_1$. Suppose $J \in \scr P(\{1, ..., d+1\})$ is non-empty. 
    If $1 \in J$ and $d+1 \in J$,
    let 
    \[
    \textbf{v}_J = \sum_{j \in J} \textbf{p}_j - \sum_{k \notin J} \textbf{p}_k
    \] 
    and let $\hat{\textbf{v}}_J = \frac{\textbf{v}_J}{||\textbf{v}_J||}$. Then $\textbf{p}_j \in H_{\hat{\textbf{v}}_J, 0}$ if and only if $j \in J$. 

    Otherwise, if $1 \notin J$ or $d+1 \notin J$, let 
    \[\textbf{v}_J = \sum_{j \in J} \textbf{p}_j.\]
    Then $\textbf{p}_j \in H_{\hat{\textbf{v}}_J, \frac{1}{||\textbf{v}_J||}}$ if and only if $j \in J$. 
    Lastly, if $J = \emptyset$, then pick $\textbf{v}_J = -\textbf{e}_d$. Then for all $k \in \{0, ..., d+1\}$, $\textbf{p}_k \notin H_{\hat{\textbf{v}}_J, \frac{1}{2}}$.
    Therefore, $\H_{\geq}$ shatters $d+1$ points. Thus,
    \[d+1 \leq \vcdim(\H_{\geq}) \leq \vcdim(\H) = d+1.\] 
    Additionally, as $\H_{\leq}^\circ$ is the collection of complement sets to $H_{\geq}$, 
    \[\vcdim(\H_{\leq}^\circ) = \vcdim(\H_{\geq}) = d+1.\qedhere
    \] 
\end{proof}

\begin{proof}[Proof of Lemma \ref{lemma vc-dimension of homogeneous half-spaces}]
As $\vcdim(\H_0^\circ) = d$, there exists a set of $d$ points $\textbf{p}_1, ..., \textbf{p}_d \in \R^d$ that is shattered by $\H_0^\circ$. Let $\textbf{p}_0 = 0$. Note that because $0 \notin H_{w, 0}^\circ$ for all $H_{w, 0}^\circ \in \H_0^\circ$, then $\textbf{p}_0 \neq \textbf{p}_j$ for all $j \in \{1, ..., d\}$. 

Let $J \in \scr P(\{0, 1, ..., d\})$. If $0 \notin J$, then there exists some $H_{w, 0}^\circ \in \H^\circ_0$ such that $\textbf{p}_j \in H_{w, 0}^\circ$ if and only if $j \in J$ by definition of $\{\textbf{p}_i\}_{i=1}^d$. Now if $0 \in J$, then $J^C \in \scr P(\{1, ..., d\})$. Hence there exists some $H_{w, 0}^\circ \in \H^\circ_0$ such that $\textbf{p}_j \in H_{w, 0}^\circ$ if and only if $j \in J^C$. Equivalently, there exists some $(H_{w, 0}^\circ)^C = H_{-w, 0} \in \H$ such that $\textbf{p}_j \in  H_{-w, 0}$ if and only if $j \in J$. Thus, $d+1 \leq \vcdim(\H_0 \cup \H_0^\circ)$.

Finally, we use a similar proof technique to the upper bound argument presented in the proof of Lemma \ref{lm:half-space_d} to derive the upper bound of $ \vcdim(\H_0 \cup \H_0^\circ) $. Because any set of points that can be separated by a closed half space $H_{w, b}$ can also be separated by a slightly perturbed open half space $H_{w, b-\eps}^\circ$, then $d+1 \leq \vcdim(\H_0 \cup \H_0^\circ) \leq \vcdim(\H^\circ) \leq d+1.$
\end{proof}

\subsection{Spheres and Hyperbolic Spaces}

We begin by proving a VC-dimension bound in hyperbolic spaces. We choose the hyperboloid model of hyperbolic space due to our choice of techniques, but naturally, the result is independent of this choice for the underlying Riemannian manifold.
    
\begin{proof}[Proof of Theorem \ref{theorem hyperbolic}]
    Let $D = \text{diag}(-1, 1, ..., 1) \in \R^{(d+1) \times (d+1)}$.
    Consider the hyperboloid model of the $d$-dimensional hyperbolic space
    \[\HH = \left\{x \in \R^{d+1} : x_0^2 -  x_1^2 - ... - x_d^2 = 1,\; x_0 > 0 \right\}\]
    and its accompanying metric
    \[d_\HH(x, y) = \text{arcosh}(-x^TDy).\]
    Fix some $\textbf{p} = (p_0, p_1, ..., p_d) \in \HH$. Let $\rho = (p_1, ..., p_d)$ so that $\textbf{p} = (p_0, \rho)$ and $\rho^T \rho = p_0^2 - 1$. Consider the $(d+1) \times (d+1)$ matrix 
    \[  M_\textbf{p} = 
    \begin{pNiceArray}{cc}
      p_0 & \rho \\
      \rho & I + \frac{\rho \rho^T}{1 + p_0}
    \end{pNiceArray}.
    \]
    Note that by definition of $M_\textbf{p}$, we have $\textbf{p} = M_\textbf{p}\textbf{e}_0$. Furthermore, $M_\textbf{p}$ is a Lorentz matrix since
    \[(M_\textbf{p})^TDM_\textbf{p} 
    = \begin{pNiceArray}{cc}
      -p_0^2 + \rho^T \rho & -p_0 \rho + \rho^T \left( I + \frac{\rho \rho^T}{1 + p_0} \right) \\
      -p_0 \rho + \left( I + \frac{\rho \rho^T}{1 + p_0} \right) \rho & \left( I + \frac{\rho \rho^T}{1 + p_0} \right)^T\left( I + \frac{\rho \rho^T}{1 + p_0} \right)
    \end{pNiceArray} = D.\]
    Lastly, consider the open ball $B_r(\textbf{p})$ with respect to $d_\HH$ for some fixed $r \in (0, \infty)$. Observe that because hyperbolic cosine is an increasing function on $(0, \infty)$,  then for all $x \in \HH$,
    \begin{align*} \label{eq:hyperbolic_half-space}
      x \in B_r(\textbf{e}_0) 
      &\iff\; d_\HH(\textbf{e}_0, x) < r 
      \;\iff\; \text{arcosh}(x_0) < r \\
      &\iff\;  x_0 < \cosh(r) 
      \;\iff\; \textbf{e}_0^T x < \cosh(r) 
      \;\iff\; x \in H_{-\textbf{e}_0, -\cosh(r)}^\circ.  
    \end{align*}
    Thus, because Lorentz matrices preserve $d_\HH$,
    \[B_r(\textbf{p}) = B_r(M_\textbf{p} \textbf{e}_0) = M_\textbf{p}  \cdot B_r(\textbf{e}_0) = M_\textbf{p}  \cdot H_{-\textbf{e}_0,\; -\cosh(r)}^\circ = H_{-(M_\textbf{p}^{-1})^T \textbf{e}_0,\; -\cosh(r)}^\circ.\]
    Finally, because $-\cosh(r) < 0$ for all $r \in (0, \infty)$, then $H_{-(M^{-1})^T \textbf{e}_0,\; -\cosh(r)}^\circ$ is an interior half space of dimension $(d+1)$. Thus, because $\B^\circ \subseteq \H_{<}^\circ$, then $\vcdim(\B^\circ) \leq \vcdim(\H_{<}^\circ) = d+1$. Furthermore, because $-\cosh(r)$ is not only negative, but in fact $-\cosh(r) < -1$ for all $r \in (0, \infty)$, the above argument can be extended to closed balls. Hence we also have that $\vcdim(\B) \leq \vcdim(\H_{<}) = d+1$.
    
    To get the reverse inequality, we need to show that there exist $d+1$ points in $\HH$ shattered by $\B_r^\circ$. First, because $r > 0$, then $\cosh^2(r) > 1$. Hence there exists some $\mu > 0$ such that $1 + \mu^2 < \cosh^2(r)$. 
    Let $\{\textbf{p}_0, ..., \textbf{p}_d\} \in \R^d$ form a $(d+1)$-regular simplex in the $(x_0 = 0)$-plane centered at the origin with $\mu = ||\textbf{p}_j||$ for all $j \in \{0, ..., d\}$. 

    Next, for every $J \in \scr P(\{0, 1, ..., d\})$ with $J \neq \emptyset$ and $J \neq \{0, 1, ..., d\}$, let 
    \[\textbf{v}_J = \lambda_J \left(\sum_{i \in J} \textbf{p}_i\right)\]
    where $\lambda_J$ is the positive solution to the following equation
    \begin{equation}
        (1 + \mu^2)(1 + ||\textbf{v}_J||^2) = \cosh^2(r).
    \end{equation}
    Note that a real solution exists because we have picked $\mu > 0$ such that $1 + \mu^2 < \cosh^2(r)$.
    Now for all such $\textbf{p}_j$ and $\textbf{v}_J$, we define 
    \begin{align*}
        \hat{\textbf{p}}_j &= \left( \sqrt{1 + ||\textbf{p}_j||^2},\; \textbf{p}_j \right) = \left( \sqrt{1 + \mu^2},\; \textbf{p}_j \right) \in \HH, \;\;\text{ and} \\
        \hat{\textbf{v}}_J &= \left( \sqrt{1 + ||\textbf{v}_J||^2},\; \textbf{v}_J \right) \in \HH. 
    \end{align*}
    Note that the central angle of a regular $d$-simplex is $\arccos(-\frac{1}{d})$. Hence for all $i \neq j$, 
    \[\textbf{p}_j^T \textbf{p}_i \;\;=\;\;  - \frac{1}{d} \cdot ||\textbf{p}_j|| \cdot ||\textbf{p}_i|| \;\;=\;\; - \frac{\mu^2}{d}.\]
    Observe that because $ 1 \leq |J| \leq d$, then for all $j \in J$, 
    \begin{align*}
    -\hat{\textbf{p}}_j^T \,D \,\hat{\textbf{v}}_J
    &= \sqrt{1 + \mu^2} \cdot \sqrt{1 + ||\textbf{v}_J||^2} - \textbf{p}_j^T\, \textbf{v}_J\\
    &= \cosh(r) - \lambda_J \left( \textbf{p}_j^T\textbf{p}_j + \sum_{i \in J,\, i \neq j} \textbf{p}_j^T\textbf{p}_i \right) \\
    &=  \cosh(r) - \lambda_J \left( \mu^2 - \sum_{i \in J,\, i \neq j} \frac{\mu^2}{d} \right) < \cosh(r),
    \end{align*}
    and for all $k \notin J$, 
    \begin{align*}
    -\hat{\textbf{p}}_j^T \,D \,\hat{\textbf{v}}_J 
    &= \sqrt{1 + \mu^2} \cdot \sqrt{1 + ||\textbf{v}_J||^2} - \textbf{p}_k^T \textbf{v}_J\\
    &= \cosh(r) - \lambda_J \left(- \sum_{i \in J} \frac{\mu^2}{d} \right) > \cosh(r).
    \end{align*}
    Therefore, for all $J \in \scr P(\{0, 1, ..., d\})$ with $J \neq \emptyset$ and $J \neq \{0, 1, ..., d\}$, we have shown that $\hat{\textbf{p}}_j \in B_r(\hat{\textbf{v}}_J)$ if and only if $j \in J$. 
    
    If $J = \emptyset$, then we utilize the fact that $\HH$ is unbounded and pick some sufficiently distant center $\hat{\textbf{v}}_J \in \HH$ such that $\hat{\textbf{p}}_j \notin B_r(\hat{\textbf{v}}
    _J)$ for all $j \notin J$. Finally, for $J = \{0, 1, ..., d\}$, let $\hat{\textbf{v}}_J = (1, 0, ..., 0) \in \HH$. Then $\hat{\textbf{p}}_j \in B_r(\hat{\textbf{v}}
    _J)$ for all $j \in J$ since we picked $\mu > 0$ such that $1 + \mu^2 < \cosh^2(r)$.

    Thus, we have found $d+1$ points in $\HH$ which can be shattered by $\B_r^\circ$. Therefore,
    \[ d+1 \leq \vcdim(\B_r^\circ) \leq \vcdim(\B^\circ) \leq d+1.\]
    Moreover, as all of the above arguments were strict inequalities, the points can also be shattered by $\B_r$, and so
    \[
    d +1 \leq \vcdim(\B_r) \leq \vcdim(\B) \leq d+1. \qedhere
    \]
\end{proof}

The proof for the bounds on the sphere uses a similar proof technique to that of the hyperbolic space in which the geodesic balls on the sphere can be written as half-spaces in $\R^{d+1}$. We will see that all of the nuances in VC-dimension of half-spaces therefore carry over to the geodesic balls on $\SS^d$.

\begin{proof}[Proof of Theorem \ref{theorem sphere}]
Fix some $\textbf{p} \in \SS^d$. As $\textbf{p}$ is a unit vector in $\R^{d+1}$, there exists $d$ additional unit vectors $\textbf{u}_1, ..., \textbf{u}_d$ such that $\{ \textbf{p}, \textbf{u}_1, ..., \textbf{u}_d\}$ form an orthonormal basis for $\R^{d+1}$. Consider the $(d+1) \times (d+1)$ matrix $M_\textbf{p} = (\textbf{p}, \textbf{u}_1, ..., \textbf{u}_d)$. As $M_\textbf{p}$ is an orthogonal matrix, it preserves distances with respect to $d_{\SS^d}$. Let $r \in (0, \pi)$. Then using the same line of argument as in the proof of Theorem \ref{theorem hyperbolic} with the exception that cosine is a \textit{decreasing} function on $(0, \pi)$, 
\begin{equation}\label{eq:ball_is_half-space}
    B_r(\textbf{p}) = H_{(M_\textbf{p}^{-1})^T\textbf{e}_0, \cos(r)}^\circ  =  H_{\textbf{p}, \cos(r)}^\circ .
\end{equation}
Utilizing this fact, we now prove Theorem \ref{theorem sphere} in a series of claims.

\begin{claim}
For $r \in (0, \pi/2)$, $\vcdim(\B_r^\circ) = \vcdim(\B_r) = d+1$.
\end{claim}

    \begin{claimproof}
    As $r \in (0, \pi/2)$, then $\cos(r) > 0$. Hence $B_r(p) = H_{p, \cos(r)}^\circ \in \cal H_>^\circ$ for all $B_r(p) \in \B_r^\circ$. In other words, $\B_r^\circ \subseteq \H_{>}^\circ$. Thus, by Lemma \ref{lm:half-space_d},
    \begin{align*}
        &\vcdim(\B_r^\circ) \leq \vcdim(\H_{>}^\circ) = d+1 & r \in (0, \pi/2) 
    \end{align*}

    Now to show the reverse inequality, pick $\mu \in (0, 1)$ such that $\cos(r) < 1 - \mu^2 < 1$. Let $\textbf{p}_j$, $0 \leq j \leq d$, be as defined in proof of Theorem \ref{theorem hyperbolic} with respect to this redefined value of $\mu$. Now for $J \in \scr P(\{0, 1, ..., d\})$ with $J \neq \emptyset$ and $J \neq \{0, 1, ..., d\}$, let $\textbf{v}_J$ be as defined in proof of Theorem \ref{theorem hyperbolic} but where $\lambda_J$ is now the positive solution to the equation
    \[(1 - \mu^2)(1 - ||\textbf{v}_J||^2) = \cos^2(r).\]
    Note that a real solution exists since we have picked $\mu \in (0, 1)$ such that $\cos(r) < 1 - \mu^2 < 1$. Additionally, this condition implies that $||\textbf{v}_J||^2 \in (0, 1)$. Hence for each $\textbf{p}_j$ and $\textbf{v}_J$, the following are well-defined points in $\SS^d$:
    \begin{align*}
        \hat{\textbf{p}}_j &= \left( \sqrt{1 - ||\textbf{p}_j||^2},\; \textbf{p}_j \right) = \left( \sqrt{1 - \mu^2},\; \textbf{p}_j \right) \in \SS^d, \;\;\text{ and} \\
        \hat{\textbf{v}}_J &= \left( \sqrt{1 - ||\textbf{v}_J||^2},\; \textbf{v}_J \right) \in \SS^d. 
    \end{align*}
    
    As previously noted, because cosine is decreasing on $(0, \pi/2)$, in order to show $\hat{\textbf{p}}_j \in B_r(\hat{\textbf{v}}_J)$ if and only if $j \in J$, one needs to show that $\hat{\textbf{p}}_j^T\, \hat{\textbf{v}}_J > \cos(r)$ if and only if $j \in J$. And indeed, one can use analogous calculations to the ones used in the proof of Theorem \ref{theorem hyperbolic} to verify this claim.
    
    Next, for $J = \emptyset$, let 
    $\hat{\textbf{v}}_J = (-1, 0, ..., 0)$. Then $\hat{\textbf{p}}_k \notin B_r(\hat{\textbf{v}}_J)$ for all $k \notin J$ since $\cos(r) > 0$. Lastly, for $J = \{0, 1, ..., d\}$, let $\hat{\textbf{v}}_J = (1, 0, ..., 0)$. Then $\hat{\textbf{p}}_j \in B_r(\hat{\textbf{v}}_J)$ for all $j \in J$ since $\cos(r) < 1 - \mu^2$. Thus, we have found a set of $d+1$ points on $\SS^d$ that can be shattered by $\B_r^\circ$, and so $\vcdim(\B_r^\circ) = d+1$. Similarly, because all of the above inequalities were strict inequalities, we also have that $\vcdim(\B_r) = d+1$. 
\end{claimproof}

\begin{claim}
    For all $r \in (\pi/2, \pi)$, $\vcdim(\B_r^\circ) = \vcdim(\B_r) = d+1$.
\end{claim}

    \begin{claimproof}
    Note that the complement of all $B_r^\circ(p) \in \B_r^\circ$ is $B_{\pi-r}(-p) \in \B_{\pi-r}$. Therefore, because  complement sets preserve VC-dimension and $\pi - r \in (0, \pi/2)$, it follows from Claim 1 that
    \begin{align*}
        \vcdim(\B_r^\circ) &= \vcdim(\B_{\pi-r}) = d+1, \\
        \vcdim(\B_r) &= \vcdim(\B_{\pi-r}^\circ) = d+1.
    \end{align*}
    \end{claimproof}

\begin{claim}
     For $r = \pi/2$, $\vcdim(\B_r^\circ) = \vcdim(\B_r) = d+1$.
\end{claim}

    \begin{claimproof}
        For any $\lambda > 0$,
        \[B_{\pi/2}^\circ(p) \in \B_{\pi/2}^\circ \iff  H_{p, 0}^\circ = H_{\lambda p, 0}^\circ \in \H_0^\circ,\]
        \[B_{\pi/2}(p) \in \B_{\pi/2} \iff  H_{p, 0} = H_{\lambda p, 0}  \in \H_0.\]
        by Equation (\ref{eq:ball_is_half-space}).
        Therefore
        \begin{align*}
            \vcdim(\B_{\pi/2}^\circ) &= \vcdim(\H_0^\circ) = d+1, \\
            \vcdim(\B_{\pi/2}) &= \vcdim(\H_0) = d+1. 
        \end{align*}
    \end{claimproof}

\begin{claim}
    $\vcdim(\B_{(0, \pi/2]}^\circ) = \vcdim(\B_{(0, \pi/2)}^\circ) = \vcdim(\B_{(0, \pi/2)}) = d+1$.
\end{claim}

    \begin{claimproof}
        For all $r \in (0, \pi/2)$, we have $\B_r^\circ \subseteq \H_{>}^\circ$. Hence $\B_{(0, \pi/2)}^\circ \subseteq \H_{>}^\circ$. Therefore, by Lemma \ref{lm:half-space_d}
        \begin{align*}
            &d + 1 \;\leq\; \vcdim(\B_r^\circ) \;\leq\;\vcdim(\B_{(0, \pi/2)}^\circ) \leq \vcdim(\H_{>}^\circ ) \leq d+1, \\
            &d + 1 \;\leq\; \vcdim(\B_r) \;\leq\;\vcdim(\B_{(0, \pi/2)}) \leq \vcdim(\H_{>}) \leq d+1, \\
            &d + 1 \;\leq\; \vcdim(\B_r^\circ) \;\leq\;\vcdim(\B_{(0, \pi/2]}^\circ) \leq \vcdim(\H_{\geq}^\circ) \leq d+1.
        \end{align*}
    \end{claimproof}

\begin{claim}
    $\vcdim(\B_{(0, \pi/2]}) = d+2$
\end{claim}

\begin{claimproof}
     Let $\left\{ \hat{\textbf{p}}_j\right\}_{j=0}^{d+1}$ and $\left\{ \hat{\textbf{v}}_J\right\}_{J \in \scr P(\{0, ..., d+1\})}$ be the set of points constructed in the proof of Lemma \ref{lm:half-space_d+1} for dimension $d+1$. Note that all $\hat{\textbf{p}}_j \in \SS^d$ and $\hat{\textbf{v}}_J \in \SS^d$. Moreover observe that all half-space arguments made in the proof translate to $\B_{(0, \pi/2]}$. Namely, by Equation (\ref{eq:ball_is_half-space}), 
     \begin{align*}
         &H_{\hat{\textbf{v}}_J, 0} = B_{\pi/2}(\hat{\textbf{v}}_J) \in \B_{(0, \pi/2]} ,\\
         &H_{\hat{\textbf{v}}_J, \frac{1}{||\textbf{v}_J||}} = B_{\arccos(\frac{1}{||\textbf{v}_J||})}(\hat{\textbf{v}}_J) \in \B_{(0, \pi/2]}, \\
        &H_{\hat{\textbf{v}}_J, \frac{1}{2}} = B_{\pi/3}(\hat{\textbf{v}}_J) \in \B_{(0, \pi/2]} .
     \end{align*}
    Therefore, the set of $d+2$ points $\left\{ \hat{\textbf{p}}_j\right\}_{j=0}^{d+1} \subseteq \SS^d$ can be shattered by $\B_{(0, \pi/2]}$. Thus,
    \[ d+2 \leq \vcdim(\B_{(0, \pi/2]}) \leq \vcdim(\H) \leq d+2.\] 
\end{claimproof}

\begin{claim}
    $\vcdim(\B^\circ) = \vcdim(\B) = d+2$.
\end{claim}

    \begin{claimproof}
        Because $\B^\circ \subseteq \H^\circ$ and $\B \subseteq \H$, then it immediately follows that 
        \begin{align*}
            \vcdim(\B^\circ) &\leq \vcdim(\H^\circ) = d+2, \\
            \vcdim(\B) &\leq \vcdim(\H) = d+2. 
        \end{align*}
        To see the reverse inequality, recall the set of points $\left\{ \hat{\textbf{p}}_0, ..., \hat{\textbf{p}}_d\right\}$ and $\left\{ \hat{\textbf{v}}_J : J \in \scr P(\{0, ..., d\}), J \neq \emptyset \right\}$ constructed in the proof of Claim 1 for some fixed $r \in (0, \pi/2)$.
        We now let the $(d+2)$-th point $\hat{\textbf{p}}_{d+1} = (-1, 0, ..., 0)$. 
        
        First of all, as we may pick a ball of any radius $r \in (0, \pi)$, there exists some sufficiently small radius and point $\hat{\textbf{v}}_\emptyset \notin \left\{ \hat{\textbf{p}}_0, ..., \hat{\textbf{p}}_{d+1}\right\}$ for which the intersection of $B_r(\hat{\textbf{v}}_\emptyset)$ with $\left\{ \hat{\textbf{p}}_0, ..., \hat{\textbf{p}}_d\right\}$ is empty. Additionally, for $J = \{0, ..., d+1\}$, let $\textbf{v}_J = - \textbf{v}_\emptyset$. Observe that $B_\pi(\textbf{v}_J) = \SS^d - \{\textbf{v}_\emptyset\}$, and so $\{\textbf{p}_0, ..., \textbf{p}_{d+1}\} \subseteq B_\pi(\textbf{v}_J)$.
        
        Let $J \in \scr P(\{0, ..., d+1\})$, $J \neq \emptyset$, $J \neq \{0, ..., d+1\}$. If $d+1 \notin J$, then $J \in \scr P(\{0, ..., d\})$. Hence by the proof of Claim 1, for all $j \neq d+1$, 
        $\hat{\textbf{p}}_j \in B_r^\circ(\hat{\textbf{v}}_J)$ if and only if $j \in J$. It is left to check that $\hat{\textbf{p}}_{d+1} \notin   B_r^\circ(\hat{\textbf{v}}_J)$. This follows from the fact that $\hat{\textbf{v}}_J = (\sqrt{1 - ||\textbf{v}_J||^2},\, \textbf{v}_J)$ lies on the upper hemisphere. Therefore, because $r \in (0, \pi/2)$, then $\hat{\textbf{p}}_{d+1} = (-1, 0, ..., 0) \notin  B_r^\circ(\hat{\textbf{v}}_J)$.

        Now suppose that $d+1 \in J$. Then the complement $J^c = \{0, ..., d+1\} \setminus J$ does not contain $d+1$. Hence $\hat{\textbf{p}}_{k} \in  B_r^\circ(\hat{\textbf{v}}_{J^c})$ if and only if $k \in J^c$. Therefore $\hat{\textbf{p}}_{j}  \in B_{\pi-r} (-\hat{\textbf{v}}_{J^c})$ if and only if $j \in J$. In particular, because all of the arguments in the proof of Claim 1 were strict inequalities, $\hat{\textbf{p}}_{j}  \in B_{\pi-r}^\circ (-\hat{\textbf{v}}_{J^c})$ if and only if $j \in J$. 

        Thus, $\B^\circ$ shatters the set of $d+2$ points. One can take the closure of the above arguments and see that $\B$ also shatters the same set of $d+2$ points.
    \end{claimproof}
\end{proof}

Resembling Theorem \ref{theorem sphere} for balls of variable radius, we establish an analogous result for Euclidean spaces.

\begin{proof}[Proof of Corollary \ref{corollary complements}]
As $\vcdim(\B) = d+1$, there exist $d+1$ points $p_0, ..., p_d$ in $\R^d$ which can be shattered by a subset $\{ B_J \in \B : J \in \scr P(\{0, ..., d\})\}$ of balls in $\B$. Consider the finite union $\cup_J B_J$ of such balls. In particular, note the fact that $\cup_J B_J \neq \R^d$. Pick some point $p_{d+1} \in \R^d \setminus \cup_J B_J$. 
Let $J \in \scr P(\{0, ..., d+1\})$. 

\begin{itemize} 
\item If $d+1 \notin J$, then $J \in \scr P(\{0, ..., d\})$, and by assumption that $\{ B_J \in \B : J \in \scr P(\{0, ..., d\})\}$ shatters $\{p_0, ..., p_d\}$, we have that $p_j \in B_J \in \B $ if and only if $j \in J$. 
\item If $d+1 \in J$, then $J^c \in \scr P(\{0, ..., d\})$, and again by assumption, $p_j \in B_{J^c}$ if and only if $j \in J^c$. Equivalently, this gives us $p_j \in (B_{J^c})^c \in \B^c$ if and only if $j \in J$.
\end{itemize}
Therefore, $\{p_0, ..., p_{d+1}\}$ is shattered by $\B \cup \B^c$. Thus, $\vcdim(\B \cup \B^c) \geq d+2$.

Now suppose to the contrary that there exists a subset classifiers in $\B \cup \B^c$ that shatters $d+3$ points $p_0, ..., p_{d+2}$ in $\R^d$. That is, for every $J \in \scr P(\{0, ..., d+2\})$,  there exists some $\textbf{v}_J \in \R^d$ and $r_J > 0$ such that either
\begin{itemize}
    \item $||\textbf{p}_j - \textbf{v}_J ||^2 \leq r_J^2$ if and only if $j \in J$, or
    \item $||\textbf{p}_j - \textbf{v}_J ||^2 > r_J^2$ if and only if $j \in J$.
\end{itemize}
Equivalently, 
\begin{itemize}
    \item $\textbf{p}_j^T \textbf{v}_J - \frac{1}{2} (||\textbf{p}_j||^2 + ||\textbf{v}_J||^2 - r_J^2) \geq 0$ if and only if $j \in J$, or
    \item $\textbf{p}_j^T \textbf{v}_J - \frac{1}{2}(||\textbf{p}_j||^2 + ||\textbf{v}_J||^2 - r_J^2) < 0$ if and only if $j \in J$.
\end{itemize}
Therefore, if we let $\hat{\textbf{p}}_j = (\textbf{p}_j,\, ||\textbf{p}_j||^2) \in \R^{d+1}$, $\hat{\textbf{v}}_J = (\textbf{v}_J,\, -\frac{1}{2}) \in \R^{d+1}$ and $b_J = (||\textbf{v}_J||^2 - r_J^2)/2$, then
\begin{itemize}
    \item $\hat{\textbf{p}}_j \in H_{\hat{\textbf{v}}_J, b_J} \in \H$ if and only if $j \in J$, or
    \item $\hat{\textbf{p}}_j \in (H_{\hat{\textbf{v}}_J, b_J})^C \in \H^\circ$ if and only if $j \in J$.
\end{itemize}
Thus, we have found $d+3$ points $\hat{\textbf{p}}_0, ..., \hat{\textbf{p}}_{d+2}$ in $\R^{d+1}$ that can be shattered by $\H \cup \H^\circ$. However, this contradicts the fact that any labeling on a finite set can be realized by open subspaces if and only if it can be realized by closed half-spaces of $\R^{d+1}$, meaning that the set could be shattered by just $\H$. Thus, $\vcdim(\B \cup \B^c) = d+2$. 
\end{proof}

\subsection{Quotients of Model Spaces}

We illustrate that small balls in all constant curvature geometries behave just as Euclidean balls.

\begin{proof}[Proof of Theorem \ref{theorem quotient spaces}]
    Let $X\in\{\R^d, \mathbb H^d, S^d\}$ be a model space. By \cite[Corollary 11.13]{lee2018introduction}, there exists a discrete sub-group $\Gamma$ of the isometry group of $X$ which acts freely and properly discontinuously on $X$ such that $M = X/\Gamma$ and $d_M([p],[q]) = \inf_{\phi\in\Gamma} d_X(p, \phi(q))$ for two equivalence classes $[p], [q]\in X/\Gamma$.

    The claim is already proved if $\Gamma = \{\id\}$, i.e.\ $M=X$ for classes $\B_r$ and $\B_{<r}$ etc,  given that the injectivity radius of the Euclidean and hyperbolic space is infinite and that of any constant curvature sphere coincides with its diameter. Balls of radius $r<\inj(M)/2$ thus do not cross the equator relative to their center and form a set of VC-dimension $d+1$. From now on, we focus on the case in which $\Gamma$ contains multiple elements.

    Any geodesic in $X$ projects to a geodesic in $M$ under the quotient map. In particular, the geodesic connecting $p, \phi(p)$ in $X$ for $\phi\in\Gamma\setminus \{\id\}$ projects to a closed geodesic in $M$. Thus, the exponential map ceases to be injective at $d(p, \phi(p))/2$ at the latest since the equivalence class $[p]$ in $M$ can be realized as $\exp_{[q],M}(v) = \exp_{[q],M}(-v)$ where $q$ is the midpoint of the geodesic connecting $p$ and $\phi(p)$ in $X$ and $v$ is its velocity vector passing through $q$.

    Since the Jacobi equation is local, no new conjugate points are induced in the quotient. The injectivity radius of $M$ is therefore
    \[
    \inj(M) = \frac12\,\inf \big\{d_{X}(p, \phi(p)) : p\in X, \ \phi\in \Gamma\setminus \{\id\}\big\} = \frac12\,\inf \big\{d_{X}(p_0, \phi(p_0)) : \phi\in \Gamma\setminus \{\id\}\big\}
    \]
    for any fixed $p_0\in X$ (since constant curvature spaces are homogeneous and isotropic). The first equality is trivial in spaces of non-positive curvature (since the only closed geodesics are induced by the quotient map). In spherical spaces, it is true since 
    \[
    \inf \big\{d_{X}(p_0, \phi(p_0)) : \phi\in \Gamma\setminus \{\id\}\big\} \leq \diam(X) = \inj(X),
    \]
    meaning that the quotient by $\Gamma\neq \{\id\}$ always identifies a stricter constraint on the injectivity radius. For a more general discussion linking the injectivity radius and the length of shortest closed geodesics, see e.g.\ \cite[Chapter 6, Section 6]{petersen2006riemannian}.
    
    Take now $S= \{p_1, \dots, p_n\}\subset M$. If $S$ is shattered by metric balls of radius $r>0$ in the quotient space $M$, then there exists $p^*\in M$ such that $d_M(p^*, p_i) < r$ for all $i=1,\dots,n$. We select representatives $x_1,\dots, x_n, x^*\in X$ of the equivalence class in $M$ such that $d_X(x^*, x_i) = d_M(p^*, p_i)< r$.

    Let $I\subseteq\{1,\dots,n\}$. For $q_I\in M$ such that $B_r(q_I) \cap S = \{x_i : i\in I\}$. We select a representative $z_I \in X$ of the class of $q_I$ such that $d_X(z_I, x^*) = d_M(p^*, q_I)$. If $I\neq \emptyset$, then 
    \[
    d_X(x^*, z_I) = d_M(p^*, q_I) \leq d_M(p, p_i) + d_M(p_i, q_I) < 2r
    \]
    for $i\in I$. We claim that $d_X(z_I, x_j) = d_M(p_j, q_I)$ for all $j=1,\dots,n$. Specifically
    \[
    d_X(z_I, x_j) \leq d_X(z_I, x^*) + d(x^*, x_j) < 3r.
    \]
    On the other hand, if $\phi\in \Gamma\setminus\{\id\}$, then
    \[
    d_X\big(\phi(z_I), x_j\big) \geq d_X\big(\phi(z_I), z_I\big) - d_X(z_I, x_j) > 6r - 3r = 3r. 
    \]
    In particular, the set $S_X = \{x_1,\dots,x_n\}$ of lifts of $p_i\in S$ to $X$ is shattered by balls of the same radius in $X$. The empty set shattering point is easier to achieve in $X$ than the quotient space.
    
    The converse direction follows similarly, taking the projection of a shattered set in the model space $X$ to $M$ rather than lifting a set in $M$ to one in $X$. The proof is technically simpler as the projection is unique, but relies on the same isometric identification.
\end{proof}

\begin{table}
\begin{align*}
x_1 &= (0.348,\ 0.582)&
x_2 &= (0.57,\ 0.292)&
x_3 &= (0.294,\ 0.026)\\
x_4 &= (0.952,\ 0.854)&
x_5 &= (0.203,\ 0.731)
\end{align*}

\begin{tabular}{l |l || l |l } {\bf Subset} $I$ & {\bf Center} $p_I$ &  {\bf Subset} $I$ & {\bf Center} $p_I$\\ \hline\hline
$\emptyset$ & $(0.966,\ 0.252)$ & $\{1,2,3,4,5\}$ & $(0.5,\ 0.5)$\\
\hline
$\{1\}$ & $(0.087,\ 0.358)$ & $\{2,3,4,5\}$ & $(0.54,\ 0.018)$\\
$\{2\}$ & $(0.796,\ 0.339)$ & $\{1,3,4,5\}$ & $(0.262,\ 0.663)$\\
$\{3\}$ & $(0.251,\ 0.1)$ & $\{1,2,4,5\}$ & $(0.935,\ 0.5)$\\
$\{4\}$ & $(0.76,\ 0.745)$ & $\{1,2,3,5\}$ & $(0.452,\ 0.368)$\\
$\{5\}$ & $(0.97,\ 0.407)$ & $\{1,2,3,4\}$ & $(0.599,\ 0.894)$\\
\hline
$\{1,2\}$ & $(0.703,\ 0.376)$ & $\{3,4,5\}$ & $(0.186,\ 0.816)$\\
$\{1,3\}$ & $(0.17,\ 0.32)$ & $\{2,4,5\}$ & $(0.848,\ 0.575)$\\
$\{1,4\}$ & $(0.669,\ 0.747)$ & $\{2,3,5\}$ & $(0.389,\ 0.014)$\\
$\{1,5\}$ & $(0.045,\ 0.408)$ & $\{2,3,4\}$ & $(0.605,\ 0.938)$\\
$\{2,3\}$ & $(0.334,\ 0.147)$ & $\{1,4,5\}$ & $(0.031,\ 0.562)$\\
$\{2,4\}$ & $(0.814,\ 0.43)$ & $\{1,3,5\}$ & $(0.373,\ 0.915)$\\
$\{2,5\}$ & $(0.944,\ 0.418)$ & $\{1,3,4\}$ & $(0.624,\ 0.841)$\\
$\{3,4\}$ & $(0.629,\ 0.884)$ & $\{1,2,5\}$ & $(0.965,\ 0.422)$\\
$\{3,5\}$ & $(0.313,\ 0.953)$ & $\{1,2,4\}$ & $(0.703,\ 0.66)$\\
$\{4,5\}$ & $(0.894,\ 0.617)$ & $\{1,2,3\}$ & $(0.389,\ 0.231)$\\
\end{tabular}

\caption{\label{table shattering on torus}
Representatives $x_i,\ P_I \in[0,1)^2$ of five equivalence classes in $\T^2 = \R^2/\Z^2$ which are shattered by metric balls and the centers for those balls. Each ball in the shattering has radius $r(p) = d_{\T^2}(p,[0]) = \min_{z\in\Z^2}\|p-z\|_2$, the distance from $x_i$ to the decision boundary $\partial B_{r(P_I)}(P_I)$ is at least $0.001$ in all cases, so we do not need to distinguish between open and closed balls.
}
\end{table}

\begin{proof}[Proof of Lemma \ref{lemma flat tori}]
{\bf Lower bound.} The proof is similar to that of the claims in Theorem \ref{theorem vc normed vector spaces} on $\ell^1$-norms. We refer the reader to Table \ref{table shattering scaling radius} for an example of a dataset $\{x_1,\dots, x_5\}$ in $\T^2$ which is shattered by $B_{r(p)}(p)$ with $r(p) = d_{\T^2}(p, p^*)$ where $p^*$ is an arbitrary fixed point on $\T^2$, which we selected to be represented by the equivalence class of integer points in $\R^2/\Z^2$.

If $d = 2k+1$ is odd, we find that the VC-dimension of metric ball classifiers in $\T^d$ is at least that of metric ball classifiers in $\T^{2k}$ since $\T^{2k}$ embeds isometrically into $\T^d$. If $d=2k$ is even, we select $n = 5k$ points $x_{i,j}$ with $1\leq i \leq 5$ and $0\leq j\leq k-1$ such that $x_{i,j}$ has a copy of $x_i$ as in Table \ref{table shattering on torus} in coordinates $2j,\ 2j+1$ and zero entries in all others (projected to $\T^d$). For
\[
I \subseteq \{ (i,j) : i=1,\dots,5,\ j=0,\dots,k-1\}\qquad \text{set}\quad I_j = \{i : (i,j)\in I\}.
\]
With the centers $P_J\in\T^2$ given in Table \ref{table shattering on torus}, set $Q_I:= (P_{I_0},\dots,P_{I_{k-1}})\in\R^{2k}$ (projected to the torus) and $r_I:= d_{\T^d}(Q_I, [0])$. Then
\[
d_{\T^d}(P_I, x_{i,j})^2 = \sum_{l\neq j} d_{\T^2}(P_I, [0]_{\T^2})^2 +  d_{\T^2}(P_{I_j}, x_i)^2= r_I^2 +  d_{\T^2}(P_{I_j}, x_i)^2 -  d_{\T^2}(P_{I_j}, [0]_{\T^2})^2
\]
meaning that
\[
d_{\T^d}(P_I, x_{i,j}) \leq r_I \qquad\LRa\quad  d_{\T^2}(P_{I_j}, x_i)^2 \leq  d_{\T^2}(P_{I_j}\qquad\LRa\quad i\in I_j \qquad\LRa\quad (i,j) \in I.
\]
Thus the $5k$ points $x_{ij}$ are shattered by (a sub-class of) Riemannian balls in $\T^d$.

{\bf Upper bound:} The argument proceeds in full analogy to the proof that the VC-dimension of the class of $\ell^1$-balls in $\R^d$ has VC-dimension growing at most like $d\log(d)$ proved in Appendix \ref{appendix normed spaces}. We take the fundamental domain $[0,1)^d$ of $\T^d$ and $n$ representatives $p_1,\dots, p_n \in [0,1)^d$ such that balls $B_r(x)\subseteq \T^d$ shatter $p_1,\dots, p_n$. 

We note that $p_i \in B_r(x)$ in $\T^d$ if and only if there exists $n\in\Z^d$ such that $p_i + n \in B_r(x)$. Taking representatives $p_i, x\in[0,1)^d$, it naturally suffices to consider $n\in\{-1,0,1\}^d$. In fact, for every $i$, it suffices to consider $2^n$ vectors since the choice $n_j = 1$ is guaranteed to be suboptimal if $p_{i,j} \geq 1/2$ and the choice $n_j = -1$ is guaranteed to be suboptimal if $n_j<0.5$. 

We divide $[0,1)$ into at most $n+1$ segments at threshold values $t_{i,j} = p_{i,j} \pm 1/2\in[0,1)$ and $\T^d$ into at most $(n+1)^d$ regions by sub-dividing all coordinates accordingly. Inside any of the regions, the closest point representative $p_i + n_i$, $n_i\in\{-1,0,1\}^d$ does not change. Thus any sign pattern which can be realized on $p_i$ by balls on $\T^d$ with center in a fixed sub-region can also be realized as a sign pattern by Euclidean balls on the points $p_i+n_i$ for a fixed representative $n_i$. Since the VC-dimension of Euclidean balls is $d+1$, the counting argument can be taken verbatim from the proof of Theorem \ref{theorem vc normed vector spaces}.
\end{proof}

\subsection{Ambient dimension and intrinsic data dimension: Subspaces and graphs}
\label{appendix intrinsic vs ambient dimension}

We begin by noting the favorable shattering behavior for data which lie precisely on a low-dimensional subspace.

\begin{proof}[Proof of Corollary \ref{corollary vc dimension subspace}]
Without loss of generality, take $V = \R^k\times \{0\}$. If $z\in\R^d$, write $z = (x,y)$ with $x\in \R^k$ and $y\in\R^{d-k}$. For $r>0$, we have $B_r(z) \cap V = B_{\sqrt{r^2 - \|y\|^2}}(x) \times \{0\}$ (empty if $\|y\|>r$), i.e.\ the `effective hypothesis class' is that of Euclidean balls in $\R^k$. A set of points $(x_1,0), \dots, (x_n,0)$ is therefore shattered by $\B(\R^d)$ if and only if $x_1,\dots, x_n$ is shattered by $\B(\R^k)$.

Using the same notation but splitting $\R^d= \R^{k+1}\times \R^{d-(k+1)}$, we note that 
\[
S^k \cap B_r(z) =\big(S^k\cap B_{\sqrt{r^2- \|y\|^2}}(x)\big) \times \{0_{d-(k+1)}\} \in \R^{k+1} \times \R^{d-(k+1)}. 
\]
The intersection of an ambient space Euclidean ball with a sphere is a spherical cap and thus an intrinsic ball. Specifically, in $\R^{k+1}$, we have
\[
S^k \cap B_r( s e_{k+1}) = \{(\hat x, x_{k+1}) \in S^k : \|\hat x\|^2 + (x_{k+1}-s)^2 < r^2\}.
\]
We simplify
\begin{align*}
r^2 &> \|\hat x\|^2 + (x_{k+1}-s)^2
    = 1 - x_{k+1}^2 + x_{k+1}^2 - 2 x_{k+1}s + s^2 = 1-2x_{k+1}s + s^2 = 1 + s^2 - 2s\,\langle x, e_{k+1}\rangle.
\end{align*}
This can be rewritten in terms of the spherical distance as
\[
d_{S^k}(x, e_{k+1}) = \arccos (\langle x, e_{k+1}\rangle) < \arccos\left(\frac{1+s^2-r^2}{2s}\right)
\]
since the $\arccos$ function is monotonically decreasing and assuming that the right hand side is well-defined. This is the case if and only if 
\[
-1\leq \frac{1+s^2-r^2}{2s} \leq 1 \quad\LRa\quad -2|s|\leq 1+s^2 -r^2 \leq 2|s| \quad \LRa \quad r^2 \leq (1+|s|)^2 \quad \text{ and } (1-|s|)^2 \leq r^2
\]
and thus if and only if $1-|s| \leq r\leq 1+|s|$. By rotational symmetry, the intersection of an ambient space Euclidean ball $B_r(z)$ with the sphere is either empty or an intrinsic metric ball of radius
\[
R = \arccos \left(\frac{1+ \|x\|^2 - (r^2 - \|y\|^2)}{2\|x\|}\right)\qquad\text{centered at } \frac x{\|x\|}\text{ for }x\neq 0.
\]
If $x=0$, the intersection is either empty or the entire sphere. The balls are largest for $y=0$, so we are able to represent radii $R\geq \pi/2$ if and only if we can achieve
\[
\frac{1+ \|x\|^2-r^2}{2\|x\|}\leq 0 \qquad\LRa\quad 1+ \|x\|^2 - r^2 < 0 \quad \LRa\quad r\geq 1.
\]
The non-strict inequality suffices for closed balls since it allows us to achieve the full set for shattering by the effective hypothesis class $\B_{(0,\pi/2)}(S^k)\cup \{S^k\}$. Conversely, if $r<1$, the balls of the `effective hypothesis class' have intrinsic radius $< \pi/2$ and cannot shatter $k+2$ points by Theorem \ref{theorem sphere}.
\end{proof}

\begin{proof}[Computations for Example \ref{example 2d graph}]
\begin{itemize}
    \item All four points are in $\overline B_1(0)$.
    \item None of the points are in $\overline B_1(2e_1)$.
    \item The point $x_1$ is the only one of the four points lying in $\overline B_1(e_1)$.
    \item The points $x_1,x_2,x_3$ lie in $\overline B_1(0, \eps, \eps)$ but $x_4$ does not
    since
    \begin{align*}
    |x_1 - (0,\eps,\eps)|^2 &= |x_2 - (0,\eps,\eps)|^2 =  \beta^2 +\eps^2 = 1\\
    |x_4 - (0,\eps,\eps)| &= (\beta+\eps)^2 + \eps^2 > \beta^2 + \eps^2 = 1\\
    |x_3-(0,\eps,\eps)|^2 &= (\beta-\eps)^2 + (2\eps)^2 = \beta^2 -2\beta\eps + 5\eps^2 = 1 - 2\beta\eps + 4\eps^2\leq 1
    \end{align*}
    if 
    \[
    \beta\geq 2\eps \qquad \LRa \quad 1- \eps^2 \geq 4\eps^2 \qquad \LRa\quad \eps \leq \frac1{\sqrt 5}
    \]
    which is true by assumption.
    \item The points $x_1, x_3$ lie in the ball $\overline B_1(\beta/2, \beta/2,0)$ since
    \begin{align*}
    |x_1 - (\beta/2,\beta/2,0)|^2 &= 2(\beta/2)^2 + \eps^2 = \frac{\beta}2 + \eps^2 < 1\\
    |x_2 - (\beta/2,\beta/2,0)|^2 &= (3\beta/2)^2 + (\beta/2)^2 + \eps^2> \beta + \eps^2 = 1
    \end{align*}
    and analogously for $x_3,x_4$.
    \item The points $x_1,x_2$ lie in $B(0,0,\eps)$, but $x_3, x_4$ do not since
    \begin{align*}
    |x_1 - (0,0,\eps)|^2 &= |x_2 - (0,0,\eps)|^2 = \beta^2 <1\\
    |x_3 - (0,0,\eps)|^2 &= |x_4 - (0,0,\eps)|^2 = \beta^2 + (2\eps)^2 = 1+3\eps^2>1.
    \end{align*}
    \end{itemize}
    Up to a relabeling of points, these are the only subsets of the four-point set $\{x_1,x_2,x_3,x_4\}$, so the set is shattered. 
\end{proof}

In the following, we present various examples of point sets which can be shattered by metric balls although they are `close to' being low-dimensional. 

\begin{proof}[Proof of Theorem \ref{theorem ambient dimension graph}]
{\bf First case: $n=k$.} Assume that $\eps < 4/k$. The hypothesis class $\B_1(\R^k)$ shatters the $k$ unit vectors $e_1, \dots, e_k$ using the centers
\[
\hat x_I = \sqrt{\frac \eps{|I|}}\sum_{i\in I}e_i\quad \text{for }I\subseteq\{1,\dots,n\}
\]
since
\[
\|x_j-\hat x_I\|_2^2 
    = \begin{cases}
    1+\sum_{i\in I} \frac{\eps}{|I|} & j\notin I\\
    \big(1- \sqrt{\eps/|I|}\big)^2 +\sum_{i\in I,\ i\neq j} \frac{\eps}{|I|} &j\in I
    \end{cases} 
= \begin{cases} 1+\eps &j\notin I\\
    1+\eps - 2\sqrt{\eps/|I|} &j\in I.
    \end{cases}
\]
Note that
\[
1+\eps - 2\sqrt{\eps/|I|} \leq 1+\eps - 2\sqrt{\eps/k} < 1\quad\LRa\quad \eps< 2\sqrt{\frac\eps k} \quad \LRa \quad \eps< \frac4k.
\]
The result is thus trivial since $\B_1(\R^d)$ restricts to $\B_{[0,1]}(\R^k)$, which in particular contains $\B_1(\R^k)$.
We thus set $z_i = (e_i, 0) \in \R^k\times \R^{d-k}$ and choose centers $\hat z_I = (\hat x_I, 0)$. The function $f_\eps\equiv 0$ satisfies all conditions.

{\bf Second case: $n = d-k$.} Let $x_1,\dots, x_n$ be any set of $n$ points in $\R^k$ such that $\|x_i\|=1$ for all $i$ and $y_i = e_i\in\R^{d-k}$ as for Case 1. We denote the corresponding centers for shattering $\{y_1,\dots, y_n\}$ by $\hat y_I$. 
Set $z_i := \big(\sqrt{1-\eps^2 }x_i,\ \eps y_i\big)$. This set is shattered by balls of radius $1$ centered at points $\hat z_I = (0, \eps\,\hat y_I)$ since
\[
\|z_j-\hat z_I\|_2^2 
    = 1 - \eps^2 + \eps^2\|y_i - \hat y_I\|^2 \leq 1 
    \qquad\LRa \quad \|y_i - \hat y_I\|^2 \leq 1.
\]

It remains to show that the points $z_i$ lie on the graph of a smooth function. Let $\eta \in C^\infty(\R)$ be a cut-off function such that $\eta(r) = 1$ if $r \leq 0$ and $\eta(r) = 0$ if $r\geq 1$. Then the function 
\[
f_\eps :\R^k\to \R^{d-k}, \qquad f_\eps(x) := \eps \sum_{i=1}^n \eta\left( 2\,\frac{\big\|x - \sqrt{1-\eps^2}\,x_i\big\|}{\sqrt{1-\eps^2}\,\delta}\right)\,y_i
\]
satisfies the assumptions. While the sum involves $n$ terms, the supports of the different `bumps' do not overlap, meaning that suprema of $\|D^mf(x)\|$ over $x$ are unaffected.

{\bf Distant points on the sphere:} The $2k$ points $e_1, \dots, e_k, -e_1, \dots, -e_k$ are always admissible. The exponential bound follows by a measure argument.
For $X\in S^{k-1}$, consider the spherical cap
\[
C(X, \alpha) = \{x\in S^{k-1} : \langle x, X\rangle > \alpha\}.
\]
If $x\in S^{k-1}\setminus C(X,\alpha)$, then
\[
\|X- x\|^2 = \|X\|^2 - 2\langle X, x\rangle + \|x\|^2 \geq 2(1-\alpha).
\]

It is well-known that a spherical cap in a high-dimensional space has small volume: With respect to the natural $(k-1)$-dimensional area measure (i.e.\ the Hausdorff measure $\H^{k-1}$) we have
\[
\frac{|C(e_k, \alpha)|}{|S^{k-1}|} = 
\frac{|\{x \in S^{k-1} : x_k > \alpha\}|}{|S^{k-1}|} 
    \leq \frac{|\{x \in S^{k-1} : x_k > \sin\alpha\}|}{|S^{k-1}|} \leq \frac12 \sqrt{\frac\pi2} \exp\left(- \frac{(k-2)\alpha^2}2\right)
\]
using the bound of \cite[Theorem B]{gimeno2025fat}. Naturally, the volume of the cap is independent of the choice of point $X$.
By the volume bound, we cannot cover $S^{k-1}$ using a union of fewer than
\[
N:= \left\lceil \frac1{\frac12 \sqrt{\frac\pi2} \exp\left(- \frac{(k-2)\alpha^2}2\right)}\right\rceil =  \left\lceil \sqrt{\frac 8\pi} \exp\left(\frac{(k-2)\alpha^2}2\right)\right\rceil
\]
spherical caps $C(x_1,\alpha), \dots, C(x_N, \alpha)$. In particular, we can inductively find $N$ points $x_i$ such that $\|x_i-x_j\| \geq \sqrt{2(1-\alpha)}$. The bound follows by solving
\[
\sqrt{2(1-\alpha)} = \delta \qquad \LRa \quad 2 - 2\alpha = \delta^2 \quad \LRa \quad \alpha = \frac{2-\delta^2}2.\qedhere
\]
\end{proof}

The bound of Theorem \ref{theorem ambient dimension graph} is somewhat looser than that of Example \ref{example 2d graph} since we only rely on dimension $k$ or co-dimension $d-k$ for shattering rather than using both simultaneously.

\subsection{VC-dimension of unions of balls}

The proof combines two observations: A stability result as in Lemma \ref{lemma closed vs open balls}, which states that a shattered set can in fact be shattered by a family of sets\footnote{\ At least for `nice' sets like half-spaces and balls.}\ for which the distance to the decision boundary remains uniformly positive, and the fact that half-spaces can be approximated locally by balls of large radius. 

\begin{proof}[Proof of Corollary \ref{corollary vc-dimension union lower bound}]
{\bf Step 1. Stable shattering.} Let $x_1, \dots, x_n$ be a set which is shattered by a $k$-fold union of closed half-spaces. We recall that finite unions of closed sets are closed. 

Let $I\subseteq \{1,\dots, n\}$ and $H_{I, 1}, \dots, H_{I,k}$ such that
\[
\{x_1, \dots, x_n\} \cap \bigcup_{j=1}^k H_{I,j} = \{x_i : i\in I\}.
\]
Recall that the distance between two disjoint sets, one closed and one compact, is always positive. The minimum of a finite collection of positive numbers is positive. Thus 
\[
\min\left\{\dist\left(x_i, \bigcup_{j=1}^k H_{I, j} \right) : I\subseteq\{1,\dots,n\} \text{ and }i\notin I\right\} >0.
\]
It is therefore possible to shift all half-spaces to their exterior normal direction by a non-vanishing amount without changing the intersection with $\{x_1,\dots, x_n\}$. The distance from $\{x_i : i\notin I\}$ to the boundary remains positive, while that of $\{x_i : i\in I\}$ to the boundary is newly guaranteed to be positive. 

We conclude that for every $I\subseteq \{1,\dots,n\}$ there exists a collection of closed half-spaces $H_{I, 1}', \dots, H_{I,k}'$ such that 
\[
\{x_1, \dots, x_n\} \cap \bigcup_{j=1}^k H_{I,j}' = \{x_i : i\in I\}\quad\forall\ I\subseteq\{1,\dots,n\} \text{ and } \dist\left(\{x_i :i=1,\dots, n\}, \partial \bigcup_{j=1}^k H_{I,j}'\right)>0.
\]

{\bf Step 2. The expressivity of Euclidean balls.} Let $H = H_{w,b} = \{x : \langle w,x\rangle \geq b\}$ a closed half-space and $x\in H_{w,b}^\circ$. We want to approximate $H_{w,b}$ locally by balls of large radius. Note that
\[
\|z-tw\|^2 = \|z\|^2 - 2 t\langle w, z\rangle + t^2 \|w\|^2< (t\|w\|-b/\|w\|)^2 = t^2\|w\|^2 -2t b + b^2/\|w\|^2
\]
if and only if
\[
\|z\|^2 - b^2/\|w\|^2 < 2t \big(\langle w,z\rangle -b\big).
\]
In particular
\begin{enumerate}
\item If $z\in H_{w,b}^\circ$, then $z \in B_{t\|w\| - b/\|w\|}(tw)$ for all sufficiently large choices of $t$, specifically for all $t> (\|z\|^2 - b^2/\|w\|^2) / (\langle w,z\rangle - b)$.

\item Assume that $z\notin H_{w,b}$, i.e.\ $\langle w,z\rangle <b$. If $t> b/\|w\|^2$, then 
\begin{align*}
\|z\|^2 - b^2/\|w\|^2 - 2t \big(\langle w,z\rangle -b\big)
    &\geq \|z\|^2 - b^2/\|w\|^2 - 2\frac b{\|w\|^2} \big(\langle w,z\rangle -b\big)\\
    &= \|z\|^2 - 2 \left\langle \frac{b}{\|w\|}\,\frac{w}{\|w\|}, z\right\rangle + \frac{b^2}{\|w\|^2}\\
    &= \left\|z - \frac{b}{\|w\|}\,\frac{w}{\|w\|}\right\|^2 \geq 0,
\end{align*}
i.e.\ $z\notin B_{t\|w\|-b/\|w\|}(tw)$. 
\end{enumerate}
So if $K\subset H_{w,b}^\circ$ is compact (e.g.\ finite), and $t$ is sufficiently large, then
\[
K\subseteq B_{r(t)}(tw) \subseteq H_{w,b}\qquad\text{for }r(t) = t\|w\|-b/\|w\|.
\]
Thus for the purposes of shattering, large balls are indistinguishable from half-spaces. The same is true for a union of half-spaces. As the result holds for all large radii, it is sufficient to consider a single large radius given a {\em finite} union of half-spaces. 
\end{proof}

\section{VC-dimension: Normed Vector Spaces}\label{appendix normed spaces}

\begin{table}
\begin{align*}
x_1 &= (24,\ 32,\ -47)&
x_2 &= (37,\ 27,\ 6)&
x_3 &= (31,\ -8,\ -41)\\
x_4 &= (2,\ -12,\ -12)&
x_5 &= (-24,\ 15,\ -11)
\end{align*}

\begin{tabular}{l |l || l |l } {\bf Subset} $I$ & {\bf Center} $p_I$ &  {\bf Subset} $I$ & {\bf Center} $p_I$\\ \hline\hline
$\emptyset$ & $(-2,\ 0,\ 0)$ & $\{1,2,3,4,5\}$ & $(98,\ 98,\ -102)$\\
\hline
$\{1\}$ & $(0,\ 13,\ -39)$ & $\{2,3,4,5\}$ & $(98,\ 15,\ -12)$\\
$\{2\}$ & $(27,\ 13,\ 98)$ & $\{1,3,4,5\}$ & $(5,\ 98,\ -102)$\\
$\{3\}$ & $(31,\ -2,\ -8)$ & $\{1,2,4,5\}$ & $(19,\ 17,\ -18)$\\
$\{4\}$ & $(2,\ -102,\ -16)$ & $\{1,2,3,5\}$ & $(99,\ 99,\ -10.5)$\\
$\{5\}$ & $(-102,\ 14,\ 0)$ & $\{1,2,3,4\}$ & $(98,\ -102,\ -102)$\\
\hline
$\{1,2\}$ & $(24,\ 98,\ -3)$ & $\{3,4,5\}$ & $(-102,\ -12,\ -42)$\\
$\{1,3\}$ & $(-2,\ 6,\ -102)$ & $\{2,4,5\}$ & $(9,\ 28,\ -13)$\\
$\{1,4\}$ & $(2,\ 13,\ -37)$ & $\{2,3,5\}$ & $(99,\ 15,\ -10.5)$\\
$\{1,5\}$ & $(-2,\ 98,\ -37)$ & $\{2,3,4\}$ & $(37,\ -3,\ -14)$\\
$\{2,3\}$ & $(36,\ -8,\ -2)$ & $\{1,4,5\}$ & $(2,\ 98,\ -20)$\\
$\{2,4\}$ & $(23,\ 13,\ -14)$ & $\{1,3,5\}$ & $(-2,\ 13,\ -102)$\\
$\{2,5\}$ & $(19.5,\ 16,\ -10.5)$ & $\{1,3,4\}$ & $(18,\ -14,\ -102)$\\
$\{3,4\}$ & $(25,\ -102,\ -17)$ & $\{1,2,5\}$ & $(9.5,\ 99,\ -10.5)$\\
$\{3,5\}$ & $(-17,\ 0,\ -43)$ & $\{1,2,4\}$ & $(23,\ 13,\ -16)$\\
$\{4,5\}$ & $(-26,\ -14,\ -13)$ & $\{1,2,3\}$ & $(99,\ 99,\ -9.5)$\\
\end{tabular}

\caption{\label{table shattering scaling radius}
The points $x_1,\dots,x_5$ are shattered by the class of $\ell^1$-balls $B_{r(}^{\ell^1}(p)$ with scaled radius $r(p) = \|p\|_1$. For any $I$, the distance from the decision boundary is positive as $\|x_i - P_I\| \leq r(P_I) -1 \leq r(P_I)+1 \leq \|x_j - P_I\|$ if $i\in I$ and $j\notin I$.
}
\end{table}

The construction of norms with highly irregular unit balls builds on the construction of \cite{naszodi2010transversal} of compact convex subsets of $\R^3$ whose translations shatter sets of arbitrary cardinality.

\begin{proof}[Proof of Theorem \ref{theorem vc normed vector spaces}]
{\bf Claim 1.} The proof follows Lemma \ref{lemma closed vs open balls}: If a set is shattered by $\ell^{p_0}$-balls $\{B_r^{p_0}(x) : x\in \R^d\}$, then it can also be shattered `stably', i.e.\ in such a fashion that its points have strictly positive distance from the boundary of the shattering balls. Since a finite set of balls suffices for shattering, there is a margin within which the distance can be perturbed. Since $\|\cdot \|_p \to \|\cdot \|_{p_0}$ locally uniformly and all points lie within a compact set, this establishes that a set which can be shattered by $p$-balls can also be shattered by $p_0$-balls for $p$ sufficiently close to $p_0$. In other words, the VC-dimension of $\B(p,d)$ is lower semi-continuous in $p$.

{\bf Claim 2: Lower bound.} The proof is similar to that of Lemma \ref{lemma flat tori}, exploiting the product structure of the space. We refer the reader to Table \ref{table shattering scaling radius} for an example of a dataset $\{x_1,\dots, x_5\}$ in $\R^3$ which is shattered by $\ell^1$-balls $B_{r(x)}^{\ell^1}(x)$ with $r(x) = \|x\|_1$.

If $d= 3k+1$ or $d=3k+2$ with $k\in\N_0$, we find that $\vcdim(\B(1,d)) \geq \vcdim(\B(1, 3k))$ as we can embed $\R^{3k}$ isometrically in $\R^d$ by setting $x_i=0$ for $3k<i\leq d$. If $d=3k$, we select $n = 5k$ points $x_{i,j}$ with $1\leq i \leq 5$ and $0\leq j\leq k-1$ such that $x_{i,j}$ has a copy of $x_i$ as in Table \ref{table shattering scaling radius} in coordinates $3j,\ 3j+1,\ 3j+2$ and zero entries in all others. For
\[
I \subseteq \{ (i,j) : i=1,\dots,5,\ j=0,\dots,k-1\}\qquad \text{set}\quad I_j = \{i : (i,j)\in I\}.
\]
With the centers $P_J\in\R^3$ given in Table \ref{table shattering scaling radius}, set $Q_I:= (P_{I_0},\dots,P_{I_{k-1}})\in\R^{3k}$, $r_I:= \|Q_I\|_1 = \sum_{j=0}^{k-1}\|P_{I_j}\|_1$ and note that
\[
\|Q_I - x_{i,j}\|_1 = \sum_{l\neq j}\|P_{I_j}\|_1 + \|P_{I_j}- x_i\|_1 = \sum_{l}\|P_I\|_1 + \|P_{I_j}- x_i\|_1 - \|P_{I_j}\|_1 = r_I + \|P_{I_j} - x_i\|_1 - r(P_{I_j})
\]
meaning that
\[
\|Q_I - x_{i,j}\| \leq r_I \qquad\LRa\quad  \|P_{I_j} - x_i\|_1 \leq r(P_{I_j})\qquad\LRa\quad i\in I_j \qquad\LRa\quad (i,j) \in I.
\]
Thus the $5k$ points $x_{ij}$ are shattered by (a sub-class of) $\ell^1$-balls in $\R^{3k}$.

{\bf Claim 2: Upper bound.} Assume that the set $p_1,\dots, p_n$ is shattered. In light of the lower bound, we may assume that $n\geq d+2$. Without loss of generality, we assume that the coordinate entries $p_{1,j},\dots, p_{n,j}$ are distinct for all $j$; otherwise we may perturb the points as they are shattered `stably' by Lemma \ref{lemma closed vs open balls}. We denote by $t_{i,j}$ the ordered `threshold values' $p_{i,j}$ along with $t_{0,j} = -\infty$, $t_{n+1,j} = +\infty$ and divide $\R^d$ into $n+1$ regions according to the $j$-th coordinate
\[
I_j^i = \{ x\in\R^d : t_{i,j} \leq x_i < t_{i+1,j}\}, \qquad i= 0, \dots, n, \qquad j = 1,\dots, d
\]
and set $Q(K) = \bigcap_{j=1}^d I_j^{K_j}$ for $K\in \{0,\dots, n\}^d$. The set of all sign patterns realizable by $\ell^1$-balls with centers $x\in\R^d$ is the union over $K$ of the sign patterns realizable by $\ell^1$-balls with centers in $Q_K$.

As $x$ varies across the domain $Q_K$, we note that
\[
\|x-p_i\|_1 = \sum_{j=1}^d |x_j - p_{i,j}| = \sum_{j=1}^d \sigma_{i,j} x_j + b_i\qquad\text{where }\sigma_{ij} \in\{-1,1\} \quad\text{and } b_i = -\sum_{j=1}^d \sigma_{i,j}p_{i,j}
\]
for $\sigma_{i,j}$ depending only on $K$ and not $x$ since $x_j - p_{i,j}$ does not change sign within $K$. Thus, any pattern of $n$ signs which is achievable on the set of vectors $\{p_1,\dots, p_n\}$ by $\ell^1$-balls with centers in $Q_K$ is also realized by the set of linear classifiers on the vectors $(\sigma_{i,1},\dots, \sigma_{i,d}, b_i) \in\R^{d+1}$. Since the VC-dimension of linear classifiers in $d$ dimensions is $d+1$, the Sauer-Shelah Lemma \cite[lemma 6.10]{shalev2014understanding} shows that at most $(en/(d+1))^{d+1}$ sign patterns can be realized in the second scenario, and thus also in the first scenario. 

Since 
\begin{enumerate}
    \item $p_1,\dots, p_n$ is shattered by $\ell^1$-balls (i.e.\ all $2^n$ sign patterns are realized),
    \item balls with centers in $Q_K$ realize at most $(en/(d+1))^{d+1}$ sign patterns, and
    \item there are $(n+1)^d$ values for $K$,
\end{enumerate}
we find that
\[
2^n \leq (n+1)^d \left(\frac{en}{d+1}\right)^{d+1} \leq \left(\frac{2e\,n^2}{d+1}\right)^{d+1} \quad \Ra \quad n \leq (d+1)\,\log_2\left(\frac{2en^2}{d+1}\right)% = d\log_2\left(\frac{2e}d\right) + 2d\,\log_2(n) 
\leq 2(d+1)\,\log_2(n)
\]
if $2e/(d+1)\leq 1$, i.e.\ if $d\geq 5$. This reduces to $n/\log_2(n) \leq 2(d+1)$. The function $f(s) = s/\log_2 s$ satisfies
\[
f'(s) = \frac{\log_2 s - s\,\frac{1}{s\,\log 2}}{(\log_2s)^2} = \frac{\frac{\log s}{\log 2} - \frac1{\log 2}}{(\log_2 s)^2} = \frac{\log s -1}{\log2\,(\log_2s)^2} \geq 0 \qquad\forall s\geq e.
\]
In particular, abbreviating $D = d+1$, we note that 
\[
f(n) \leq 2D = \frac{4D\,\log_2(4D)}{2\log_2(4D)} = \frac{4D\,\log_2(4D)}{\log_2\big((4D)^2\big)} \leq \frac{4D\,\log_2(4D)}{\log_2\big(4D\,\log_2(4D)\big)} = f\big(4D\,\log_2(4D)\big)
\]
since the logarithm function is both monotone and convex, implying that $\log_2z\leq  1 + z/\log 2$ by the first order concavity condition at $z=1$ and thus that $4D \geq 1+ D/\log 2 \geq \log_2z$ if $D\geq 4$. Overall, we see that
\[
n\leq 4D\,\log_2(4D)  = 4(d+1)\,\log_2(4d+4)\qquad \forall\ d\geq 5.
\]

{\bf Claim 3.} This follows directly from Claims 1 and 2 and Example \ref{example linfty}.

{\bf Claim 4.} This is a special case of \cite[Theorem 3]{naszodi2010transversal}.

{\bf Claim 5.} 
Consider the countable set 
\[
\scr P_{fin}(\N) = \bigcup_{n\in\N} \scr(\{1,\dots,n\}) \subseteq \scr P(\N)
\]
of finite subsets of $\mathbb N$. We select an injective function $\phi:\scr P_{fin}\to (0,1)$ such that 
\begin{enumerate}
    \item $\phi(\mathscr P(\{1,\dots,n\}) \subseteq [1/n, 1)$ and
    \item $n+1\in I \Ra \phi(I) < 1/n$. 
\end{enumerate}
The set $\scr P_{fin}(\N)$ is countable. Let $f(x_1, x_2) = x_1^2 + x_2^2 - 2$. Define
\[
S = \left \{ \left(\phi_J , \frac{1}{n}, f\left(\phi_J, \frac{1}{n}\right) \right) : J \in \scr P_{fin}(\N),\ n \in J \right\}.
\]
Let $C$ be the convex hull of the signed points of $S$,
\[
C: = \conv \left( \left\{ (\sigma_1 x_1, \sigma_2  x_2, \sigma_3  x_3) : x \in S, \,\sigma_i\in\{-1,1\}^3 \right\} \right).
\]
Observe that $C$ is a compact convex set (since $0$ is the only accumulation point of $\phi_J$ and $J$ is finite for $\phi_J\geq \eps>0$) that is symmetric about the origin and contains $0$ in its interior. Thus $C$ is the closed unit ball $\overline{B_1(0)}$ with respect to the norm $\|\cdot\|_C$ induced by Minkowski functional. 

For each $n \in \N$, let $p_n = (0, \frac{1}{n}, \frac{1}{n^2} )$. Next, for each $N \in \N$, let $P_N = \left \{p_n : n \in\{1,\dots, N\} \right\}$. We now prove that the family of closed unit balls with respect to $\|\cdot\|_C$ (i.e.\ translates of $C$) shatter $P_N$, meaning that their VC-dimension is at least $N$. Taking the supremum over $N \in \N$, we find that the VC-dimension of closed unit $\|\cdot\|_C$-balls is infinite.
 
Fix $J \in \scr P(\{1,\dots,n\}) \subseteq \scr P_{fin}(\N)$ and denote $\phi(J)$ by $\phi_J$. Let $\overline{B_J}$ be the translated closed unit ball, 
\[
\overline{B_J} \;\;=\;\; \overline{B_1( \phi_J, 0, f(\phi_J))} \;\;=\;\; \overline{B_1(0)} - \big( \phi_J,\, 0,\, f(\phi_J, 0) \big) \;\;=\;\; C - \big( \phi_J,\, 0,\, f(\phi_J, 0) \big).
\]
Then for each $n \in J$, 
\[ 
p_n = \left(0,\, \frac{1}{n},\, \frac{1}{n^2}\right) = 
\left(\phi_J,\, \frac{1}{n},\, f\left(\phi_J, \frac{1}{n}\right)\right) - \big( \phi_J,\, 0, \,f(\phi_J, 0) \big) \in \overline{B_J}.
\]
Let $m \notin J$. By the strict convexity of $f$ as a function of {\em two} variables $(x,y)$, any point $(x,y,f(x,y))$ lies in the convex hull of points on the graph of $f$ if and only if it is in the closure of that point set. By the locally finite nature of $\phi_J$, this the set coincides with its closure except at $x=0$, so $(\phi_J, 1/m, f(\phi_J), 1/m))$ is {\em not} in $C$, meaning that $(0, 1/m, 1/m^2)\notin \overline B_J$.

If $d> 3$, select any norm $|\cdot|$ on $\R^{d-3}$ and set $\|x\|= \|(x_1, x_2,x_3)\|_C + |(x_4, \dots, x_d)|$. This norm coincides with $\|\cdot \|_C$ on the embedded subspace $V_3:= \R^3\times \{0_{d-3}\}$ and thus has the same shattering capacity for sets $S\subseteq V_3\subseteq\R^d$.

{\bf Claim 6.} The proof is similar to that of Claim 5, but instead, we consider $f(x,y) = \sqrt{1-x^2-y^2}$ and select the points $\phi_J\in(0,\eps)$. As above, we focus on the case $d=3$ and modify $\|x\| = \sqrt{\|x_{1:3}\|_C^2 + \|x_{4:d}\|_2^2}$ if $d>3$. To make the unit ball similar to the Euclidean one, we additionally add the necessary segments:
\[
S =  \left \{ \left(\phi_J , \frac{1}{n}, f\left(\phi_J, \frac{1}{n}\right) \right) : J \in \scr P_{fin}(\N),\ n \in J \right\}, \quad S_{sign} = \{ \sigma\odot x : \sigma\in\{-1,1\}^3, x\in S\}
\]
and 
\[
C = \conv\left( S_{sign} \cup \{x\in B_1^{eucl}(0) : |x_1|>2\eps\}\right).
\]
The circular slices of the Euclidean ball can be identified by scaling up to translation, but not by translation alone as was the case for parabolas. The details are left to the reader.
\end{proof}

\begin{remark}
The argument in the Sauer-Shelah upper bound for the growth function of linear classifiers is sharp by the `counting function theorem' \cite{cover1965geometrical}. On the other hand, it is possible and even likely that $\ell^1$-balls with centers in the different regions in the decomposition of $\R^d$ may give rise to the same sign patterns on a given dataset, leading to a potentially significant overcount of the number of possible sign patterns, and thus an overestimate of the size of a shatterable set. It remains open whether the logarithmic term in the upper bound or the purely linear growth of the lower bound is closer to the true growth rate of the VC-dimension of $\ell^1$-balls as the dimension $d$ grows.
\end{remark}

\begin{table}
\begin{align*}
x_1 &= (10,\ 6,\ 7) &
x_2 &= (11,\ 4,\ -9)&
x_3 &= (3,\ -11,\ 1)&
x_4 &= (-10,\ 3,\ -4)&
x_5 &= (-6,\ -3,\ 4)
\end{align*}

\begin{tabular}{l |l || l |l } {\bf Subset} & {\bf Point} &  {\bf Subset} & {\bf Point}\\ \hline\hline
$\emptyset$ & $(40,\ 40,\ 40)$ & $\{1,2,3,4,5\}$ & $(1,\ 4,\ -1)$\\
\hline
$\{1\}$ & $(5,\ 9,\ 10)$ & $\{2,3,4,5\}$ & $(1,\ -2,\ -7)$\\
$\{2\}$ & $(11,\ 1,\ -17)$ & $\{1,3,4,5\}$ & $(0,\ 2,\ 3)$\\
$\{3\}$ & $(3,\ -21,\ 1)$ & $\{1,2,4,5\}$ & $(0,\ 5,\ -2)$\\
$\{4\}$ & $(-21,\ 3,\ -4)$ & $\{1,2,3,5\}$ & $(8,\ -3,\ 0)$\\
$\{5\}$ & $(-6,\ -4,\ 14)$ & $\{1,2,3,4\}$ & $(3,\ 4,\ -2)$\\
\hline
$\{1,2\}$ & $(11,\ 7,\ -1)$ & $\{3,4,5\}$ & $(-10,\ -10,\ -1)$\\
$\{1,3\}$ & $(10,\ -9,\ 7)$ & $\{2,4,5\}$ & $(-5,\ 1,\ -9)$\\
$\{1,4\}$ & $(0,\ 7,\ -1)$ & $\{2,3,5\}$ & $(2,\ -3,\ -7)$\\
$\{1,5\}$ & $(0,\ 2,\ 9)$ & $\{2,3,4\}$ & $(3,\ 0,\ -6)$\\
$\{2,3\}$ & $(4,\ -6,\ -9)$ & $\{1,4,5\}$ & $(-5,\ 6,\ 5)$\\
$\{2,4\}$ & $(-2,\ 4,\ -10)$ & $\{1,3,5\}$ & $(5,\ -4,\ 7)$\\
$\{2,5\}$ & $(0,\ -3,\ -9)$ & $\{1,3,4\}$ & $(2,\ 7,\ 0)$\\
$\{3,4\}$ & $(-11,\ -11,\ -4)$ & $\{1,2,5\}$ & $(6,\ 4,\ 4)$\\
$\{3,5\}$ & $(-4,\ -12,\ 4)$ & $\{1,2,4\}$ & $(4,\ 6,\ -4)$\\
$\{4,5\}$ & $(-12,\ 1,\ 3)$ & $\{1,2,3\}$ & $(7,\ 0,\ -2)$\\
\end{tabular}

\caption{\label{table shattering fixed radius R3}
The points $x_1,\dots,x_5$ are shattered by the class of $\ell^1$-balls $B_{r}^{\ell^1}(p)$ with fixed radius $r=20.5$. For any $I$, the distance from the decision boundary is positive as $\|x_i - P_I\| \leq20 < 21 \leq \|x_j - P_I\|$ if $i\in I$ and $j\notin I$.
}
\end{table}

\section{VC-dimension: General Riemannian manifolds}
\label{appendix general manifold vc}

In this Appendix, we prove Theorem \ref{theorem no vc bound}. The full proof is spread over several statements of independent interest.

\subsection{High dimension} 

The first version of Theorem \ref{theorem no vc bound} that we will prove addresses the special case of $M$ being a Euclidean space of dimension $d\geq 3$. The general version will be derived from this example by a stability argument (in dimension $\geq 3$) and by passing to a surface in $\R^3$ with the metric constructed in Theorem \ref{thm:decoupling_VCdim_in_R3} for the two-dimensional case.

\begin{theorem} \label{thm:decoupling_VCdim_in_R3}
Let $n \in \mathbb N$ and $d \geq 3$. Consider the $d$-dimensional Riemannian manifold $(\R^d,\, g=I_d)$. 
There exists an infinitely smooth Riemannian metric $\tilde g$ on $\R^d$ which is conformally equivalent to $g$ (with a conformal factor bounded both from above and below) such that the VC-dimension of the hypothesis class of indicator functions on the closed balls of radius 1 with respect to $\tilde g$,\; 
$\B^{\tilde g}_1 = \{ B_1^{\tilde g}(x) : x \in \R^d \}$,\; is at least $n$. 
\end{theorem}

\begin{proof} Recall that in $\R^d$, the VC-dimension of $H^{g} = \{ \one_{B_1^{g}(x)} : x \in \R^d \}$ is $d+1$. Hence the Riemannian metric $g$ is trivially conformal to itself and satisfies the claim for all $n \leq d+1$. Thus, we assume that $n > d+1 \geq 4$. 

Let $\{p_1, ..., p_n\}$ be $n$ evenly spaced points along the unit circle on the $(x_3 = 0)$-plane. One can verify that whenever $j \neq k$,
\[
\Dist{g}{p_j, p_k} = \left|e^{2\pi i \cdot j/n} - e^{2\pi i\cdot k/n }\right| \geq \left|e^{2\pi i/n}-1\right| \geq \left|\sin\left( \frac{2\pi}{n} \right) \right| > \frac{2}{n},
\]
where $i$ denotes $\sqrt{-1}$ in the above equation. Similarly, let $\left\{ P_J : J \in \scr P(\{1, ..., n\}),\; J \neq \emptyset \right\}$ be a set of $2^n-1$ points that lie on the positive $x_3$-axis such that $\Dist{g}{P_J, P_K} > \frac{2}{n}$ whenever $J \neq K$.

Next, consider the indexing set $\cal J = \{(J, j) : J \in \scr P(\{1, ..., n\}),\; J \neq \emptyset,\;  j \in J\}$. For each $(J, j) \in \cal J$, we define $\ell_{Jj}$ to be the line segment from $P_J$ to $p_j$. Let  $L$ denote the union of all such $\ell_{Jj}$. 

Note that if $J \neq K$ and $j \neq k$, then $\ell_{Jj}$ and $\ell_{Kk}$ do not intersect. Otherwise, if $J = K$ or $j = k$, then $\ell_{Jj}$ and $\ell_{Kk}$ intersect exactly at their shared endpoint where either $P_J = P_K$ or $p_j = p_k$, respectively. In other words, if $(J, j) \neq (K, k)$, then outside of a ball of radius $1/n$ at each endpoint, $\ell_{Jj}$ and $\ell_{Kk}$ are disjoint. Hence there exists some $\epsilon \in \left(0, \frac{1}{n}\right)$ such that the closed $\epsilon$-tube $(\ell_{Jj})_\epsilon^g := \{x : \Dist g{x,\ell_{Jj}}\leq\epsilon\}$ about $\ell_{Jj}$ and the $\epsilon$-tube $(\ell_{Kk})_\epsilon^g$ about $\ell_{Kk}$ do not intersect outside of the balls of radius $1/n$ at each end point of $\ell_{Jj}$ and $\ell_{Kk}$. Analogously, we let $\Lambda_\epsilon^g$ denote the union of all such $(\ell_{Jj})_\epsilon^g$. See Figure \ref{Lambda_structure} for a depiction of $\Lambda_\epsilon^g$.

\begin{figure}
    \centering
    \includegraphics[width=0.5\linewidth]{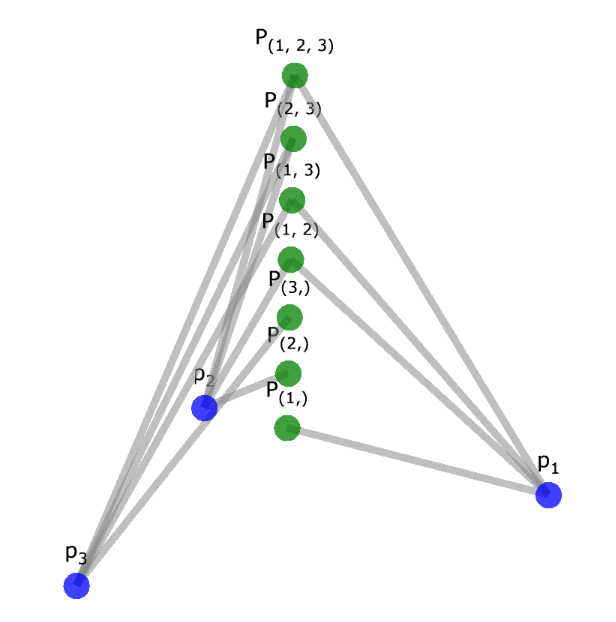}
    \caption{An example of $\Lambda_\epsilon^g$  in \textcolor{darkgray}{gray} for $n = 3$. The balls $B_{1/n}^g(P_J)$ of radius $1/n$ centered at each $P_J$ are in \textcolor{darkgreen}{green} and the balls $B_{1/n}^g(p_j)$ of radius $1/n$ centered at each $p_j$ are in \textcolor{blue}{blue}. Note that the above figure depicts a subgraph of $\Lambda_\epsilon^g$ for $n > 3$.}
    \label{Lambda_structure}
\end{figure}

For each $(J, j) \in \cal J$, we define the Lipschitz continuous function $\Delta_{Jj} : \R^d \to \R$, 
\[
\Delta_{Jj}(x) = \min \left\{1,\; \frac{3}{\epsilon} \cdot \Dist{g}{x, (\ell_{Jj})_{\epsilon/3}^g} \right\}
= \begin{cases}
    0 & \text{if $x \in (\ell_{Jj})_{\epsilon/3}^g$,} \\
    \frac{3}{\epsilon} \cdot \Dist{g}{x, (\ell_{Jj})_{\epsilon/3}^g} & \text{if $x \in (\ell_{Jj})_{2\epsilon/3}^g \setminus (\ell_{Jj})_{\epsilon/3}^g$,} \\
    1 & \text{if $x \notin (\ell_{Jj})_{2\epsilon/3}^g$.}
\end{cases}\]
Let $\delta_{Jj}$ be a smooth approximation to $\Delta_{Jj}$ defined by taking the convolution of $\Delta_{Jj}$ with a standard mollifier $\psi_\eta$  where $\eta \in \left(0, \frac{\epsilon}{12}\right]$ is sufficiently small such that $\delta_{Jj}$ satisfies
\[ 
\begin{cases}
    \delta_{Jj}(x) = 0 & \text{if $x \in (\ell_{Jj})_{\epsilon/4}^g$,} \\
    \delta_{Jj}(x) \in [0, 1] & \text{if $x \in (\ell_{Jj})_{3\epsilon/4}^g \setminus (\ell_{Jj})_{\epsilon/4}^g$,} \\
    \delta_{Jj}(x) = 1 & \text{if $x \notin (\ell_{Jj})_{3\epsilon/4}^g$.}
\end{cases}
\]

Fix some constant $\lambda \in \R$ with $\lambda > (\frac{4}{\epsilon} \cdot \frac{6}{5})^2$, and denote $\arclength{g}{\ell_{Jj}}$ by $\alpha_{Jj}$. Finally, we define the smooth non-vanishing function $f : \R^d \to \R$ by 
\begin{equation}\label{eq:conformal_factor}
    f(x) =  \lambda \cdot \left( \prod_{(J, j) \in \cal J} f_{Jj}(x) \right)
\end{equation}
where each $f_{Jj}$ given by
\[f_{Jj}(x) := \delta_{Jj}(x) \cdot \left(1 - \frac{1}{\lambda \cdot \alpha_{Jj}^2} \right) + \left(\frac{1}{\lambda \cdot \alpha_{Jj}^2}\right).\] 
Thus, the conformal factor $f$ is an infinitely smooth function that is bounded from above and below by positive numbers which induces a Riemannian metric $\tilde g := fg$ that is conformally equivalent to $g$. 

It is left to show that the VC-dimension of $\B^{\tilde g}_1$ is at least $n$. To do so, we must first show that for every non-empty $J \in \scr P(\{1, ..., n\})$, $j \in J$ if and only if $p_j \in B_1^{\tilde g}(P_J)$. Once we have proven the aforementioned claim, we complete the proof by showing that for $J = \emptyset$, all $p_j \notin B_1^{\tilde g}(P_\emptyset)$.
 
\textbf{Claim 1:} Fix $(J, j) \in \cal J$. Then there exists a curve $\gamma$ from $P_J$ to $p_j$ such that $\arclength{\tilde g}{\gamma} \leq 1$. 
    
    \begin{claimproof}
        We begin by proving an upper bound of $f$ on $\ell_{Jj}$. Observe that for all $x \in \ell_{Jj}$, 
        \begin{align} \label{eq:f_on_ell}
            \notag f(x) &= \lambda  \cdot \left( \prod_{(K, k) \in \cal J} f_{Kk}(x) \right)\\
            \notag&= \lambda \cdot f_{Jj}(x) \cdot\left( \prod_{(J, j) \neq (K, k)} f_{Kk}(x) \right) \\ 
            \notag&= \lambda \cdot \left( \frac{1}{\lambda \cdot \alpha_{Jj}^2}\right) \cdot\left( \prod_{(J, j) \neq (K, k)} f_{Kk}(x) \right) & \text{since $\delta_{Jj}(x) = 0$} \\
            \notag&\leq \lambda \cdot \left( \frac{1}{\lambda \cdot \alpha_{Jj}^2}\right) \cdot \left( \prod_{(J, j) \neq (K, k)} 1 \right) & \text{all $f_{Kk}$ are upper bounded by 1}\\ 
            &= \frac{1}{\alpha_{Jj}^2}.
        \end{align}
        Using this upper bound of $f$ on $\ell_{Jj}$, we may upper bound the arclength of $\ell_{Jj}$ with respect to $\tilde g$,
        \begin{align*}
            \arclength{\tilde g}{\ell_{Jj}} &= \int_0^1 \sqrt{f(\ell_{Jj}(t))} \cdot \sqrt{g_{\ell_{Jj}(t)}(\ell_{Jj}'(t), \ell_{Jj}'(t))}\,dt \\
            &\leq \frac{1}{\alpha_{Jj}} \cdot \int_{0}^{1} \sqrt{g_{\ell_{Jj}(t)}(\ell_{Jj}'(t), \ell_{Jj}'(t))}\,dt & \text{by Equation (\ref{eq:f_on_ell})}  \\
            &= \frac{1}{\alpha_{Jj}} \cdot \arclength{g}{\ell_{Jj}} \\
            &= \frac{1}{\alpha_{Jj}} \cdot \alpha_{Jj} = 1. 
        \end{align*}
        Thus, $\gamma := \ell_{Jj}$ satisfies the claim. 
    \end{claimproof} 

\textbf{Claim 2:} Fix some nonempty $J \in \scr P(\{1, ..., n\})$ and $k \notin J$. Let $\gamma$ be a curve from $P_J$ to $p_k$. Suppose, additionally, that $\gamma \subseteq \Lambda_\epsilon^g$. Then $\arclength{\tilde g}{\gamma} > \frac{6}{5}$. 
        
    \begin{claimproof}
        Note that  $L$ can be thought of as a graph whose vertices are given by
        \[\big\{ p_i : i \in \{1, ..., n\} \big\} \cup \big\{ P_I : I \in \scr P (\{1, ..., n\}), I \neq \emptyset \big\}\]
        and whose edges are given by $\{\ell_{Ii} : (I, i) \in \cal J\}$. In particular,  $L$ forms a bipartite graph in which a vertex in $\{ p_i : i \in \{1, ..., n\} \}$ is connected to a vertex in $\{ P_I : I \in \scr P (\{1, ..., n\})\}$ via the edge $\ell_{Ii}$ if and only if $i \in I$.
        
        Hence, because there is no direct edge connecting $P_J$ to $p_k$ since $k \notin J$,  then for any path in  $L$ from $P_J$ to $p_k$, there must exist $j \in J$ and $K \in \scr P(\{1, ..., n\})$ with $k \in K$ such that the path passes from $P_J$ to $p_j$ via the edge $\ell_{Jj}$ and passes from $p_k$ to $P_K$ via the edge $\ell_{Kk}$. For example, if $J = \{1, 2\}$, and $k = 3$, then there is no direct edge from $P_{\{1, 2\}}$ to $p_3$. Instead, to traverse from $P_{\{1, 2\}}$ to $p_3$, one possible path would be to pass from $P_{\{1, 2\}}$ to $p_1$, from $p_1$ to $P_{\{1, 3\}}$, and finally, from $P_{\{1, 3\}}$ to $p_3$. 

         Therefore, because $\gamma$ is a curve from $P_J$ to $p_k$ contained in  $\Lambda_\epsilon^g$, $\gamma$ must travel from $ B_{1/n}^g(P_J)$ to $ B_{1/n}^g(P_j)$ via the $\epsilon$-tube $(\ell_{Jj})_\epsilon^g$ and must travel from $ B_{1/n}^g(p_j)$ to $ B_{1/n}^g(P_K)$ via the $\epsilon$-tube $(\ell_{Kk})_\epsilon^g$. See Figure \ref{Lambda_strucutre_with_gamma} for an example illustration. 

        \begin{figure}
            \centering
        \includegraphics[width=0.5\linewidth]{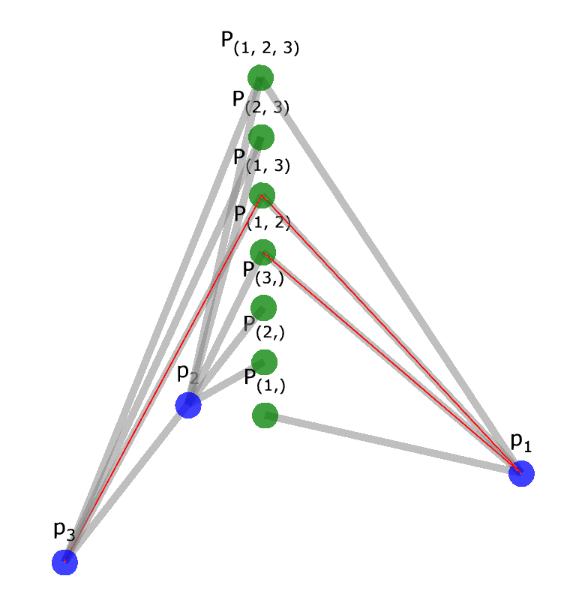}
            \caption{An example curve $\gamma$ (\textcolor{red}{red}) contained in $\Lambda_{\epsilon}^g$ from $P_{\{1, 2\}}$ to $p_3$ which passes from $B_{1/n}^g\big(P_{\{1, 2\}}\big)$ to $B_{1/n}^g(p_1)$ along the $\epsilon$-tube $\big(\ell_{\{1, 2\},\,1}\big)$ and from  $B_{1/n}^g\big(P_{\{1, 3\}}\big)$ to $B_{1/n}^g(p_3)$ along the $\epsilon$-tube $\big(\ell_{\{1, 3\},\,3}\big)$.}
        \label{Lambda_strucutre_with_gamma}
        \end{figure}
        
        In particular, we define
        \begin{align*}
            t_0 &:= \max \left\{ t \in [0, 1] : \gamma(t) \in B_{1/n}^g(P_J) \right\} \\
            t_1 &:= \min \left\{ t \in [t_0, 1] : \gamma(t) \in B_{1/n}^g(p_j) \right\} \\
            \tau_0 &:= \max \left\{ \tau \in [t_1, 1] : \gamma(\tau) \in B_{1/n}^g(P_K) \right\} \\
            \tau_1 &:= \min \left\{ \tau \in [\tau_0, 1] : \gamma(\tau) \in B_{1/n}^g(p_k)\right\}
        \end{align*}
        so that 
        \begin{align*}
            \gamma(t) \in (\ell_{Jj})^g_{\eps} \setminus \left(B_{1/n}^g(P_J) \cup B_{1/n}^g(p_j)\right) & \text{ for all $t \in [t_0, t_1]$}, \\
            \gamma(\tau) \in (\ell_{Jj})^g_{\eps} \setminus \left(B_{1/n}^g(P_K) \cup B_{1/n}^g(p_k)\right) & \text{ for all $\tau \in [\tau_0, \tau_1]$}.
        \end{align*}
        By definition of $f_{Ii}$, we have that $f_{Ii}(x) = 1$ for all $x$ such that $\Dist{g}{x, \ell_{Ii}} > \frac{3\epsilon}{4}$. Furthermore, $f_{Jj}(x) = \frac{1}{\lambda \cdot \alpha_{Jj}^2}$ for $x \in (\ell_{Jj})_{\epsilon/4}^g$ and $f_{Kk}(x) = \frac{1}{\lambda \cdot \alpha_{Kk}^2}$ for $x \in (\ell_{Kk})_{\epsilon/4}^g$. 
        Lastly, because $(\ell_{Ii})_{\epsilon}^g$ do not intersect outside of the balls of radius $\frac{1}{n}$ based at the end points of $\ell_{Ii}$, then $f(\gamma(t)) \geq \frac{1}{\alpha_{Jj}^2}$ whenever $t \in [t_0, t_1]$ and $f(\gamma(\tau)) \geq \frac{1}{\alpha_{Jj}^2}$ whenever $\tau \in [\tau_0, \tau_1]$. Hence we have that
            \begin{align*}
                \arclength{\tilde g}{\gamma} 
                &= \int_0^1 \sqrt{f(\gamma(t))} \cdot \sqrt{g_{\gamma(t)}(\gamma'(t), \gamma'(t))}\,dt \\
                &\geq \int_{t_0}^{t_1} \sqrt{f(\gamma(t))} \cdot \sqrt{g_{\gamma(t)}(\gamma'(t), \gamma'(t))}\,dt + \int_{\tau_0}^{\tau_1} \sqrt{f(\gamma(\tau))} \cdot \sqrt{g_{\gamma(\tau)}(\gamma'(\tau), \gamma'(\tau))}\,d\tau \\
                &\geq \frac{1}{\alpha_{Jj}} \cdot \int_{t_0}^{t_1}  \sqrt{g_{\gamma(t)}(\gamma'(t), \gamma'(t))}\,dt + \frac{1}{\alpha_{Kk}} \cdot \int_{\tau_0}^{\tau_1}  \sqrt{g_{\gamma(\tau)}(\gamma'(\tau), \gamma'(\tau))}\,d\tau \\
                &\geq \frac{1}{\alpha_{Jj}} \cdot \Dist{g}{B_{1/n}^g(P_J), B_{1/n}^g(p_j)} + \frac{1}{\alpha_{Kk}} \cdot \Dist{g}{B_{1/n}^g(P_K), B_{1/n}^g(p_k)} \\
                &= \frac{1}{\alpha_{Jj}} \cdot \left(\alpha_{Jj} - \frac{2}{n}\right) + \frac{1}{\alpha_{Kk}} \cdot \left(\alpha_{Kk} - \frac{2}{n}\right) \\
                &= 2 - \frac{2}{\alpha_{Jj} \cdot n} - \frac{2}{\alpha_{Kk} \cdot n}  
            \end{align*}
            Lastly, note that because $\ell_{Jj}$ and $\ell_{Kk}$ are line segments connecting a point on the unit circle in the $(x_3 = 0)$-plane to a point on the positive $x_3$-axis, we have that $\alpha_{Jj}$, $\alpha_{Kk} > 1$. Thus, because $n \geq 5$ and $\alpha_{Jj}$, $\alpha_{Kk} > 1$,
            \[
            \arclength{\tilde g}{\gamma} \geq \left(2 - \frac{2}{\alpha_{Jj} \cdot n} - \frac{2}{\alpha_{Kk} \cdot n} \right)  > \left(2 - \frac{2}{5} - \frac{2}{5} \right) = \frac{6}{5}.
            \]
    \end{claimproof} 

Recall that in the above claim, we considered the case in which $\gamma \subseteq \Lambda_{\epsilon}^g$. We now show that any curve that leaves $\Lambda_{\epsilon}^g$ will also have a sufficiently large arclength. The explanation for this claim can be intuitively described by the idea that we have constructed $f$ to be sufficiently large outside of $\Lambda_{3\epsilon/4}^g$ so that if a curve $\gamma$ travels from inside of $\Lambda_{3\epsilon/4}^g$ to outside $\Lambda_{\epsilon}^g$, then the arclength of the subcurve is already at least $6/5$ with respect to the constructed Riemannian metric $\tilde g = fg$. 

\textbf{Claim 3:} Let $\gamma$ be a curve in $\R^d$. Suppose that there exist two points $\gamma(t_0)$ and $\gamma(t_1)$ on the curve such that $\gamma(t) \notin \Lambda_{3\eps/4}^g$ for all $t \in (t_0, t_1)$ and $\Dist{g}{\gamma(t_0), \gamma(t_1)} > \frac{\epsilon}{4}$. Then $\arclength{\tilde g}{\gamma} > \frac{6}{5}$.

\begin{claimproof}
     First, observe that whenever $x \notin \Lambda_{3\epsilon/4}^g$, we have $\delta_{Jj}(x) = 1$ for all $(J, j) \in \cal J$, so
        \begin{equation} \label{eq:f_outside_L}
            f(x) = \lambda \cdot \left( \prod_{(J, j) \in \cal J} f_{Jj}(x) \right) = \lambda \cdot \left( \prod_{(J, j) \in \cal J} 1 \right) = \lambda.
        \end{equation}
    Therefore,
        \begin{align*}
            \arclength{\tilde g}{\gamma} &= \int_0^1 \sqrt{f(\gamma(t))} \cdot \sqrt{g_{\gamma(t)}(\gamma'(t), \gamma'(t))}\,dt \\
            &\geq \int_{t_0}^{t_1} \sqrt{f(\gamma(t))} \cdot \sqrt{g_{\gamma(t)}(\gamma'(t), \gamma'(t))}\,dt \\
            &= \sqrt{\lambda} \cdot \int_{t_0}^{t_1} \sqrt{g_{\gamma(t)}(\gamma'(t), \gamma'(t))}\,dt & \text{by Equation (\ref{eq:f_outside_L})}  \\
            &\geq \sqrt{\lambda} \cdot \Dist{g}{\gamma(t_0), \gamma(t_1)} \\
            &> \sqrt{\lambda} \cdot \frac{\epsilon}{4}.
        \end{align*}
        Thus, because we picked $\lambda$ to be greater than $\left(\frac{4}{\epsilon} \cdot \frac{6}{5}\right)^2$, then we have that $\arclength{\tilde g}{\gamma} > \frac{6}{5}$.
\end{claimproof}

Let $\gamma$ be a curve from $P_J$ to $p_k$ with $k \notin J$. Now suppose that $\gamma \not \subseteq \Lambda_{\epsilon}^g$. Then there exists some $\gamma(t_0) \notin \Lambda_\epsilon^g$, and because $\gamma$ ends at $\gamma(1) = p_k \in \Lambda_{3\epsilon/4}^g$, there exists some $t_1$ such that $\gamma(t_1)$ is the first entry point into  $\Lambda_{3\epsilon/4}^g$ after $\gamma(t_0)$. In other words, there exists $t_1 := \min \left\{ t \in [t_0, 1] : \gamma(t) \in \scr  \Lambda_{3\epsilon/4}^g \right\} $
such that $\gamma(t) \notin \scr  \Lambda_{3\epsilon/4}^g$ for all $t \in (t_0, t_1)$. Hence we satisfy the conditions of Claim 3, and so $\arclength{\tilde g}{\gamma} > \frac{6}{5}$. 

Therefore, for a given nonempty $J \in \scr P(\{1, ..., n\})$, if $j \in J$, then $\Dist{\tilde g}{P_J, p_j} \leq 1$ by Claim 1. Moreover, combining Claim 2 and Claim 3, we have shown that if $k \notin J$, then any curve $\gamma$ from $P_J$ to $p_k$ has $\arclength{\tilde g}{\gamma} > \frac{6}{5}$. 
Hence $\Dist{\tilde g}{P_J, p_k} \geq \frac{6}{5} > 1$. \\
\hspace*{5mm} Lastly, we must consider the case where $J = \emptyset$. Let $P_\emptyset \in \R^d$ such that $\Dist{g}{P_\emptyset, \Lambda_{3\epsilon/4}^g} > \frac{\epsilon}{4}$. Then $P_\emptyset$ satisfies the conditions of Claim 3, and we have that $\Dist{\tilde g}{P_\emptyset, p_j} \geq \frac{6}{5} > 1$ for all $j \in \{1, ..., n\}$. Hence all $p_j \notin B_1^{\tilde g}(P_\emptyset)$. \\
\hspace*{5mm} Therefore, for all $J \in \scr P(\{1, ..., n\})$, the predictor $B_1^{\tilde g}(P_J) \in \B^{\tilde g}_1$ is 1 at $p_j$ for $j \in J$ and 0 at $p_k$ if $k \notin J$. Thus, the VC-dimension of $\B^{\tilde g}_1$ is at least $n$. \qedhere
\end{proof}

\begin{remark}
    The proof of Theorem \ref{thm:decoupling_VCdim_in_R3} also reveals that the VC-dimension of the hypothesis class of indicator functions on the \textit{open} balls of $(6/5)$-radius with respect to $\tilde g$ is at least $n$. Additionally, the proof of Theorem \ref{thm:decoupling_VCdim_in_R3} does not depend on the radius being 1 and may be rescaled so that the claim applies for hypothesis classes of metric balls with any fixed radius. 
\end{remark}

\begin{corollary}   
Let $n \in \mathbb{N}$ and $(M, h)$ be a Riemannian manifold of dimension $d \geq 3$. Then there exists a Riemannian metric $\tilde h$ on $M$ which is conformally equivalent to $h$ such that the VC-dimension of the hypothesis class of indicator functions on the closed balls of radius 1 with respect to $\tilde h$,\; $\B^{\tilde h}_1 = \{ B_1^{\tilde h}(m) : m \in M \}$,\; is at least $n$. 
\end{corollary}

\begin{proof}
Fix some $m_0 \in M$. Let $\phi : U \subseteq M \to \R^d$ be the normal coordinates on an open neighborhood $U$ of $m_0$ such that $\phi$ is a smooth map with $\phi(m_0) = 0$. Restrict $\phi$ to an open ball $B^h_{r_0}(m_0) \subseteq U$ with sufficiently small radius $r_0 > 0$ such that inside of this ball, $g$ is locally equivalent to $h$ via $\phi$ in the sense that there exists some $\xi \in (0, \frac{1}{6})$ such that
\begin{equation} \label{eq:g_is_equivalent_to_h}
(1 - \xi) \cdot \big|\big| \phi \big(\gamma(t)\big)' \big|\big|_g  \leq \big|\big| \gamma'(t) \big|\big|_h \leq \frac{6}{5} \cdot \left(1 - \xi \right) \cdot \big|\big| \phi\big(\gamma(t)\big)'\big|\big|_g,
\end{equation}
for all curves $\gamma$ in $B^h_{r_0}(m_0)$ and $t \in [0, 1]$. 

As $\phi$ is the normal coordinate map about $m_0$, there exists some $r > 0$ such that $\phi^{-1}\big(B_r^g(0)\big) \subseteq B_{r_0}^h(m_0)$. Rescale the arguments in the proof of Theorem \ref{thm:decoupling_VCdim_in_R3} by some constant factor $c$ such that $\Lambda_{\epsilon}^g \subseteq B_r^g(0)$ and rescale the conformal factor $\sqrt{f}$ defined in the proof of Theorem \ref{thm:decoupling_VCdim_in_R3} by $\left(c \cdot \frac{6}{5} \cdot (1 - \xi)\right)^{-1}$ such that for all $J \in \scr P(\{1, ..., n\})$,
\begin{align} 
    &\Dist{\tilde g}{P_J, p_j} \leq \left(\frac{6}{5} \cdot (1 - \xi)\right)^{-1} & \text{if } j \in J, \label{eq:scaled_tilde_g1}\\
    &\Dist{\tilde g}{P_J, p_k} > \frac{6}{5} \cdot \left(\frac{6}{5} \cdot (1 - \xi)\right)^{-1} = (1 - \xi)^{-1} & \text{if } k \notin J. \label{eq:scaled_tilde_g2}
\end{align}
Secondly, we use the rescaled, positive, smooth function $f : \R^d \to \R$ constructed in the proof of Theorem \ref{thm:decoupling_VCdim_in_R3} to define a conformally equivalent Riemannian metric $\tilde h := (f \circ \phi) \cdot h$ on $M$. We now aim to prove the claim that VC-dimension of $\B^{\tilde h}_1$ is at least $n$ by proving  that $p_j \in B^{\tilde h}_{1}(P_J)$ if and only if $j \in J$ for all $J \in \scr P(\{1, ..., n\})$. Equivalently, we aim to show that
\begin{center}
    $\Dist{\tilde h}{\phi^{-1}(P_J), \phi^{-1}(p_j)} \leq 1$ if and only if $j \in J$ for all $J \in \scr P(\{1, ..., n\})$.
\end{center}

Let $\gamma$ be a curve in $B_r^h(m_0)$, let $\Gamma = \phi \circ \gamma$, and let $t \in [0, 1]$. Then by Equation (\ref{eq:g_is_equivalent_to_h}), 
\[ (1 - \xi) \cdot \big|\big| \Gamma'(t) \big|\big|_g \leq \big|\big| \gamma'(t) \big|\big|_h \leq \frac{6}{5} \cdot \left(1 - \xi \right) \cdot \big|\big| \Gamma'(t)\big|\big|_g,\]
\[(1 - \xi) \cdot \sqrt{g_{\Gamma(t)}\big( \Gamma'(t), \Gamma'(t)\big)} \leq \sqrt{h_{\gamma(t)}\big( \gamma'(t), \gamma'(t)\big)} \leq  \frac{6}{5} \cdot \left(1 - \xi \right) \cdot \sqrt{g_{\Gamma(t)}\big( \Gamma'(t),\Gamma'(t)\big)}. \]
Next, we multiply by $\sqrt{(f \circ \phi \circ \gamma)(t)} > 0$ to get
\[(1 - \xi) \cdot \sqrt{f\big(\Gamma(t)\big) \cdot g_{\Gamma(t)}\big( \Gamma'(t), \Gamma'(t)\big)}  \leq \sqrt{(f \circ \phi)\big(\gamma(t)\big) \cdot h_{\gamma(t)}\big( \gamma'(t), \gamma'(t)\big)} \leq  \frac{6}{5} \cdot \left(1 - \xi \right) \cdot \sqrt{f\big(\Gamma(t)\big) \cdot g\big( \Gamma'(t), \Gamma'(t)\big)},\]
\[ (1 - \xi) \cdot \sqrt{\tilde g_{\Gamma(t)}\big( \Gamma'(t), \Gamma'(t)\big) } \leq \sqrt{\tilde h_{\gamma(t)}\big( \gamma'(t), \gamma'(t)\big) } \leq  \frac{6}{5} \cdot \left(1 - \xi \right) \cdot \sqrt{\tilde g\big( \Gamma'(t), \Gamma'(t)\big) }.\]
As the above inequality holds for all $t \in [0, 1]$, then
\[ (1 - \xi) \cdot \int_0^1 \sqrt{\tilde g_{\Gamma(t)}\big( \Gamma'(t), \Gamma'(t)\big) }\,dt \leq \int_0^1 \sqrt{\tilde h_{\gamma(t)}\big( \gamma'(t), \gamma'(t)\big) }\,dt \leq  \frac{6}{5} \cdot \left(1 - \xi \right) \cdot \int_0^1 \sqrt{\tilde g\big( \Gamma'(t), \Gamma'(t)\big) }\,dt\]
\[ (1 - \xi) \cdot \arclength{\tilde g}{\Gamma} \leq  \arclength{\tilde h}{\gamma} \leq  \frac{6}{5} \cdot \left(1 - \xi \right) \cdot  \arclength{\tilde g}{\Gamma}\]
Lastly, because the above inequality holds for all curves $\gamma$ in $B_r^h(m_0)$ and because infimums respect ordering,
\[(1 - \xi) \cdot \Dist{\tilde g}{\phi(x), \phi(y)} 
\leq \Dist{\tilde h}{x, y} \leq  \frac{6}{5} \cdot \left(1 - \xi \right) \cdot \Dist{\tilde g}{\phi(x), \phi(y)}.\]

Therefore, for all $J \in \scr P(\{1, ..., n\})$, if $j \in J$ and $k \notin J$, then by Equations (\ref{eq:scaled_tilde_g1}) and (\ref{eq:scaled_tilde_g2}),
\begin{align*}
    \Dist{\tilde h}{\phi^{-1}(P_J), \phi^{-1}(p_j)}  &\leq  \frac{6}{5} \cdot \left(1 - \xi \right) \cdot \Dist{\tilde g}{P_J, p_j} \leq  1, \\
    \Dist{\tilde h}{\phi^{-1}(P_J),\phi^{-1}(p_k)} &\geq (1 - \xi) \cdot \Dist{\tilde g}{P_J, p_k} > 1 .
\end{align*}
Thus, $p_j \in B^{\tilde h}_{1}(P_J) $ if and only if $j \in J$. Hence the VC-dimension of $\B^{\tilde h}_1$ is at least $n$.

Note that the argument above assumed that $\gamma \subseteq B_r^h(m_0)$. However, if $\gamma \not \subseteq B_r^h(m_0)$, then $\gamma \not \subseteq \Lambda_\epsilon^g$. Hence it follows from Claim 3 in the proof Theorem \ref{thm:decoupling_VCdim_in_R3} that 
\[\arclength{\tilde h}{\gamma} \geq (1 - \xi) \cdot \arclength{\tilde g}{\Gamma} > (1 - \xi) \cdot \frac{6}{5} \geq 1.\]
Hence the above argument holds without loss of generality. 
\end{proof}

We can now prove the claims of infinite VC-dimension in Theorem \ref{theorem no vc bound}.

\begin{proof}[Proof of third claim in Theorem \ref{theorem no vc bound}]
{\bf Third claim.} Let $M$ be a non-compact manifold of dimension $d\geq 3$ and $g$ a Riemannian metric on $M$. Let $x_n$ be a sequence of points in $M$ which has no convergent subsequence. Then $d_i:= \inf_{j\neq i} d_g(x_i, x_j)>0$ for any $i\in\N$. By construction, the $g$-balls 
\[
B^g_i := \{x\in M : d_g(x, x_i) < d_i/2\}
\]
are disjoint. As above, we can select a smooth conformal factor $u^2$ on $M$ which takes the value $1$ outside $\bigcup_{i=1}^\infty B^g_i$ such that 
\[
\dist_{u^2g}(x_i, \partial B^g_i) \geq 2.
\]
Thus, the local analysis of Theorem \ref{thm:decoupling_VCdim_in_R3} can be applied and we can locally modify $u^2 g$ conformally in each of the disjoint balls $B^g_i$ to create a metric whose balls of radius 1 with centers in $B^g_i$ have VC-dimension at least $i$. As the modifications are purely local on distant sets, no smoothness is lost throughout the process. Globally, the modifications yield a metric whose balls of radius 1 have {\em infinite} VC-dimension.
\end{proof}

\subsection{Surfaces}

In this section, we construct a surface $M$ in $(\R^3, g)$ where $g = I_3$. With suitable care, the same VC-dimension lower bounds also apply on this sub-manifold.

\begin{theorem}
    Let $n\in \mathbb N$. Then there exists a two-dimensional Riemannian manifold $(M, \tilde g)$ such that the hypothesis class $\B_1^g = \{ B_1^g(x) : x\in M\}$ has VC-dimension at least $n$. 
\end{theorem}

\begin{proof}
If the VC-dimension of a hypothesis class $\B_1^{\tilde g}$ is at least $n$ for a particular Riemannian manifold $(M, \tilde g)$, then the VC-dimension of that hypothesis class is at least $k$ for all $k \leq n$. Hence we suppose without loss of generality that $n \geq 40$. 

Using the notation defined in the proof of Theorem \ref{thm:decoupling_VCdim_in_R3}, consider the three-dimensional solid
\[  
    S := \bigcup_{(J, j) \in \cal J} \Bigg( (\ell_{Jj})_{\epsilon/8}^g \cup B_{1/n}^g(P_J) \cup B_{1/n}^g(p_j) \Bigg).
\]

Note that $P_\emptyset$ is not contained in this solid by definition of $\cal J$. Hence we define $P_\emptyset$ to be $(0, 0, -2)$ on the $x_3$-axis, and redefine $\cal J$ to be
\[\cal J = \big\{(J, j) \;:\; J \in \scr P(\{1, ..., n\}),\; J \neq \emptyset,\; j \in J\big\} \;\cup\; \big\{(\emptyset, 1)\big\}.\]
Through an abuse of notation, we now consider the statement ``\,all $j \in J$'' to include the case in which $(J, j) = (\emptyset, 1)$ unless explicitly stated that $J$ is non-empty. Lastly, by extension, redefine $S$ with respect to the new definition of $\cal J$. \\
\hspace*{5mm} To prove the claim, we will follow a similar argument to the one presented in the proof of Theorem \ref{thm:decoupling_VCdim_in_R3}. In order to do so, we first need to construct a two-dimensional smooth sub-manifold $M$ in $\R^3$ which approximates the boundary of $S$. 

While there are many methods to implicitly construct a smooth surface, we will use the following explicit construction to more easily bound the necessary arclengths of curves on $M$ later in the proof. \\
\hspace*{5mm} Observe that for each fixed $J \in \scr P(\{1, ..., n\})$, there are $\max\{\,|J|, 1\}$ solid tubes $(\ell_{Jj})^g_{\epsilon/8}$, $j \in J$, protruding from the ball $B_{1/n}^g(P_J)$ for which each transition along the boundary of the ball to the boundary of the tube is not smooth. Hence our goal is to replace each $B_{1/n}^g(P_J)$ with a smoothly flattened ball $B_J$. Similarly, for $j \in\{1, ..., n\}$, we aim to replace $B_{1/n}^g(p_j)$ with a flattened ball $B_j$. Furthermore, we will replace each tube $(\ell_{Jj})^g_{\epsilon/4}$ with $L_{Jj} \cup L_{jJ}$ 
where $L_{Jj}$ and $L_{jJ}$ make up two halves of a flat-bottom cylinder. Thus, by smoothly flattening parts of each ball and smoothly flattening the ends of cylinders, we may glue these solids together to construct the desired solid with smooth boundary. See Figure \ref{fig:deconstruct_M}. 

\begin{figure}
    \centering
    \includegraphics[width=0.6\linewidth]{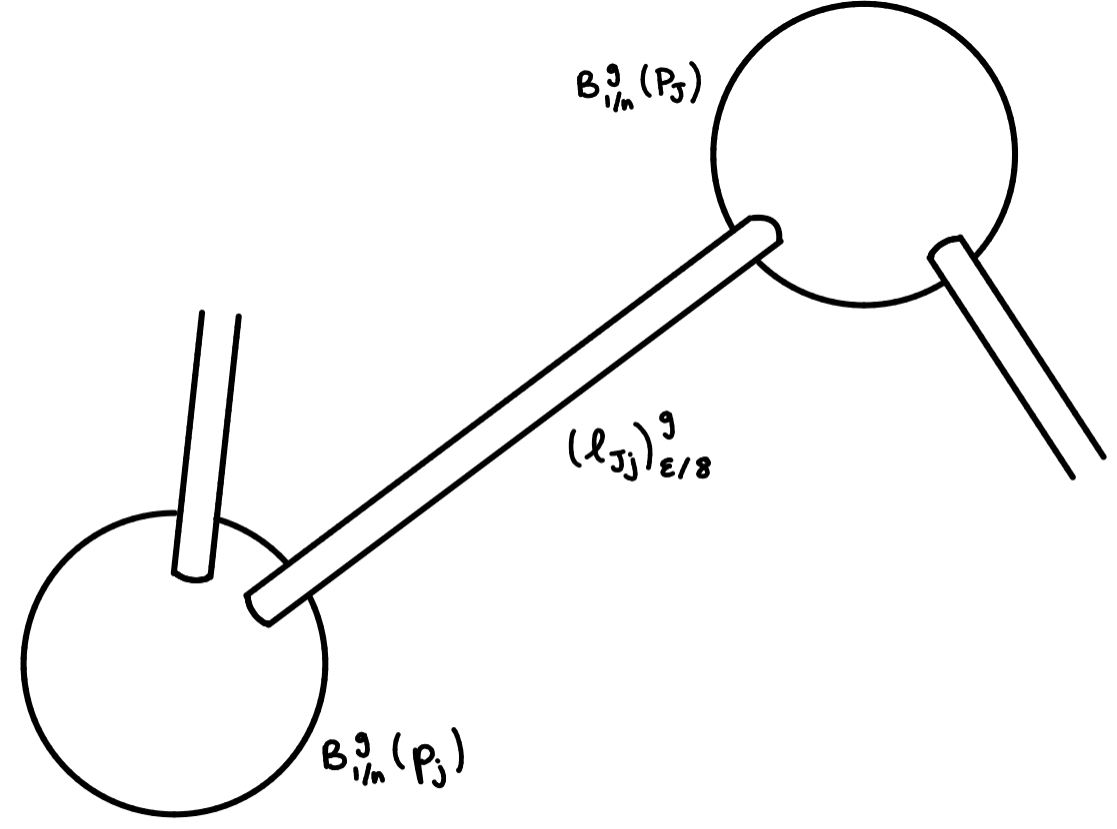}
    \includegraphics[width=0.6\linewidth]{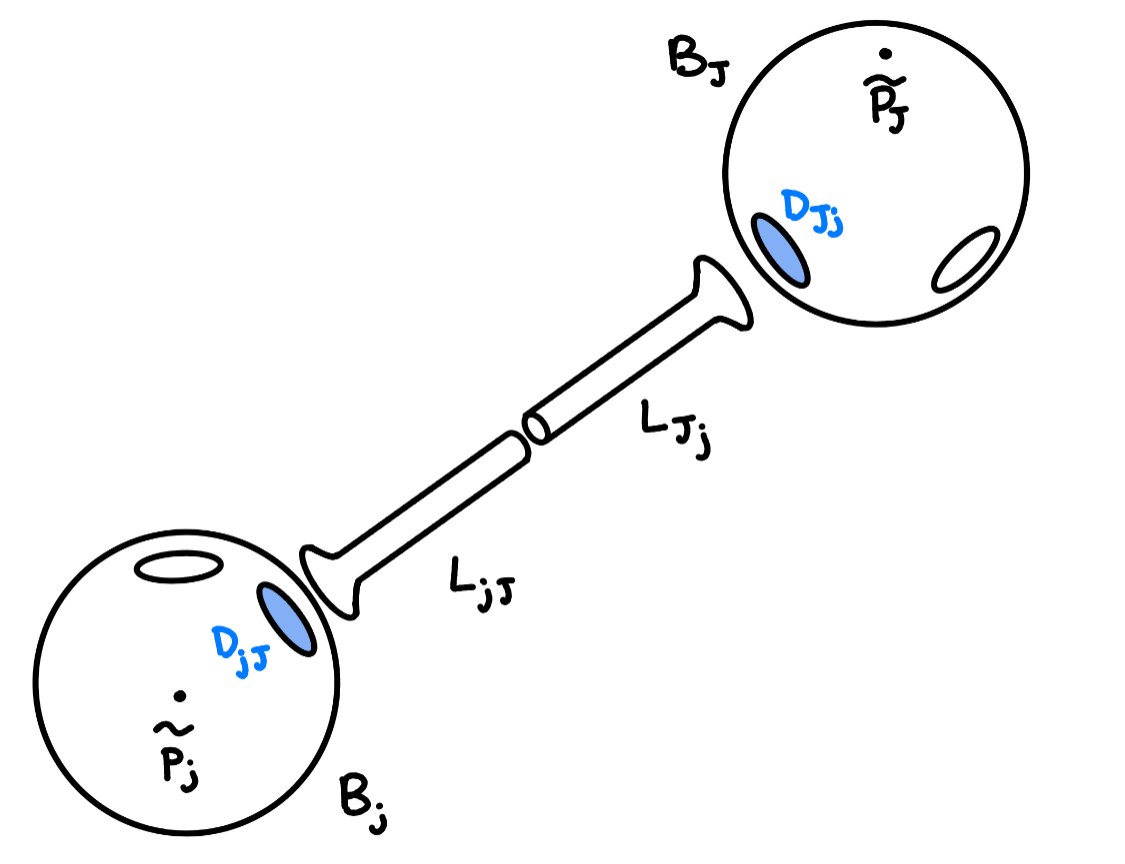}
    \caption{[Top] Components $B_{1/n}^g(P_J)$, $B_{1/n}^g(p_j)$, and $(\ell_{Jj})^g_{\epsilon/8}$ of $S$.   [Bottom] The approximation solids of the above components with smooth-boundary $B_J$, $B_j$, $L_{Jj} \cup L_{jJ}$\,, respectively.}
    \label{fig:deconstruct_M}
\end{figure}

Let us define $B_J$. For each $j \in J$, let $\partial B_{Jj} \subseteq B_{1/n}^g(0)$ be the smooth, convex, closed ``flattened sphere'' as described in \cite[Lemma 9]{wojtowytsch2016helfrich} for which 
\begin{align*}
    &\partial B_{Jj} \cap [D_{\epsilon/4} \times (0, \infty)] = 
    \left \{x_3 = \frac{1}{n} - \frac{3\epsilon}{8n} \right \} \cap [D_{\epsilon/4} \times (0, \infty)], \\
    &\partial  B_J \setminus [D_{\epsilon/2} \times (0, \infty)] = \left(\frac{1}{n}S^2\right) \setminus [D_{\epsilon/2} \times (0, \infty)]. 
\end{align*}
Next, translate $\partial B_{Jj} \subseteq B_{1/n}^g(0)$ so that $\partial B_{Jj} \subseteq B_{1/n}^g(P_J)$, and rotate $\partial B_{Jj}$ so that $\ell_{Jj}$ meets the center of flattened region of the ball perpendicularly. \\
\hspace*{5mm} Recall that $\epsilon$ is chosen such that the finite set of tubes $\{(\ell_{Jj})^g_{\epsilon}\}_{j \in J}$ have pairwise empty intersection outside of $B_{1/n}(P_J)$. Therefore, because the transition from the sphere to the flattened disks on $\partial B_{Jj}$ fits within a radius of $\frac{\eps}{2}$ about $\ell_{Jj}$, the transition regions of $\partial B_{Jj}$ and $\partial B_{Jk}$ do not overlap for all $j \neq k$. Thus, if we let $B_{Jj}$ be the solid enclosed by $\partial B_{Jj}$, then 
\[B_J := \bigcap_{j \in J} B_{Jj}\] is a convex solid whose boundary is smooth. This completes our construction of $B_J \subseteq B_{1/n}^g(P_J)$. For $j \in \{1, ..., n\}$, we use an analogous argument to construct the convex solid
\[B_j := \bigcap_{\underset{j \in J}{J \in \scr P(\{1, ..., n\}}} B_{jJ}\] 
with smooth boundary. Let $D_{Jj}$ denote each corresponding flattened region of radius $\frac{\epsilon}{4}$ on $B_J$ and analogously let $D_{jJ}$ denote each corresponding flattened region on $B_j$. 

Next, we construct $L_{Jj}$. Consider the function $V:\R \to \R$ given by $V(t) = |t|+\frac{\sqrt{2}}{8}\epsilon$. As $V$ is a piece-wise linear function and the standard mollifier $\psi_\eta$ is an even function, there exists sufficiently small $\eta > 0$ such that the mollification $U(t) = (V \ast \psi_\eta)(t)$ satisfies
\[\begin{cases}
    U(t) = V(t) & \text{if } |t| \geq \frac{\epsilon}{16} \\
    U(t) \in \left(0, \frac{\epsilon}{16} + \frac{\sqrt{2}}{8}\epsilon\right] & \text{if } |t| \leq \frac{\epsilon}{16}. 
\end{cases}\]
Let $U_{Jj}$ be the restriction of $U$ to the interval $[t_0, t_1]$ such that the distance from $U_{Jj}(t_0)$ to $(0, \frac{\sqrt{2}}{8}\epsilon)$ is $\frac{1}{2}(\alpha_{Jj} - \frac{2}{n} + \frac{3\epsilon}{4n})$ and such that the distance from $U_{Jj}(t_1)$ to the axis of rotation $y(t) = -t$ is $\frac{\epsilon}{4}$. Let $L_{Jj}$ be the solid of revolution created by rotating the region between $U_{Jj}$ and the axis of rotation $y(t) = -t$.

\begin{figure}%
    \centering
    \subfloat{{\includegraphics[width=0.55\linewidth]{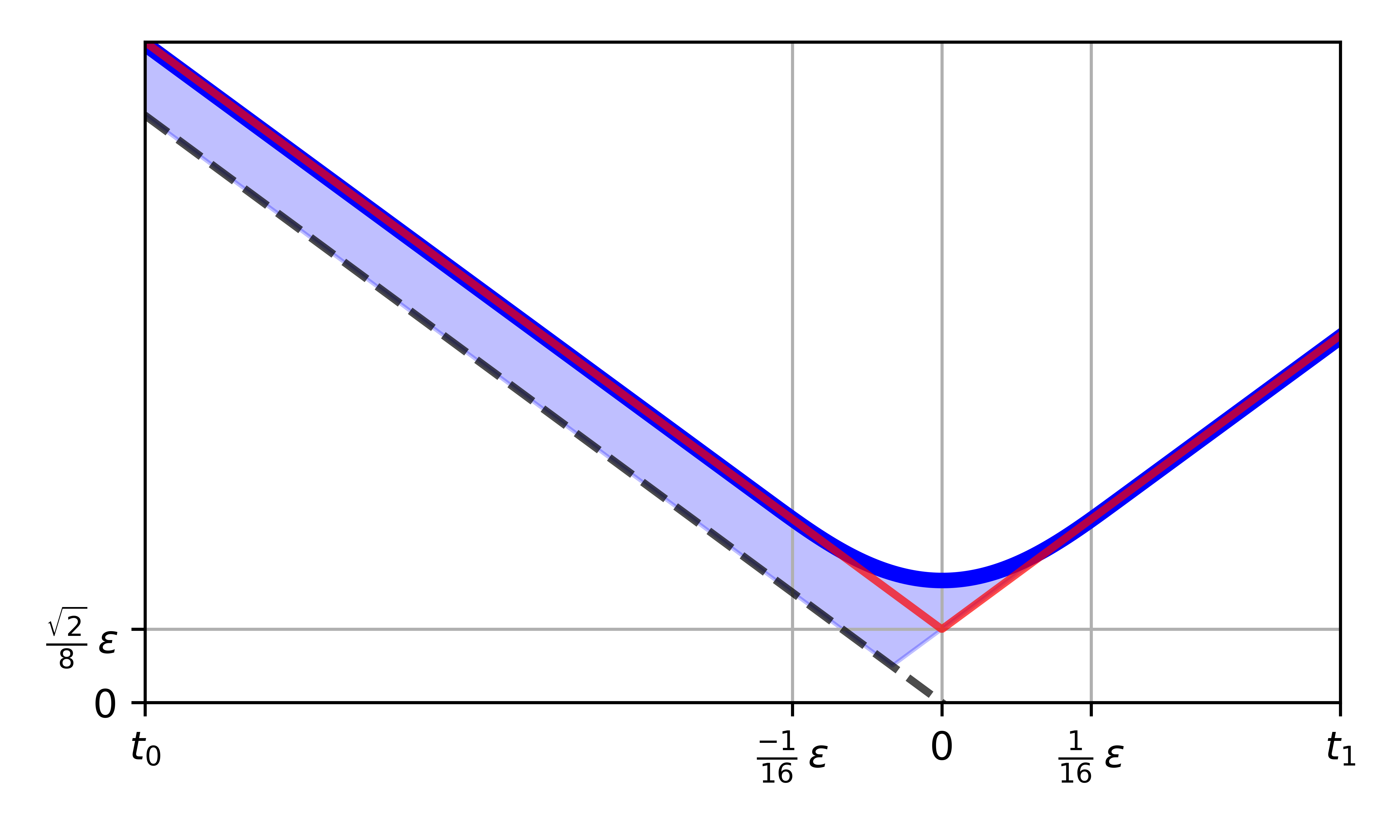} }}%
    \subfloat{{\includegraphics[width=0.45\linewidth]{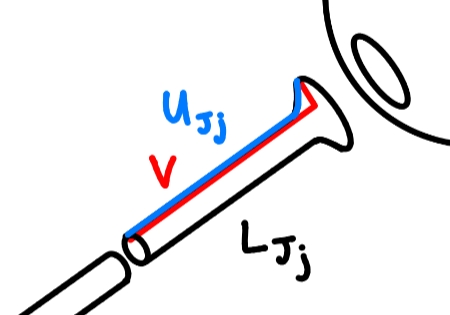} }}%
    \caption{[Left] The function $V(t)$ in \blue{blue}, its mollification $U_{Jj}(t)$ in \red{red}, the dashed axis of rotation $y(t) = -t$. [Right]  Another view of the role that $V$ and $U_{Jj}$ play in the construction of $L_{Jj}$.}
    \label{fig:U}%
\end{figure}

One can verify that for all $t \in \big[t_0, -\frac{\epsilon}{16}\big]$, the distance from $(t, U_{Jj}(t)) = \big(t, -t + \frac{\sqrt{2}}{8}\epsilon \big)$ to the axis of rotation $y(t) = -t$ is exactly $\frac{\epsilon}{8}$. Hence on $\big[t_0, - \frac{\epsilon}{16}\big]$, $U_{Jj}(t)$ generates a cylinder of radius $\frac{\epsilon}{8}$ and height $\frac{1}{2}(\alpha_{Jj} - \frac{2}{n} + \frac{3\eps}{4n})$. Moreover, for all $t \in \big[ \frac{\epsilon}{16}, t_1\big]$, $U_{Jj}(t) = t + \frac{\sqrt{2}}{8}\epsilon$ is perpendicular to the axis of rotation $y(t) = -t$. Therefore on $\big[\frac{\epsilon}{16}, t_1\big]$, $U_{Jj}(t)$ generates a disk with radius $\frac{\epsilon}{4}$. Thus, as $U_{Jj}(t)$ is a smooth curve on $(t_0, t_1)$, then $L_{Jj}$ is a solid whose boundary has a smooth transition from a cylinder to a flat disk. See Figure \ref{fig:U}. Let $L_{jJ}$ be constructed in the same manner as $L_{Jj}$.

Finally, we construct $M$. Consider some $(J, j) \in \cal J$. Recall that the flattened region $D_{Jj}$ on $B_J$ is of radius $\frac{\epsilon}{4}$. Moreover, the flattened base of $L_{Jj}$ is also of radius $\frac{\epsilon}{4}$. Hence, if we glue the flattened base of $L_{Jj}$ to $D_{Jj}$, we have created a solid $B_J \cup L_{Jj}$ whose boundary smoothly transitions from a ball to a cylinder. Similarly, we do the same for $B_j$ and $L_{jJ}$. 

Recall that the distance between $B_{1/n}(P_J)$ and $B_{1/n}(p_j)$ is $\alpha_{Jj} - \frac{2}{n}$. Observe that by flattening each ball by $\frac{\epsilon}{2n}$, the distance between $B_J$ and $B_j$ is $(\alpha_{Jj} - \frac{2}{n} + \frac{3\epsilon}{4n})$. Hence, because we have constructed the height of $L_{Jj}$ and $L_{jJ}$ to be $\frac{1}{2}(\alpha_{Jj} - \frac{2}{n} + \frac{3\epsilon}{4n})$, then the gluing of $(B_J \cup L_{Jj})$ with $(B_j \cup L_{jJ})$ is well-defined. Let $C_{Jj}$ denote the circle of radius $\frac{\epsilon}{8}$ created by the intersection of the boundary of $L_{Jj}$ and the boundary of $L_{jJ}$. Thus, the solid
\[  
    \bigcup_{(J, j) \in \cal J} \bigg( \big(L_{Jj} \cup L_{jJ}\big) \cup B_J \cup B_j \bigg).
\]
has smooth boundary. Finally, we define $M$ to be the boundary of this solid.

Recall that $\{p_j\}_{j=1}^n$ is a set of $n$ evenly spaced points on the unit circle in the $(x_3 = 0)$-plane and that $\{P_J : J \in \scr P(\{1, ..., n\}), J \neq \emptyset\}$ is a set of points along the positive $x_3$-axis. Hence by symmetry, for non-empty $J \in \scr P(\{1, ..., n\})$, the point of intersection of each $\ell_{Jj}$ with $B_{1/n}(P_J)$ has the same $x_3$-value for all $j \in J$. In other words, the points of intersection all lie on a plane parallel to the $(x_3 = 0)$-plane. Let $\tilde P_j \in M$ be a point on the boundary of $B_{1/n}(P_j)$ such that the vector from $P_j$ to $\tilde P_j$ is perpendicular to the aforementioned plane. 

Analogously, observe that for a fixed $j \in \{1, ..., n\}$, the lines $\{\ell_{Jj}\}_{j \in J}$ are co-planar. Hence their points of intersection with $B_{1/n}(p_j)$ are co-planar. Let $\tilde p_j \in M$ be a point on the boundary of $B_{1/n}(p_j)$ such that the vector from $p_j$ to $\tilde p_j$ is perpendicular to the plane. 

Lastly, we define a Riemannian metric $\tilde g$ on $M$. Let $\tilde \alpha_{Jj} = \big(\frac{n}{n-8}\big)\alpha_{Jj}$. Define $f : \R^3 \to \R$ by
\[f(x) =  \prod_{\substack{(J, j) \in \cal J \\J \neq \emptyset}} f_{Jj}(x)\]
where each $f_{Jj}$ is given by
\[f_{Jj}(x) =  \delta_{Jj}(x) \left(1 - \frac{1}{\tilde \alpha_{Jj}^2} \right) + \frac{1}{\tilde \alpha_{Jj}^2} \]
and each $\delta_{Jj}$ is as defined in the proof of Theorem \ref{thm:decoupling_VCdim_in_R3}. Let $\tilde g := fg$. Thus, we have constructed a 2-dimensional Riemannian sub-manifold $(M, \tilde g)$ for which we will now show the following. 

\begin{center}
    \begin{minipage}{0.8\textwidth}
    \begin{enumerate} [label=Claim \arabic*:]
        \item For all  nonempty $J \in \scr P(\{1, ..., n\})$, $\Dist{\tilde g}{\tilde P_J, \tilde p_j} \leq 1$ if $j \in J$.
        \item For all  nonempty $J \in \scr P(\{1, ..., n\})$, $\Dist{\tilde g}{\tilde P_J, \tilde p_k} > 1$ if $k \notin J$. 
        \item If $J = \emptyset$, $\Dist{\tilde g}{\tilde P_\emptyset, \tilde p_k} > 1$ for all $k \in \{1, ..., n\}$.
    \end{enumerate}
    \end{minipage}
\end{center}

\textbf{Claim 1:} Fix some non-empty $J \in \scr P(\{1, ..., n\})$ and $j \in J$. There exists some curve $\gamma$ on $M$ from $\tilde P_J$ to $\tilde p_j$ such that $\arclength{\tilde g}{\gamma} \leq 1$. 
        
    \begin{claimproof} 
    We define a curve $\gamma = \sum_{i=1}^5 \gamma_i$ from $\tilde P_J$ to $\tilde p_j$ in five parts. 
        
    We begin with the curve $(\gamma_1+\gamma_2)$ from $\tilde P_J$ to $C_{Jj}$ for which $\gamma_1$ is a curve on $B_J$ and $\gamma_2$ is a curve on $L_{Jj}$. Geometrically, one can describe $(\gamma_1+\gamma_2)$ as the curve running along the intersection of $B_J \cup L_{Jj}$ with the plane containing $\tilde P_J$ and $\ell_{Jj}$. \\
    \hspace*{5mm} Analogously, we define the curve $(\gamma_4 + \gamma_5)$ from $\tilde p_j$ to $C_{Jj}$ for which $(\gamma_4 + \gamma_5)$ will follow the intersection curve of $B_j \cup L_{jJ}$ and the plane containing $\tilde p_j$ and $\ell_{Jj}$. Note that by symmetric argument, the arclength of $(\gamma_1 + \gamma_2)$ is equal to the arclength of $(\gamma_4 + \gamma_5)$. \\
    \hspace*{5mm} However, as the aforementioned planes are perpendicular to each other, the end point of $(\gamma_1 +\gamma_2)$ is not the same point that $(\gamma_4 + \gamma_5)$ begins with. Thus, we complete the construction of $\gamma$ by defining $\gamma_3$ to be the quarter rotation along $C_{Jj}$ which connects the end point of $(\gamma_1 +\gamma_2)$ to the start point of $(\gamma_4 + \gamma_5)$. See Figure \ref{fig:gamma_on_M}.

    \begin{figure}
    \centering
    \includegraphics[width=0.6\linewidth]{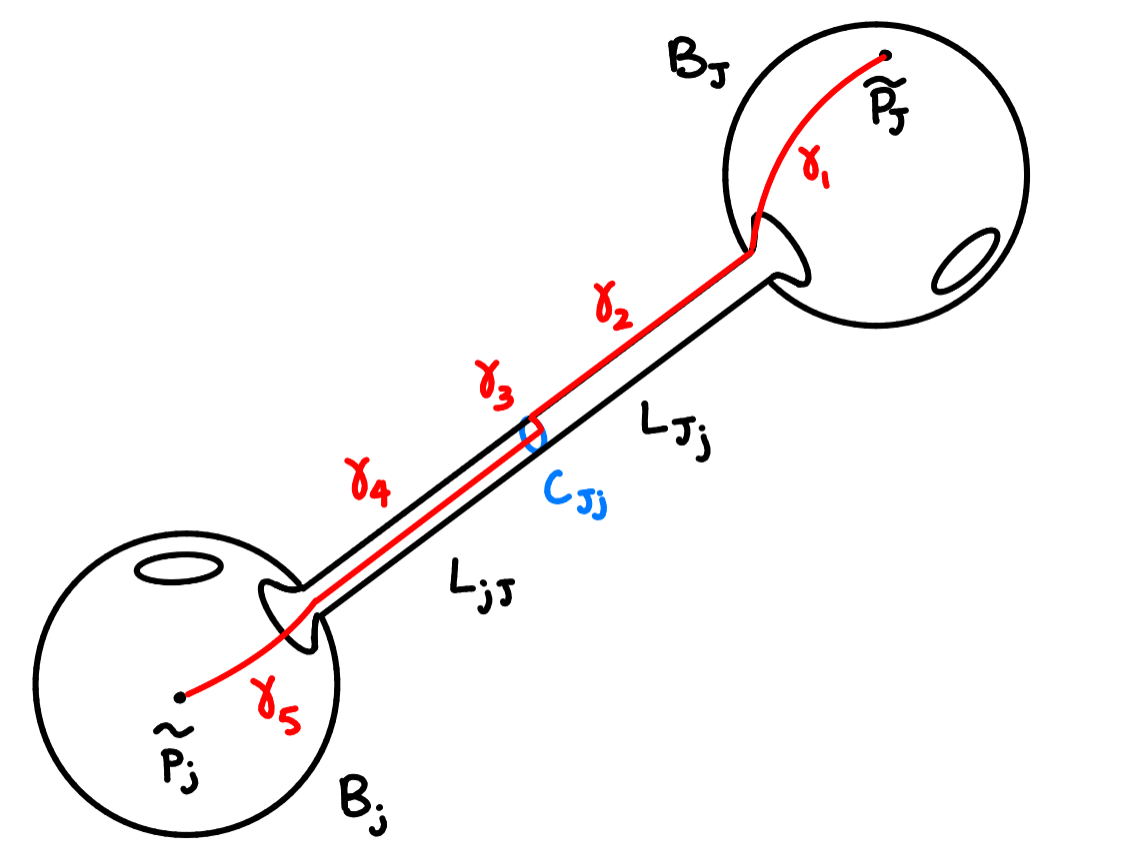}
    \caption{The five subcurves that form $\gamma = \gamma_1 + ... + \gamma_5$.}
    \label{fig:gamma_on_M}
\end{figure}
        
    Let us define $\gamma_1$. Consider the geodesic $\tilde \gamma_1$ on $B_{1/n}(P_J)$ from $\tilde P_J$ to the point of intersection of $B_{1/n}(P_J)$ and $\ell_{Jj}$. Note that the distance between any two points on a sphere of radius $\frac{1}{n}$ is at most $\frac{\pi}{n}$. Therefore, because $\tilde \gamma_1$ is defined to be a geodesic, $\arclength{g}{\tilde \gamma_1} \leq \frac{\pi}{n}$. 
    
    Let $\gamma_1$ be the nearest-point projection of $\tilde \gamma_1$ onto $B_J$. As $B_J$ is convex, we apply Lemma \ref{lm:nearest_point_projection} (given at the end of this proof) and get that
    \[\arclength{g}{\gamma_1} \leq \arclength{g}{\tilde \gamma_1} \leq \frac{\pi}{n} < \frac{4}{n}.\]
    Furthermore, because $f(x) \leq 1$ for all $x \in \R^3$, then 
    \begin{equation} \label{eq:d=2_claim1_gamma_1}
        \arclength{\tilde g}{\gamma_1} \leq 1 \cdot \arclength{g}{\gamma_1} < \frac{4}{n}.
    \end{equation}
    Lastly, we redefine $\gamma_1$ to be the restriction of itself from $\tilde P_J$ to the boundary of $D_{Jj}$. Note that because we are restricting $\gamma_1$, we preserve the upper bound on its arclength given in Equation (\ref{eq:d=2_claim1_gamma_1}). Finally, $\gamma_1$ is a well-defined curve on $B_{1/n}(P_J)$, but to see that it is well-defined on $M$ after we glue on the $L_{Ji}$ to $B_{1/n}(P_J)$, we utilize the fact that $\tilde P_J$ was defined to be orthogonal to the centers of all $D_{Ji}$, $i \in J$. Hence $\gamma_1 \cap D_{Ji} = \emptyset$ for all $i \neq j$. Therefore, $\gamma_1 \cap L_{Ji} = \emptyset$ for all $i \neq j$. Thus, $\gamma_1$ is well-defined on $M$ with arclength at most $\frac{4}{n}$.

    Next, we define $\gamma_2$. Recall $U_{Jj}$ used to define the solid of revolution $L_{Jj}$. Let $\gamma_2$ be the meridian $U_{Jj}$ running along boundary of $L_{Jj}$ such that $\gamma_2(0) = \gamma_1(1)$. Hence, by definition, the arclength of $\gamma_2$ is equal to the arclength of $U_{Jj}$. As $U_{Jj}$ is the mollification of a convex function, the epigraph $\{(t, s) : t \in [t_0, t_1], s \geq U_{Jj}(t)\}$ is convex. Thus, because the nearest-point projection of $V$ over $[t_0, t_1]$ onto the convex epigraph is $U_{Jj}$, then by Lemma \ref{lm:nearest_point_projection}, 
    \[\arclength{g}{\gamma_2} = \arclength{g}{U_{Jj}} \leq \arclength{g}{V} = \frac{1}{2}\left(\alpha_{Jj} - \frac{2}{n} + \frac{3\epsilon}{4n} \right) + \frac{\epsilon}{4} < \frac{\alpha_{Jj}}{2} - \frac{1}{n} + \frac{3\epsilon}{4} .\]
    Note that the last inequality holds because we assumed $n \geq 40$.
    Next, in order to bound the arclength of $\gamma_2$ with respect to $\tilde g$, we first observe the following. Recall that $f_{Ii}(x) = 1$ for all $x$ such that $\Dist{g}{x, \ell_{Ii}} > \frac{3\epsilon}{4}$, $f_{Jj}(x) = \frac{1}{\tilde \alpha_{Jj}^2}$ for $x \in (\ell_{Jj})_{\epsilon/4}^g$, and all $(\ell_{Ii})_{\epsilon/4}^g$ do not intersect outside of the balls of radius $1/n$ based at the end points of $\ell_{Ii}$. Hence because $L_{Jj}$ is constructed so that all points in $L_{Jj}$ are at most $\frac{\epsilon}{4}$ away from $\ell_{Jj}$, then $f(\gamma_2(t)) \leq \frac{1}{\tilde \alpha_{Jj}^2}$ for all $t \in [t_0, t_1]$. Therefore
    \begin{equation} \label{eq:d=2_claim1_gamma_2}
    \arclength{\tilde g}{\gamma_2} \;=\; \frac{1}{\tilde \alpha_{Jj}} \cdot \arclength{g}{\gamma_2} \;=\; \frac{1}{\tilde \alpha_{Jj}} \cdot\left(\frac{\alpha_{Jj}}{2} - \frac{1}{n} +   \frac{3\epsilon}{4} \right).
    \end{equation}

    As previously described, now that we have defined $(\gamma_1 + \gamma_2)$, we define $\gamma_4$ to be a curve on $B_j$ analogous to $\gamma_1$ and define $\gamma_5$ to be a curve on $L_{jJ}$ analogous to $\gamma_2$. Lastly, let $\gamma_3$ be the quarter arc of the intersection circle $C_{Jj}$ that connects the end point of $(\gamma_1 + \gamma_2)$ to the start point of $(\gamma_4 + \gamma_5)$. As $C_{Jj}$ is a circle of radius $\frac{\epsilon}{8}$ on $L_{Jj} \cap L_{jJ}$, then $f(\gamma_3(t)) \leq \frac{1}{\tilde \alpha_{Jj}^2}$ for all $t \in [0, 1]$, and therefore
    \begin{equation}\label{eq:d=2_claim1_gamma_3}
        \arclength{\tilde g}{\gamma_3} \;\leq\; \frac{1}{\tilde \alpha_{Jj}} \cdot \left(\frac{\pi}{2} \cdot \frac{\epsilon}{4}\right) \;<\; \frac{1}{\tilde \alpha_{Jj}} \cdot \frac{\epsilon}{2} 
    \end{equation}
    
    Finally, we now show that the arclength of $\gamma$ with respect of $\tilde g$ is indeed at most 1 using Equations (\ref{eq:d=2_claim1_gamma_1}), (\ref{eq:d=2_claim1_gamma_2}), and (\ref{eq:d=2_claim1_gamma_3}). 
    \begin{align*}
        \arclength{\tilde g}{\gamma} 
         &\leq \sum_{i=1}^5 \;\arclength{\tilde g}{\gamma_i}  \\
        &= 2 \cdot \arclength{\tilde g}{\gamma_1}  +  2 \cdot \arclength{\tilde g}{\gamma_2} + \arclength{\tilde g}{\gamma_3}  \\
        &\leq  2 \cdot \left( \frac{4}{n} \right) + 2 \cdot \left(\frac{1}{\tilde \alpha_{Jj}} \cdot\left(\frac{\alpha_{Jj}}{2} - \frac{1}{n} + \frac{3\epsilon}{4}\right) \right) + \bigg(\frac{1}{\tilde \alpha_{Jj}} \cdot \frac{\epsilon}{2} \bigg) \\
        &= \frac{8}{n} + \frac{1}{\tilde \alpha_{Jj}} \cdot \left(\alpha_{Jj} - \frac{2}{n} + 2 \epsilon \right)  \\ 
        &\leq \frac{8}{n} + \frac{1}{\tilde \alpha_{Jj}} \cdot \alpha_{Jj} &\text{since $\epsilon < \frac{1}{n}$}\\ 
        &= \frac{8}{n} + \frac{n-8}{n} = 1 & \text{by definition of $\tilde \alpha_{Jj}$.}
    \end{align*}
    \end{claimproof}
    
\textbf{Claim 2:} Fix some nonempty $J \in \scr P(\{1, ..., n\})$ and $k \notin J$. Let $\gamma$ be a curve on $M$ from $\tilde P_J$ to $\tilde p_k$. Then $\arclength{\tilde g}{\gamma} > \frac{6}{5}$. 
        
    \begin{claimproof}
    Note that the arclength of any curve on $M$ from $\tilde P_J$ to $\tilde p_j$ must be larger than $\Dist{g}{B_{1/n}^g(P_J), B_{1/n}^g(p_j)}$. 
    Hence using an analogous argument to the proof of \textit{Claim 2} in Theorem \ref{thm:decoupling_VCdim_in_R3}, there exists some $j \in J$ and some $K \in \scr P(\{1, ..., n\})$ with $k \in K$ such that
     \begin{align*}
        \arclength{\tilde g}{\gamma} 
        &\geq \frac{1}{\tilde \alpha_{Jj}} \cdot \Dist{g}{B_{1/n}^g(P_J), B_{1/n}^g(p_j)} + \frac{1}{\tilde \alpha_{Kk}} \cdot \Dist{g}{B_{1/n}^g(P_K), B_{1/n}^g(p_k)} \\
        &= \frac{1}{\tilde \alpha_{Jj}} \left(\alpha_{Jj} - \frac{2}{n}\right) +  \frac{1}{\tilde \alpha_{Kk}} \left(\alpha_{Kk} - \frac{2}{n}\right)
    \end{align*}
    In the proof of Theorem \ref{thm:decoupling_VCdim_in_R3}, we showed that 
    $\displaystyle \frac{1}{\alpha_{Ii}} \left(\alpha_{Ii} - \frac{2}{n} \right) \geq \frac{3}{5}$ for all $(I, i) \in \cal J$.
    We now show that despite $\tilde \alpha_{Ii} > \alpha_{Ii}$, we still have that $\displaystyle  \frac{1}{\tilde \alpha_{Ii}} \left(\alpha_{Ii} - \frac{2}{n} \right) \geq \frac{3}{5}$.
    % keep displaystyle because tilde alpha is visually hard to differentiate from alpha
    
    First, because we assumed that $n \geq 40$, it follows that 
    \begin{equation} \label{eq:d=2_claim2_1}
        \frac{n-8}{n} \geq \frac{40-8}{40} = \frac{4}{5}.
    \end{equation} 
    Similarly, as $\alpha_{Ii} > 1$ and $n \geq 40$, then     
    \begin{equation} \label{eq:d=2_claim2_2}
    \frac{n \cdot \alpha_{Ii} - 2}{n \cdot \alpha_{Ii}} = \frac{n - \frac{2}{\alpha_{Ii}}}{n} \geq \frac{n - 2}{n} \geq \frac{40 - 2}{40} \geq \frac{4}{5}.
    \end{equation}
    Therefore, by Equations  (\ref{eq:d=2_claim2_1}) and (\ref{eq:d=2_claim2_2}), for all $(I, i)\in \cal J$, 
    \begin{align*}
        \frac{1}{\tilde \alpha_{Ii}}\left(\alpha_{Ii} - \frac{2}{n}\right)
        &= \left( \frac{n - 8}{n\cdot \alpha_{Ii}} \right) \left(\alpha_{Ii} - \frac{2}{n}\right)
        =  \left( \frac{n - 8}{n\cdot \alpha_{Ii}}\right) \left(\frac{n \cdot \alpha_{Ii}-2}{n}\right) \\
        &=  \left( \frac{n - 8}{n}\right) \left(\frac{n \cdot \alpha_{Ii}-2}{n\cdot \alpha_{Ii}}\right) 
        \geq \left( \frac{4}{5} \right)\left( \frac{4}{5} \right) > \frac{3}{5}. 
    \end{align*}
    Thus, we have that
    \begin{align*}
        \arclength{\tilde g}{\gamma} 
        \geq \frac{1}{\tilde \alpha_{Jj}} \left(\alpha_{Jj} - \frac{2}{n}\right) +  \frac{1}{\tilde \alpha_{Kk}} \left(\alpha_{Kk} - \frac{2}{n}\right) 
        > \frac{3}{5} + \frac{3}{5} = \frac{6}{5}.
    \end{align*}
    \end{claimproof}

\textbf{Claim 3:} Fix some $k \in \{1, ..., n\}$. Let $\gamma$ be a curve on $M$ from $\tilde P_\emptyset$ to $\tilde p_k$. Then $\Dist{\tilde g}{P_\emptyset, p_k} \geq \frac{6}{5}$. 

\begin{claimproof}
    Note that any curve on $M$ from $\tilde P_\emptyset$ to $\tilde p_k$ must have a subcurve which spans across $(L_{\emptyset,1} \cup L_{1, \emptyset})$. Recall that
    \[f(x) = \prod_{\substack{J \in \scr P(\{1, ..., n\}) \\J \neq \emptyset}} f_{Jj}(x)\]
    is only defined for non-empty $J$. Hence $f(x) = 1$ for all $x \in (L_{\emptyset,1} \cup L_{1, \emptyset})$. Thus, 
    \[
        \arclength{\tilde g}{\gamma} \geq 1 \cdot \Dist{g}{B_{1/n}(P_\emptyset), B_{1/n}(p_k)} \geq \frac{6}{5}
    \]
    where the last inequality holds because $P_\emptyset = (0, 0, -2)$ and all $p_k$ lie on the $x_1x_2$-plane.
\end{claimproof}

Therefore, by Claim 1, Claim 2, and Claim 3, we have proved that $\cal H_1$ shatters $\{\tilde p_j\}_{j = 1}^n$. Thus we have proven the claim. 
\end{proof}

\begin{lemma} \label{lm:nearest_point_projection}
    Let $S$ be a  closed, convex set in $\R^d$, and let $\tilde \gamma$ be a curve in $\R^d \setminus S$. Then the arclength of curve $\gamma$ created by the nearest-point-projection of $\tilde \gamma$ onto $S$ is at most the arclength of $\tilde \gamma$.
\end{lemma}

\begin{proof}
    On convex sets, nearest-point projections are non-expansive. Therefore
    \begin{align*}
        ||\gamma'(t)||^g 
        &= \left|\left|\; \lim_{h \to 0}\; \frac{\gamma(t+h) - \gamma(t)}{h} \;\right|\right|^g\\
        &=  \lim_{h \to 0} \left[\; \frac{||\gamma(t+h) - \gamma(t)||^g}{h}\;\right] & \text{$||\cdot||^g$ is continuous and the limit exists} \\
        &\leq  \lim_{h \to 0} \left[\; \frac{||\tilde \gamma(t+h) - \tilde \gamma(t)||^g}{h} \;\right] &\text{by non-expansiveness of projections on $S$} \\
        &=  \left|\left|\; \lim_{h \to 0}\; \frac{\tilde \gamma(t+h) - \tilde \gamma(t)}{h} \;\right|\right|^g & \text{$||\cdot||^g$ is continuous and the limit exists} \\\\
        &=  ||\tilde \gamma'(t)||^g.
    \end{align*}
    Thus, we immediately see that
    \[
    \arclength{g}{\gamma} = \int_0^1 ||\gamma'(t)||^g \,dt \leq \int_0^1 ||\tilde \gamma'(t)||^g \,dt = \arclength{g}{\tilde \gamma}.
    \qedhere
    \]
\end{proof}

\section{Generalization of Metric Ball Classifiers}

\subsection{VC-dimension based generalization bounds}

\begin{proof}[Proof of Theorem \ref{thm:probability_bounds_using_VCdim}]
We present the bound for $k=1$ as the notation is more transparent; the general case follows by applying the bound \eqref{eq vc-dimension of unions} of \cite{blumer1989learnability} with the same computation. Specifically:
    \begin{align*}
        \P\left(\sup_{x\in X} \big|\mu(B_r(x)) - \hat\mu_n(B_r(x))\big| \geq \epsilon \right) 
        &\leq \frac{1}{\epsilon} \cdot \E \left[ \sup_{x\in X} \big|\mu(B_r(x)) - \hat\mu_n(B_r(x))\big| \right] \\
        &=\frac{1}{\epsilon} \cdot \E \left[ \sup_{h \in \cal H} \left|\int h\,d\mu - \int h\,d\hat \mu_n\right| \right]  \\
        &\leq \frac{1}{\epsilon} \cdot C \sqrt{\frac{\vcdim(\cal H)}{n}}
    \end{align*}
    where the last inequality follows from \cite[Theorem 8.3.15]{vershynin2019high} for a universal constant $C$.
\end{proof}

\subsection{Covering-based generalization bounds: An abstract result}

We recall an auxiliary result on the convergence of empirical measures.

\begin{lemma} \label{lm:empirical_measure_and_true_measure_bound}
Let $(X, \mathcal A, \mu)$ be a probability space and $S_1, ..., S_N \in \mathcal A$. Let $x_1, ..., x_n \in X$ be i.i.d.\ samples drawn with respect to $\mu$. Let $\hat \mu_n = \frac{1}{n}\sum_{i=1}^n \delta_{x_i}$ be the empirical measure. Then
\[
\E \left[ \underset{1 \leq j \leq N}{\max} \Big| \hat \mu_n(S_j) - \mu(S_j) \Big| \right] \leq  \sqrt{\frac{\log 2N}{2n}}.
\]
\end{lemma}

\begin{proof}
    For fixed $j \in \{1, ..., N\}$, $\one_{S_j}(x)$ is a Bernoulli random variable with mean $\mu(S_j)$. Therefore $-\mu(S_j)/n \leq \big(\one_{S_j}(x) - \mu(S_j)\big)/n \leq (1-\mu(S_j))/n$ is a bounded random variable with mean zero and a range of width $1/n$. Hence
\begin{align} \label{eq:moment_of_Y}
\E\left[ \exp \Big(\lambda \cdot \big(\hat \mu_n(S_j) - \mu(S_j) \big) \Big)\right] 
    \notag &= \E \left[\exp \left(\frac{\lambda}{n} \cdot \sum_{i=1}^n \Big(\one_{S_j}(x_i) - \mu(S_j) \Big) \right)\right]\\
    \notag &= \prod_{i=1}^n \E \left[\exp \left(\frac{\lambda}{n} \cdot \Big(\one_{S_j}(x_i) - \mu(S_j)\Big) \right) \right]& \text{by independence} \\
    \notag &\leq \prod_{i=1}^n \exp \left( \frac{\lambda^2}{8n^2}\right)  &\text{by Hoeffding's Lemma} \\
        &= \exp \left( \frac{\lambda^2}{8n}\right)
\end{align}
Let $Y_j = \hat \mu_n(S_j) - \mu(S_j)$ and $Y_{N+j} = -Y_j$ for $j=1,\dots,N$. Then 
\begin{align*}
        \E \left[ \underset{1 \leq j \leq N}{\max} \Big| \hat \mu_n(S_j) - \mu(S_j) \Big| \right]
        &= \frac{1}{\lambda}  \log 
        \exp \left( \lambda\, \E \left[ \underset{1 \leq j \leq 2N}{\max} Y_j  \right] \right)  \\
        &\leq \frac{1}{\lambda} \log \left( \E \left[\exp \left(  \lambda \cdot \underset{1 \leq j \leq 2N}{\max} Y_j  \right) \right] \right) & \text{by Jensen's Inequality} \\
        &\leq \frac{1}{\lambda} \log \left( \sum_{j=1}^{2N} \E\big[ e^{\lambda Y_j}\big] \right) \\
        &\leq \frac{1}{\lambda} \cdot \log \left(2N\exp \left(\frac{\lambda^2}{8n}  \right) \right) & \text{by Equation (\ref{eq:moment_of_Y})}\\
        &= \frac{\log(2N)}{\lambda} + \frac{\lambda}{8n}.
    \end{align*}

    Optimizing for $\lambda$, we find that $\lambda = \sqrt{8n \log(2N)}$ gives us
    \[
    \E \left[ \underset{1 \leq j \leq N}{\max} \Big| \hat \mu_n(S_j) - \mu(S_j) \Big| \right] \leq \sqrt{\frac{\log(2N)}{2n}}.\qedhere
    \]
\end{proof}

We are now in a position to prove the abstract covering result. 

\begin{proof}[Proof of Theorem \ref{theorem generalization covering}]
We first prove the result for a union of $k=1$ balls (i.e.\ for metric ball classifiers) and sketch below how to generalize the result.

Let $\eps > 0$. Pick some positive $\delta < \min\left(\frac{\eps}{4C}, \delta_0 \right)$. By the lower Ahlfors condition, there exists a finite $\delta$-cover with centers at $\hat x_1, \dots, \hat x_N \in X$, $N\leq \frac1c (2/\delta)^d$, i.e.\ $X = \bigcup_{i=1}^N B_\delta(\hat x_i)$. 

To see this, consider a maximal set of points $\mathcal X$ such that the balls $B_{\delta/2}(x)$ are disjoint for all $x\in\mathcal X$ (i.e.\ adding any additional points would violate this property). Since $X$ is separable (and hence second countable), any such subset is automatically at most countable. We observe that $X = \bigcup_{x\in\mathcal X} B_\delta(x)$ since otherwise, we could add any $z\in X\setminus \bigcup_{x\in\mathcal X} B_\delta(x)$ to $\mathcal X$ and make the set strictly larger. By the Ahlfors property and the $\sigma$-additivity of measures, we have
\[
1 = \nu(X) \geq \nu\left(\bigcup_{x\in\mathcal X} B_{\delta/2}(x)\right) = \sum_{x\in\mathcal X}\nu(B_{\delta/2}(x)) \geq \sum_{x\in\mathcal X} c(\delta/2)^d = |\mathcal X| \cdot c(\delta/2)^d.
\]

For every $x \in X$, let $\displaystyle \hat x = {\text{argmin}} \{d(x, \hat x_i) : 1 \leq i \leq N\}$. Observe that because $\hat x_1, ..., \hat x_N$ forms a $\delta$-cover, $d(x, \hat x) < \delta$. Hence for all $y \in B_r(x)$, $d(\hat x, y) \leq d(\hat x, x) + d(x, y) < \delta + r$. That is, $B_r(x) \subseteq B_{r+\delta}(\hat x)$. Similarly, $B_{r-\delta}(\hat x) \subseteq B_r(x)$. Therefore,
    \begin{equation} \label{eq:delta_cover}
         \mu \big(B_r(x)\big) - \mu \big(B_r(\hat x)\big) \leq \mu \big(B_{r+\delta}(\hat x)\big) - \mu \big(B_{r-\delta}(\hat x)\big) = \mu \big(B_{r+\delta}(\hat x) \setminus B_{r-\delta}(\hat x)\big).
    \end{equation}
    Let $S_r^\delta(\hat x) = B_{r+\delta}(\hat x) \setminus B_{r-\delta}(\hat x)$. Then
\begin{align*}
\Big|\mu&\big(B_r(x)\big) - \hat\mu_n\big(B_r(x)\big)\Big| \\
        &\leq\;  \Big|\mu\big(B_r(x)\big) - \mu\big(B_r(\hat x)\big)\Big| + \Big|\mu\big(B_r(\hat x)\big) - \hat \mu_n\big(B_r(\hat x)\big)\Big| + \Big | \hat \mu_n \big(B_r(\hat x)\big) - \hat \mu_n\big(B_r(x)\big)\Big|\\
        &\leq\; \Big|\mu\big(S_r^\delta(\hat x)\big)\Big| + \Big|\mu\big(B_r(\hat x)\big) - \hat \mu_n\big(B_r(\hat x)\big)\Big| + \Big | \hat \mu_n \big(S_r^\delta(\hat x)\big)\Big| & \\%\text{by Equation (\ref{eq:delta_cover})} \\
        &\leq\ 2\Big|\mu\big(S_r^\delta(\hat x)\big)\Big| + \Big|\mu\big(B_{r}(\hat x)\big) - \hat \mu_n\big(B_{r}(\hat x)\big)\Big| + \Big | \mu\big(S_r^\delta(\hat x)\big) - \hat \mu_n \big(S_r^\delta(\hat x)\big)\Big|\\% &\text{by triangle inequality}\\
        &\leq\; 2C\delta + \Big|\mu\big(B_{r}(\hat x)\big) - \hat \mu_n\big(B_{r}(\hat x)\big)\Big| + \Big | \mu\big(S_r^\delta(\hat x)\big) - \hat \mu_n \big(S_r^\delta(\hat x)\big)\Big| .%& \text{by assumption on $\mu$.}
    \end{align*}
    Thus,
    \[
       \sup_{x \in X} \Big|\mu\big(B_r(x)\big) - \hat\mu_n\big(B_r(x)\big)\Big| \leq \max_{1 \leq j \leq N} \left( 2C\delta + \Big|\mu\big(B_{r}(\hat x_j)\big) - \hat \mu_n\big(B_{r}(\hat x_j)\big)\Big| + \Big | \mu\big(S_r^\delta(\hat x_j)\big) - \hat \mu_n \big(S_r^\delta(\hat x_j)\big)\Big| \right) 
    \]
    Recall that $\delta < \eps/4C$, so $\eps - 2C\delta > \eps/2$.
    Finally, by Lemma \ref{lm:empirical_measure_and_true_measure_bound}, we have that
    \begin{align*}
        \P&\left(\sup_{x\in X} \big|\mu(B_r(x)) - \hat\mu_n(B_r(x))\big| \geq \epsilon \right) \\
        &\leq\; \P \left( \max_{1 \leq j \leq N} \Big( 2C\delta + \Big|\mu(B_{r}(\hat x_j)) - \hat \mu_n(B_{r}(\hat x_j))\Big| + \Big | \mu(S_r^\delta(\hat x)) - \hat \mu_n (S_r^\delta(\hat x))\Big|\Big) \geq \epsilon \right) \\
        &\leq\; \P \left( \max_{1 \leq j \leq N} \Big( \Big|\mu(B_{r}(\hat x_j)) - \hat \mu_n(B_{r}(\hat x_j))\Big| + \Big | \mu(S_r^\delta(\hat x)) - \hat \mu_n (S_r^\delta(\hat x))\Big|\Big) \geq \frac\eps2 \right) \\
        &\leq\; \P \left( \max_{1 \leq j \leq N}  \Big|\mu(B_{r}(\hat x_j)) - \hat \mu_n(B_{r}(\hat x_j))\Big| \geq \frac\eps4 \right)  + \P \left( \max_{1 \leq j \leq N} \Big| \mu(S_r^\delta(\hat x)) - \hat \mu_n (S_r^\delta(\hat x))\Big| \geq \frac\eps4 \right) \\
        &\leq\; \frac{4}{\epsilon} \cdot  \E \left[ \max_{1 \leq j \leq N}  \Big|\mu(B_{r}(\hat x_j)) - \hat \mu_n(B_{r}(\hat x_j))\Big| \right]  + \frac {4}{\epsilon} \cdot \E \left[ \max_{1 \leq j \leq N} \Big| \mu(S_r^\delta(\hat x)) - \hat \mu_n (S_r^\delta(\hat x))\Big|\Big) \right] \\
        &\leq\; 2\,\frac{4}{\epsilon} \sqrt{\frac{\log(2N)}{2n}}
    \end{align*}
since if the sum of two terms exceeds $\eps/2$, then at least one must exceed $\eps/4$. We now use the Ahlfors bound on the number $N$ of points
\[
N \leq \frac1 c\left(\frac2\delta\right)^d = \frac1c \left(\frac{8C}\eps\right)^d \qquad \Ra \quad \log N \leq \log(1/c) + d \,\log(8C/\eps)
\]
if $\eps$ is so small that $\eps/4C < \delta_0$. Then, since $8/\sqrt 2\leq 6$, we have
\[
\P\left(\sup_{x\in X} \big|\mu(B_r(x)) - \hat\mu_n(B_r(x))\big| \geq \epsilon \right)
\leq 6\,\frac{\sqrt{\log(1/c) + d\,\log(8C/\eps)}}{\eps}\,\frac1{\sqrt n}.
\]

The result is now proved for $k=1$. If $k\geq 2$, we take $k$ centers $x_i$, nearest points $\hat x_i$ in the fixed set of centers, and observe that
\[
\bigcup_{i=1}^k B_r(x_i) \setminus \bigcup_{i=1}^kB_r(\hat x_i) \subseteq \bigcup_{i=1}^k \big(B_r(x_i)\setminus B_r(\hat x_i)\big) \subseteq \bigcup_{i=1}^k \big( B_{r+\delta}(\hat x_i) \setminus B_{r-\delta}(\hat x_i)\big),
\]
so
\[
\mu\left(\bigcup_{i=1}^k B_r(x_i) \setminus \bigcup_{i=1}^kB_r(\hat x_i)\right) \leq \sum_{i=1}^k \mu \big( B_{r+\delta}(\hat x_i) \setminus B_{r-\delta}(\hat x_i)\big) \leq 2Ck\delta
\]
and by an analogous calculation we get 
\begin{align*}
\bigg|\mu&\left(\bigcup_{i=1}^kB_r(x_i)\right) - \hat\mu_n\left(\bigcup_{i=1}^kB_r(x_i)\right)\bigg| \\
    &\leq 2Ck\delta + \left|(\mu-\hat\mu_n)\left(\bigcup_{i=1}^k B_r(\hat x_i)\right)\right| + \left|(\hat\mu_n -\mu) \left(\bigcup_{i=1}^k \big( B_{r+\delta}(\hat x_i) \setminus B_{r-\delta}(\hat x_i)\big)\right)\right|
\end{align*}
The subsets are now indexed by $k$-tuples of points $\hat x_i, \dots, \hat x_k$, of which there are fewer than $(c (2/\delta)^d)^k = c^k (2/\delta)^{dk}$, so assuming that $2Ck\delta < \eps/2$, the computation yields
\[
\P\left(\sup_{x_1,\dots,x_k\in X} \left|\mu\left(\bigcup_{i=1}^kB_r(x_i)\right) - \hat\mu_n\left(\bigcup_{i=1}^kB_r(x_i)\right)\right| \geq \epsilon \right)
\leq 6\,\frac{\sqrt{k\log(1/c) + dk\,\log(8Ck/\eps)}}{\eps\,\sqrt n}.\qedhere
\]
\end{proof}

\subsection{Covering-based generalization bounds: Ambient geometry}

\begin{proof}[Proof of Theorem \ref{theorem generalization ambient balls}]
As $r+\delta<R$ and $O$ is an isometry, it suffices to prove the results locally for graphs.
We observe that for $x\in S^{m-1}\subseteq \R^m$, we have
\[
\frac{d}{dt}\big\|(tx, f(tx))\big\|^2 = \frac{d}{dt} \big(t^2 + \|f(tx)\|^2\big) = 2t + 2\,f(tx)^TDf(tx) x.
\]
The second term can be estimated as
\[
\big|f(tx)^TDf(tx) x\big| \leq\|f(tx)\| \,\|Df(tx)\|_{op}\|x\|\leq Lt\|x\|\,L\,\|x\| = L^2t,
\]
so 
\[
0<2(1-L^2)\,t \leq \frac{d}{dt}\big\|(tx, f(tx))\big\|^2 \leq 2(1+L^2)\,t.
\]
In particular, $\|(x,f(x))\|$ is growing monotonically in radial direction and for every $x\in S^{m-1}$, there exist unique values $r_-(x), r_+(x)$ such that
\[
\big\|(r_\pm x,\ f(r_\pm x))\big\|^2 = (r\pm \delta)^2
\]
with
\[
(r\pm\delta)^2 = \int_0^{r_\pm} \frac{d}{dt} \big\|(tx, f(tx))\big\|^2\dt \geq \int_0^{r_\pm} 2(1-L^2)t\dt = (1-L^2)r_\pm^2 \qquad \Ra\quad r_\pm \leq \frac{r\pm \delta}{\sqrt{1-L^2}}
\]
and analogously $r_\pm \geq \frac{r\pm\delta}{\sqrt{1+L^2}}$. Since $\|(x,f(x))\| \geq \|x\|$, we see that $r_+ \leq r+\delta<R$, i.e.\ $f$ is well-defined up to the radius $r_+$. Finally, we observe that
\[
(r+\delta)^2 = (r-\delta)^2 + \int_{r_-}^{r_+} \frac{d}{dt} \big\|(tx,f(tx))\big\|^2\dt \geq  (r-\delta)^2 +\int_{r_-}^{r_+}2(1-L^2)t\dt  
\]
meaning that 
\[
(r_+ - r_-)(r_++r_-) = r_+^2 - r_-^2 \leq \frac{(r+\delta)^2- (r-\delta)^2}{1-L^2} = \frac{4r\delta}{1-L^2}
\]
and hence 
\[
r_+-r_- \leq \frac{4r\delta }{(1-L^2)(r_++r_-)}\leq \frac{4r\delta}{(1-L^2)(r+\delta + r-\delta)/\sqrt{1+L^2}} = \frac{2\sqrt{1+L^2}}{1-L^2}\,\delta.
\]
Hence, we have
\begin{align*}
\H^m \big( \Gamma_f \cap B_{r+\delta}(0) \setminus B_{r-\delta}(0)\big) 
    &= \int_{S^{m-1}}\int_{r_-(x)}^{r_+(x)}\sqrt{\det\big(\id_{m\times m}+ Df^TDf(sx)\big)}\,s^{m-1}\ds \d\H^{m-1}_x\\
    &\leq m\omega_m \,\frac{2\sqrt{1+L^2}}{1-L^2}\,\delta \,\big(1+L^2)^{m/2}(r+\delta)^{m-1}
\end{align*}
where $m\omega_m$ is the $(m-1)$-dimensional area of $S^{m-1}$, $\frac{2\sqrt{1+L^2}}{1-L^2}\,\delta$ is the upper bound on the width of the shell in radial direction, $(1+L^2)$ is an upper bound on the eigenvalues of $\id_{m\times m} + Df^TDf$, and $r+\delta$ is an upper bound on $s$.
By a simpler version of the same argument, we have 
\begin{align*}
\H^m \big( \Gamma_f \cap B_s(0)\big) &= \int_{(\id,f)^{-1}(\Gamma_f \cap B_s(0))}\sqrt{\det(\id_{m\times m} + Df^TDf)}\dx \geq \int_{B_{s/\sqrt{1+L^2}}} 1\dx \\
&= \frac{\omega_m}{(1+L^2)^{m/2}}\,s^m.
\end{align*}
since $Df^TDf$ is positive semi-definite, i.e.\ all eigenvalues of $\id_{m\times m} + Df^TDf$ are at least $1$.

The first condition of Theorem \ref{theorem generalization covering} is therefore satisfied for the measure $\nu = \frac1v\,\H^m|_{\Gamma_f}$ with the constant $c = \frac{\omega_m}{(1+L^2)^{m/2}\,v}$ and the upper bound on $\mu$ is satisfied with
\[
C = 2m\omega_m (1+L^2)^{(m+1)/2}(1-L^2)^{-1}R^{m-1} A = 2m\omega_m(1+L^2)^{m/2}(1-L)^{-1}R^{m-1}A.\qedhere
\]
\end{proof}

\subsection{Covering-based generalization bounds: Intrinsic geometry}

We begin by presenting generalization bounds for metric ball classifiers on compact manifolds with respect to the intrinsic Riemannian distance.

\begin{proof}[Proof of Corollary \ref{corollary ricci bound}]
Since $(M,g)$ is smooth and $M$ is compact, it has finite volume $v\in(0,\infty)$, a positive injectivity radius $\inj(M,g)$, and there exists $K^*$ such that the sectional curvatures of $M$ are bounded from above by $K^*$. 

Assume that $s< \inj(M,g)$. Then by the G\"unther-Bishop comparison theorem \cite[Theorem 3.101]{gallot1990riemannian}, we have $\H^d(B_s(x)) \geq v(K^*, d, s)$ for all $x\in M$ where $v(K^*, d,s)$ denotes the volume of a ball of radius $s$ in the $d$-dimensional model space of constant curvature $K^*$. Recall that
\begin{equation}\label{eq volume space forms}
v(K^*, d, s) = \begin{cases} d\omega_d \int_0^s \left(\frac{\sin(\sqrt{K^*}\sigma)}{\sqrt{K^*}}\right)^{d-1}\d\sigma  & K^* >0 \text{ and } s< \frac{\pi}{\sqrt{K^*}}\\ \omega_ds^d = d\omega_d \int_0^s\sigma^{d-1}\d\sigma & K^*=0\\ d\omega_d \int_0^s \left(\frac{\sinh(\sqrt{|K^*|}\sigma)}{\sqrt{|K^*|}}\right)^{d-1}\d\sigma& K^*<0\end{cases}
\end{equation}
where $\omega_d$ is the volume of the $d$-dimensional unit ball. The square root of curvature enters as curvature scales with $\lambda^{-2}$ if length is scaled by $\lambda$. The threshold $\pi/\sqrt{K^*}$ is the injectivity radius of the sphere of radius $1/\sqrt{K^*}$. 

Depending on $K^*$, we can choose $\delta_0 \leq \inj(M,g)$ such that $v(K^*, d, s) \geq \frac{\omega_d}2\,s^d$ for all $s< \delta_0$ since $\lim_{s\to 0^+} v(K^*, d,s)/(\omega_ds^d) = 1$ independently of $K^*$ as $\sin'(0) = \sinh'(0) = 1$. Thus the Borel probability measure $\nu(B) = \H^d(B)/v$ satisfies the first assumption of Theorem \ref{theorem generalization covering} with constant $c = \omega_d/2v$ for $\delta<\delta_0$.

Conversely, if $r+\delta < \inj(M,g)$, we deduce from \cite[Lemma 34, Chapter 9]{petersen2006riemannian} that
\begin{equation}\label{eq curvature and volume of shells}
\H^d\big(B_{r+\delta}(x)\setminus B_{r-\delta}(x)\big) \leq v(K, d, r+\delta) - v(K, d, r-\delta),
\end{equation}
just as in the proof of \cite[Lemma 35, Chapter 9]{petersen2006riemannian}. In the case $K^*> 0$, we note that the restriction $s<\pi/\sqrt{K^*}$ in \eqref{eq volume space forms} is automatically satisfied in our situation since
\[
\inj(M,g) \leq \diam_g(M) \leq \frac{\pi}{\sqrt{K}}
\]
by a theorem of Myers, see e.g.\ \cite[Theorem 25]{petersen2006riemannian}. The only ingredients in the proofs of \cite[Lemma 35]{petersen2006riemannian} are \cite[Lemma 34]{petersen2006riemannian}, the radial isometry property of the exponential map and a pointwise inequality on the volume form of $(M,g)$ (which is well-defined on the orientable segment $B_{r+\delta}(x)$ of $M$ and induces the standard volume/Hausdorff measure). It thus applies not only to balls, but also to spherical shells, and \eqref{eq curvature and volume of shells} follows just as \cite[Lemma 35]{petersen2006riemannian}.
Now we can estimate
\[
\mu\big(B_{r+\delta}(x)\setminus B_{r-\delta}(x)\big) \leq Ad\omega_d \int_{r-\delta}^{r+\delta}\phi_{K}(s)^{d-1}\ds\quad\text{where } \phi_{K}(s) = 
\begin{cases}
\sin(\sqrt{K}s)/\sqrt{K} & K>0\\ s&K=0\\\sinh(\sqrt{|K|}s)/\sqrt{|K|} & K<0.
\end{cases}
\]
If $\delta < r$, the somewhat coarse bound in the statement of the result is achieved.
\end{proof}

The second proof in this appendix concerns situations where intrinsic distances are only approximated.

\begin{proof}[Proof of Theorem \ref{theorem graph approximation of sub-manifold}] 
The proof of Corollary \ref{corollary ricci bound} establishes that the assumptions of Theorem \ref{theorem generalization covering} are satisfied with constants $c, C$ and uses that the conclusion of Theorem \ref{theorem generalization covering} applies. In this version of the result, we exploit that the assumptions also apply.

Note that 
\[
x_j \in B_i \qquad \LRa \quad d_{ij} < r \qquad \Ra\quad r-\alpha \eps/k < d_{(M,g)}(x_i, x_j) < r + \alpha\eps/k,
\]
with probability at least $1-\gamma$, so
\[
\hat \mu_n(B_{r-\alpha\eps/k}(x_i)) \leq \hat\mu_n(\{x_j : j\in B_i\}) = \frac{|B_i|}{n} \leq \hat \mu_n(B_{r+\alpha\eps/k}(x_i))
\]
with probability at least $1-\gamma$. 
By assumption, we have
\[
\mu(B_{r+\alpha\eps/k} (x_i)) - \mu(B_r(x_i)) \leq C\,\frac{\alpha\eps}k =\frac\eps {2k}, \qquad \mu(B_{r-\alpha\eps/k} (x_i)) - \mu(B_r(x_i)) \geq -\frac\eps{2k}
\]
and analogously 
\[
\mu\left(\bigcup_{i\in I} B_{r+\alpha\eps/k}(x_i)\right) -\frac\eps2 \leq 
\mu\left(\bigcup_{i\in I} B_r(x_i)\right) \leq \mu\left(\bigcup_{i\in I} B_{r-\alpha\eps/k}(x_i)\right) +\frac\eps2,
\]
i.e.\ it suffices to show that $\hat\mu_n$ is margin within $\eps/2$ on the sets $B_{r\pm \alpha\eps/k}(x_i)$. The result of Lemma \ref{lm:empirical_measure_and_true_measure_bound} cannot be applied directly since the sets on which the measures should be close are not given a priori, but depend on the dataset itself.

We compute
\begin{align*}
\mathbb E &\left[\exp \left(\lambda \big(\hat\mu_n(B_{r+\alpha\eps/k}(x_i)) - \mu(B_{r+\alpha\eps/k}(x_i))\big)\right)\right]\\
    &= \E  \left[\exp \left(\frac\lambda n\left(\sum_{j=1}^n 1_{\{d_{(M,g)}(x_i,x_j)\leq r+\alpha\eps/k\}} - \mu(B_{r+\alpha\eps/k}(x_i))\right)\right)\right]\\
    &= \E\left[\E  \left[\exp \left(\frac\lambda n\left(\sum_{j=1}^n 1_{\{d_{(M,g)}(x_i,x_j)\leq r+\alpha\eps/k\}} - \mu(B_{r+\alpha\eps/k}(x_i))\right)\right) \bigg| x_i\right]\right]\\
    &= \E\left[\prod_{j=1}^n\E\left[ \exp\left(\frac\lambda n\big(1_{\{d_{(M,g)}(x_i,x_j)\leq r+\alpha\eps/k\}} - \mu(B_{r+\alpha\eps/k}(x_i))\big)\right)\bigg|x_i\right]\right]\\
    &\leq \E\left[\exp\left(\frac\lambda n \big(1- \mu(B_{r+\alpha\eps/k}(x_i))\big)\right)\right]\prod_{j\neq i} \exp\left(\frac{\lambda^2}{8n^2}\right)\\
    &\leq \exp\left(1+(n-1)\frac{\lambda^2}{8n^2}\right) \leq \exp(1+\lambda^2/8n)
\end{align*}
since $1_{d_{(M,g)}(x_i,x_i) <r}\equiv 1$ and $1_{d_{(M,g}(x_i,x_j)} - \mu(B_{r+\alpha_\eps/k}(x_i))$ is a sub-Gaussian random variable of mean zero, conditioned on $x_i$.
The same argument applies to 
\[
\mu(B_{r+\alpha\eps/k}(x_i)) - \hat\mu_n(B_{r-\alpha\eps/k}(x_i)).
\]
This gives us $2n$ variables to take the maximum over (as we are only worried about overestimating the larger ball and underestimating the smaller ball). In particular:
\[
\E\left[\max_{1\leq i\leq n} \left|\mu(B_r(x_i)) -  \frac{|B_{i}|}n\right| \right] \leq \sqrt{\frac{\log (2en)}n}.
\]
The Lemma now follows by Markov's inequality in the case $k=1$ as
\begin{align*}
\P&\left(\max_{K\subseteq\{1,\dots,n\},\ |K| = 1}\left|\mu\left(\bigcup_{i\in I}B_r(x_i)\right) - \frac{|\bigcup_{i\in I}B_i|}n\right|\geq \eps\right)\\
    &\leq \P\left(\max_{i,j}|d_{ij}-d_{(M,g)}(x_i,x_j)| \geq \frac{\alpha\eps}k\right)\\
    &\qquad + \P\left(\max_{1\leq i}\max\left\{(\mu - \hat\mu_n)(B_{r-   \alpha\eps/k}(x_i),\ (\mu - \hat\mu_n) (B_{r+\alpha\eps/k}(x_i)\right\}\geq \frac\eps2\right)\\
    &\leq \gamma + \frac2\eps\,\min_{\lambda>0}\frac{\log (2en)}{\lambda} + \frac{\lambda}{8n}\\
    &= \gamma + \frac2\eps\,\sqrt{\frac{\log(2en)}{2n}}.
\end{align*}
For $k>1$, we use the same argument as in Theorem \ref{theorem generalization covering}: While it is not sufficient to show that the estimates are accurate for balls due to the possibility of `overlap', the analysis remains essentially intact since it relies only on symmetric differences having small measure, which is preserved under finite unions.

We consider unions of up to $k$ balls, indexed by $I\in\{0, \dots,n\}^k$. This is an overestimate of the actual size of the index set due to permutation invariance, but mild as we will take the logarithm, and yields the estimates
\[
\E\left[\max_{K\subseteq\{0,\dots, n\},\ |K| = k} \left|\mu\left(\bigcup_{i\in I}B_r(x_i)\right) - \frac{|\bigcup_{i\in I}B_i|}n\right|\right] \leq \sqrt{\frac{k\,\log(2en)}{2n}}
\]
and thus ultimately
\begin{align*}
\P&\left(\max_{K\subseteq\{1,\dots,n\},\ |K| = 1}\left|\mu\left(\bigcup_{i\in I}B_r(x_i)\right) - \frac{|\bigcup_{i\in I}B_i|}n\right|\geq \eps\right) \leq \gamma + \frac1\eps \sqrt{\frac{2k\log(2en)}n}.
\end{align*}
So far, we have considered the direct generalization, estimating discrepancies at a finite collection of centers $x_i$. This may often yield tight bounds in practice, but we observe that the statement of Corollary \ref{corollary ricci bound} yields guarantees for the generalization of {\em all} metric balls with constants which depend on geometric quantities of $M$, but scale with $n^{-1/2}$ rather than $\sqrt{(\log n)/n}$. We assume that the same bounds apply to the radii $r_\pm = r\pm \alpha\eps/k$ and with $\eps/2$, so we choose $\eps$ smaller if needed. By a union bound, the constant additionally increases by a factor $2$.
\end{proof}

\stopcontents[appendix]

\end{document}